\documentclass[11pt]{article}

\usepackage[margin=1in]{geometry}
\usepackage{amsmath, amssymb}
\usepackage{graphicx}
\usepackage{xcolor} 
\usepackage{natbib}
\usepackage{soul}
\usepackage{comment}
\usepackage{xcolor}
\usepackage{booktabs}
\usepackage{amsfonts}
\usepackage{algorithmic}
\usepackage{algorithm}
\usepackage{array}
\usepackage{stfloats}
\usepackage{tabularx}
\usepackage{float}
\usepackage{subcaption}
\usepackage{setspace}
\usepackage{threeparttable}
\usepackage{rotating}
\usepackage{booktabs}  % For better table formatting
\usepackage{multirow}  % Allows multi-row cells
\usepackage{array}     % Allows better column formatting
\usepackage{pdflscape}
\usepackage{pifont}
\usepackage{caption}
\usepackage{longtable}
\usepackage[utf8]{inputenc}
\usepackage{lmodern}
\usepackage{hyperref}

\title{MIRACL: A Diverse Meta-Reinforcement Learning for Multi-Objective Multi-Echelon Combinatorial Supply Chain Optimisation}

\author{
  Rifny Rachman\thanks{Corresponding author. Email: \texttt{rifny.rachman@manchester.ac.uk}}\\
  The University of Manchester, United Kingdom
  \and
  Josh Tingey\\
  Peak AI, Ltd, United Kingdom\\
  \texttt{josh.tingey@peak.ai}
  \and
  Richard Allmendinger\\
  The University of Manchester, United Kingdom\\
  \texttt{richard.allmendinger@manchester.ac.uk}
  \and
  Pradyumn Shukla\\
  The University of Manchester, United Kingdom\\
  \texttt{pradyumn.shukla@manchester.ac.uk}
  \and
  Wei Pan\\
  The University of Manchester, United Kingdom\\
  \texttt{wei.pan@manchester.ac.uk}
  \and
  Bahrul Ilmi Nasution\\
  The University of Manchester, United Kingdom\\
  \texttt{bahrul.nasution@manchester.ac.uk}
}

\date{} % leave empty for arXiv

\begin{document}
\maketitle

\begin{abstract}
\noindent Multi-objective reinforcement learning (MORL) is effective for multi-echelon combinatorial supply chain optimisation, where tasks involve high dimensionality, uncertainty, and competing objectives. However, its deployment in dynamic environments is hindered by the need for task-specific retraining and substantial computational cost. We introduce \textbf{MIRACL} (\textbf{M}eta mult\textbf{I}-objective \textbf{R}einforcement le\textbf{A}rning with \textbf{C}omposite \textbf{L}earning), a hierarchical Meta-MORL framework that allows for a few-shot generalisation across diverse tasks. MIRACL decomposes each task into structured subproblems for efficient policy adaptation and meta-learns a global policy across tasks using a Pareto-based adaptation strategy to encourage diversity in meta-training and fine-tuning. To our knowledge, this is the first integration of Meta-MORL with such mechanisms in combinatorial optimisation. Although validated in the supply chain domain, MIRACL is theoretically domain-agnostic and applicable to broader dynamic multi-objective decision-making problems. Empirical evaluations show that MIRACL outperforms conventional MORL baselines in simple to moderate tasks, achieving up to 10\% higher hypervolume and 5\% better expected utility. These results underscore the potential of MIRACL for robust, efficient adaptation in multi-objective problems.

\end{abstract}


%\item 
%\item 
%\item 

% Keywords
\vspace{1em}
\noindent\textbf{Keywords:} Multi-objective optimisation, Reinforcement Learning, Meta-learning, Combinatorial problems, Supply chain.

%\maketitle
\onehalfspacing
% Main text
\section{Introduction}\label{introduction}
Multi-echelon combinatorial supply chain (SC) optimisation is challenging due to the scale and interdependence of facilities, echelons, and transportation routes, further compounded by conflicting objectives, uncertainty, and parameter fluctuations. Multi-objective reinforcement learning (MORL), formulated as a multi-objective Markov decision process (MOMDP), has been developed to address such settings by learning sequential decisions through continuous agent--environment interaction~\citep{shar_multi-objective_2023,zhao_research_2025,singh_multi-objective_2025,sutton_reinforcement_2015}. MORL can adapt to feedback, respond to operational changes, and recover trade-off solutions, but often requires substantial computation due to exploration-intensive trial-and-error.

In practice, MORL policies are typically specialised to a given SC configuration, so changes in architecture or parameters require retraining. This is problematic in dynamic operations where costs, lead times, and network connectivity can shift (e.g., route disruptions) and decisions must be made quickly. Meta-learning for MORL (Meta-MORL) aims to reduce this burden by \emph{learning how to learn} across tasks with different configurations, enabling rapid adaptation to new instances without repeating full training~\citep{schmidthuber_evolutionary_1987,finn_model-agnostic_2017,thrun_learning_1998}.

However, Meta-MORL remains underexplored in SC: existing work focuses mainly on inventory control~\citep{wang_dynamic_2024} and combinatorial vehicle routing~\citep{zhang_meta-learning-based_2023}. Moreover, prior combinatorial approaches often meta-train over decomposed subproblems within a single problem type, which limits generalisation when both decision variables and parameters vary substantially. These limitations are particularly salient in multi-echelon SC problems, where task heterogeneity is larger, and meta-training can be unstable.

To address this gap, we cast multi-objective, multi-echelon combinatorial SC optimisation as a MOMDP meta-learning problem and propose \textbf{MIRACL} (\textbf{M}eta mult\textbf{I}-objective \textbf{R}einforcement le\textbf{A}rning with \textbf{C}omposite \textbf{L}earning). MIRACL extends Meta-MORL by (i) using hierarchical composite learning to organise adaptation through multiple within-task scalarised subproblems, and (ii) introducing an archive-guided Pareto simulated annealing (PSA) mechanism that perturbs preference weights during meta-training to promote broader PF coverage beyond passive sampling. MIRACL further applies the same PSA-based weight adaptation during fine-tuning, where the weight updates more directly translate into diverse final solutions on the target task. We evaluate MIRACL against conventional MORL methods and a metaheuristic on SC instances of increasing complexity, and analyse the resulting operational behaviours to highlight when each approach is most effective and the trade-offs they induce in practice. Moreover, we demonstrate MIRACL's domain-agnosticism beyond SC.

\section{Related Work} \label{sec:related_work}
Recent MORL advances aim to learn \emph{preference-aware} policies while improving PF coverage. Most remain utility-based, using scalarisation to make trade-offs explicit: envelope Q-learning generalises across linear preferences~\citep{yang_generalized_2019}, and Pareto-conditioned networks capture multiple trade-offs via preference conditioning~\citep{reymond_pareto_2022}. For high-dimensional control, MORL based on decomposition (MORL/D) decomposes learning into scalarised subproblems with continuous policies~\citep{felten_multi-objective_2023}, fitting multi-echelon SC where decisions scale, and preferences shift. Yet MORL alone is sample-intensive and often requires retraining as tasks change, motivating meta-RL for rapid adaptation (e.g., RL$^2$ and MAML)~\citep{duan_rl2_2016,finn_model-agnostic_2017}.

Meta-MORL~\citep{chen_meta-learning_2019} bridges this gap by targeting generalisation across both tasks and preferences.~\citet{liu_prediction_2021} extend Meta-MORL with a predictor that selects weights to maximise expected PF improvement in continuous control, but they meta-train on a single task decomposed by weight vectors rather than across distinct tasks.~\citet{lu_meta-learning_2024} detect environment changes via an unsupervised auto-encoder and treat them as contexts for a Reptile-based meta-learner~\citep{nichol_first-order_2018} with generalised policy improvement.~\citet{10439641} apply Meta-MORL to vehicular networks by decomposing into subproblems, training per-subproblem, and using MAML to learn a transferable initialisation.

Although the application of Meta-MORL in the SC domain appears to have great potential, there is a scarcity of research concentrating on this area. For example,~\citet{wang_dynamic_2024} applied MAML along with proximal policy optimisation (PPO) to address a single-facility inventory issue, while~\citet{zhang_meta-learning-based_2023} utilised Reptile with actor-critic reinforcement learning (RL) to solve travelling salesman and vehicle routing problems. Consequently, the opportunities for exploration in this domain remain substantial, particularly in complex multi-echelon structures like SC network design.

\section{Problem Definition}
We formulate the combinatorial SC optimisation problem as a finite-horizon MOMDP~\citep{hayes_practical_2022}, where each episode represents a sequence of operational decisions over time $t$ within time horizon $T$. The tuple $\langle \mathcal{S}, \mathcal{A}, ST, \gamma, \mu, \mathbf{R} \rangle$, where: $\mathcal{S}$ is the state space, including inventory levels, outstanding orders, cumulative emissions, and average service level (SL) inequality; $\mathcal{A}$ is the action space, including manufacturing and delivery quantities; $ST: \mathcal{S} \times \mathcal{A} \rightarrow \mathcal{S}$ is the state transition function that governs how states evolve given actions; $\gamma \in [0,1]$ is the discount factor; $\mu$ is the initial state distribution; and $\mathbf{R}: \mathcal{S} \times \mathcal{A} \rightarrow \mathbb{R}^d$ is a vector-valued reward function that captures conflicting objectives $d$, such as profit, emissions, and SL inequality. We denote the instantaneous vector reward on subproblem $k$ by $\mathbf{r}_k(\mathbf{s}_t, \mathbf{a}_t)=\mathbf{R}_k(\mathbf{s}_t, \mathbf{a}_t)$, where $\mathbf{s}_t$ and $\mathbf{a}_t$ represent the state and action at period $t$, respectively. The agent's goal is to learn a policy $\pi_{\theta}(\mathbf{a} \mid \mathbf{s})$ that approximates the PF over the expected cumulative rewards. The SC-specific MOMDP formulation simulated in this paper is based on the previous study~\citep{rachman_reinforcement_2026}, provided in Appendix B.

To enable generalisation across a distribution of SC tasks, we adopt a gradient-based meta-learning using RL framework that learns an initial policy parameterisation $\theta$ capable of rapid adaptation. Each task $\mathcal{T}$ is sampled from a task distribution $p(\mathcal{T})$, defined as a finite-horizon MOMDP with transition dynamics $ST_\mathcal{T}$, initial state distribution $\mu_\mathcal{T}$, and reward function $\mathbf{R}_\mathcal{T}$. Following~\citet{finn_model-agnostic_2017}, each task is associated with a loss function:
\begin{equation}
\label{eq:loss_function}
\mathcal{L}_{\mathcal{T}_k}(\pi_{\theta'_k}) = -\mathbb{E}_{\mathbf{s}_t, \mathbf{a}_t \sim \pi_{\theta'_k}, \mu_{\mathcal{T}_k}} \left[ \sum_{t=1}^{T} \mathbf{w}_k^\top \mathbf{r}_k(\mathbf{s}_t, \mathbf{a}_t) \right],
\end{equation}
where $\mathbf{w}_k \in \mathbb{R}^d$ is a task-specific weight vector used to scalarise multiple objectives.

During the adaptation phase, the meta-policy $\pi_\theta$ is fine-tuned to each sampled task $\mathcal{T}_k$ via gradient descent for $\alpha$ adaptation step size, yielding adapted parameters $\theta'_k$, where:
\begin{equation}
\label{eq:adaptation_policy}
    \theta'_k = \theta - \alpha \nabla_\theta \mathcal{L}_{\mathcal{T}_k}(\pi_\theta).
\end{equation}

In the meta-update phase, the meta-parameters $\theta$ are updated by aggregating losses across tasks drawn from the distribution $p(\mathcal{T})$, as follows:
\begin{equation}
\label{eq:meta_update}
    \theta = \arg\min_{\theta} \sum_{\mathcal{T}_k \sim p(\mathcal{T})} \mathcal{L}_{\mathcal{T}_k}(\pi_{\theta'_k}).
\end{equation}

This two-stage process enables the agent to quickly adapt to new SC tasks using limited interaction.

\section{Methods} \label{sec:methods}
This section presents the mechanism of the arrangement of the SC tasks, the algorithm we use in this study, and the performance measurement.

\subsection{Meta-MORL Algorithm}
~\citet{chen_meta-learning_2019} extends MAML to multi-objective RL by learning a meta-policy that rapidly adapts to different scalarisation preferences. Training has three stages: (i) \emph{adaptation}, scalarising each sampled task with a weight vector and performing a few inner updates; (ii) \emph{meta-learning}, updating meta-parameters across scalarised tasks to learn a transferable initialisation; and (iii) \emph{fine-tuning}, applying multiple weights on an unseen task to train candidate policies that are non-dominated sorted to approximate the PF. Although this supports generalisation across tasks and preferences, Meta-MORL samples tasks and weights independently each iteration, which can increase update variance when successive batches differ in dynamics or preferences. In high-dimensional combinatorial SC problems, this variability can weaken limited-step adaptation, motivating more structured within-task training and explicit diversity mechanisms.

\subsection{Proposed Algorithm}
We introduce MIRACL, an algorithm to promote solution diversity of current Meta-MORL by optimising through nested scalarised tasks. This method minimises variability across trained tasks in the adaptation phase by processing a set of similar subproblems with varied weights before transitioning to a different task. In addition, it incorporates a mechanism to explicitly improve task diversity for better learning in a more exploited search area. MIRACL builds upon MAML and Meta-MORL, both of which are established as general-purpose meta-learning frameworks applicable across diverse domains. Its hierarchical learning and PSA-based adaptation operate purely on objective structures, not domain-specific assumptions.

\subsubsection{Hierarchical composite learning}

\begin{figure}[t]
    \centering
    \includegraphics[width=\linewidth, trim={1cm 0cm 0cm 0cm}]{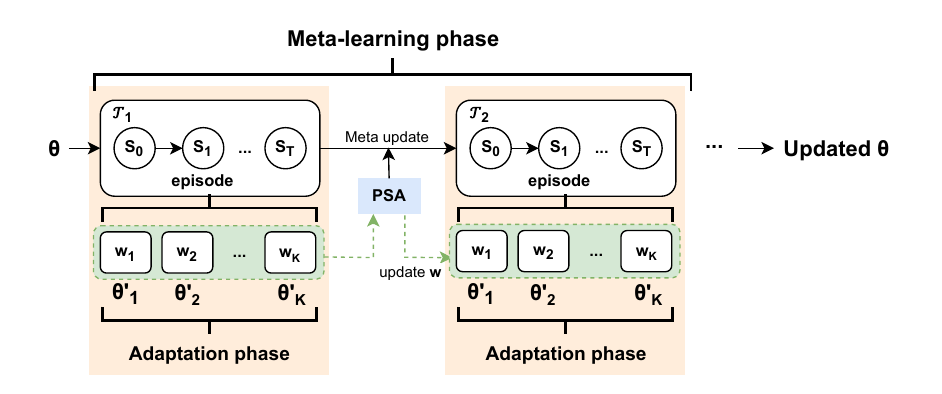}
    \caption{Meta-learning and adaptation phases in MIRACL. Each task is decomposed using several weight vectors, then PSA is applied between meta-updates to improve the task diversity.}
    \label{fig:meta-learning}
\end{figure}

As in Meta-MORL, we assume a task distribution $p(\mathcal{T})$, a weight distribution $p(\mathbf{w})$, and step sizes $\alpha$ and $\beta$ for inner adaptation and meta-updates. MIRACL introduces \emph{hierarchical composite learning}: for each sampled task, we decompose the same SC instance into $K$ scalarised subproblems by applying different weight vectors on the simplex (Figure~\ref{fig:meta-learning}). We use linear scalarisation by default, but MIRACL is agnostic to the scalarisation form and can be combined with alternatives (e.g., Tchebycheff~\citep{van_moffaert_scalarized_2013}). Unlike Meta-MORL, which treats sampled weights as fixed, MIRACL updates the $K$ weights after each meta-update using an \emph{archive-guided PSA} rule, steering subproblems toward under-covered regions of objective space without additional rollouts.

Processing multiple subproblems under the same task dynamics provides a more stable adaptation signal and enables the diversity mechanism. We initialise the meta-policy $\pi_\theta$ and maintain an archive of non-dominated vector returns across tasks and subproblems to guide PSA throughout meta-training. Standard Meta-MORL samples independent pairs $(\mathcal{T},\mathbf{w})$, entangling task and preference variation within a batch. MIRACL instead conditions on a task $\mathcal{T}$ and evaluates $\{\mathbf{w}_k\}_{k=1}^{K}$ under shared dynamics, while PSA perturbs weights using the archive to discourage revisiting previously explored objective regions and to improve coverage.

%additional theoretical analysis

\paragraph{Variance reduction intuition.} 
Let $G(\mathcal{T},\mathbf{w})$ denote the meta-gradient contribution obtained after inner adaptation
under task $\mathcal{T}$ and preference weight $\mathbf{w}$:
$G(\mathcal{T},\mathbf{w}):= \nabla_{\theta} \mathcal{L}_{\mathcal{T}}(\pi_{\theta'})$, where $\theta'$ is produced by Equation~\eqref{eq:adaptation_policy} and $\mathcal{L}_\mathcal{T}$ is the scalarised loss using weight $\mathbf{w}$ (as in Equation~\eqref{eq:loss_function}). We denote the meta-gradient estimator at a given iteration as $\widehat{G}_{\mathrm{Meta}} = \frac{1}{B}\sum_{i=1}^B G(\mathcal{T}_i, \mathbf{w}_i)$ for Meta-MORL, and $\widehat{G}_{\mathrm{MIRACL}} = \frac{1}{K}\sum_{k=1}^K G(\mathcal{T}, \mathbf{w}_k)$ for MIRACL (where MIRACL uses a single task $\mathcal{T}$ with $K$ weights). By the law of total variance~\citep{casella_statistical_2002}, Meta-MORL satisfies:
\begin{equation}
\begin{split}
\label{eq:var_meta_morl}
\mathrm{Var}(\widehat{G}_{\mathrm{Meta}}) =&
\underbrace{\mathbb{E}_{\{\mathcal{T}_i\}}\!\left[
\mathrm{Var}\!\left(\frac{1}{B}\sum_{i=1}^{B} G(\mathcal{T}_i,\mathbf{w}_i)\,\Big|\,\{\mathcal{T}_i\}\right)\right]}_{\text{preference-induced variance across heterogeneous tasks}}
+\underbrace{\mathrm{Var}_{\{\mathcal{T}_i\}}\!\left(\mathbb{E}_{\{\mathbf{w}_i\}}\!\left[\frac{1}{B}\sum_{i=1}^{B} G(\mathcal{T}_i,\mathbf{w}_i)\,\Big|\,\{\mathcal{T}_i\}\right]\right)}_{\text{task-induced variance}},
\end{split}
\end{equation}

\noindent where $\{\mathcal{T}_i\}_{i=1}^B$ is a batch of $B$ tasks sampled at a given meta-iteration. Conditioning on $\{\mathcal{T}_i\}$ isolates the variance due to weight sampling for this fixed task set.

In contrast, MIRACL conditions on a \emph{single} task $\mathcal{T}$ within the meta-iteration. Since all $K$ subproblems share the same transition dynamics $ST_{\mathcal{T}}$, their gradient contributions $\{G(\mathcal{T}, \mathbf{w}_k)\}_{k=1}^K$ are correlated through the task structure. The within-iteration (preference-induced) variance thus admits the covariance expansion:
\begin{equation}
\begin{split}
\label{eq:var_miracl_conditional}
\mathrm{Var}|&\!\left(\widehat{G}_{\mathrm{MIRACL}}\,\Big|\,\mathcal{T}\right)=
\frac{1}{K^2}\sum_{k=1}^{K}\mathrm{Var}\!\left(G(\mathcal{T},\mathbf{w}_k)\mid \mathcal{T}\right)+\frac{2}{K^2}\sum_{1\le k<\ell\le K}\mathrm{Cov}\!\left(G(\mathcal{T},\mathbf{w}_k), G(\mathcal{T},\mathbf{w}_\ell)\mid \mathcal{T} \right).
\end{split}
\end{equation}

\noindent Taking expectation over $\mathcal{T}\sim p(\mathcal{T})$ yields:
\begin{equation}
\begin{split}
\label{eq:var_miracl_total}
\mathrm{Var}(\widehat{G}_{\mathrm{MIRACL}})
=&\mathbb{E}_{\mathcal{T}}\!\left[\mathrm{Var}\!\left(\widehat{G}_{\mathrm{MIRACL}}\mid \mathcal{T}\right)\right]+\mathrm{Var}_{\mathcal{T}}\!\left(\mathbb{E}_{\{\mathbf{w}_k\}}\!\left[\widehat{G}_{\mathrm{MIRACL}}\mid \mathcal{T}\right]\right).
\end{split}
\end{equation}

Meta-MORL aggregates gradients from different tasks and preferences within a meta-batch, whereas MIRACL first averages $K$ preference-conditioned gradients under fixed task dynamics. In Equation~\eqref{eq:var_miracl_conditional}, the $1/K^2$ factors come from this averaging, and the covariance term reflects cross-preference gradient alignment. When gradients are not nearly identical (i.e., not perfectly aligned), the covariance is not dominant, so within-task averaging behaves like a sample mean: preference-driven variance shrinks as $1/K$ (standard deviation $1/\sqrt{K}$). Hence, MIRACL reduces preference-induced fluctuations in each task’s meta-update. This discussion aims to build intuition rather than provide a direct measurement. Empirically tracking meta-gradient variance would require checkpoint-level logging throughout meta-training, which we do not have for the runs reported. We therefore use observed experimental trends as indirect evidence and leave a thorough empirical verification to future work.

% The variance analysis offers intuition for the stability benefits of decomposing each task into $K$ weight-conditioned subproblems. A direct empirical test would require logging meta-gradient statistics across training checkpoints; since only the final meta-policy checkpoint is available for the reported runs, we do not include such trajectories here and instead position the analysis as theoretical support, complemented by the observed performance trends in our experiments.

Algorithm~\ref{algo:main_algo} instantiates this process: line~3 samples a single task $\mathcal{T}$, lines~9--14 compute $K$ adapted gradients under shared dynamics (yielding the covariance structure in Equation~\eqref{eq:var_miracl_conditional}), line~15 applies the meta-update using $\widehat{G}_{\mathrm{MIRACL}}$, and the diversity mechanism applies PSA to update weights using the archive. The meta-learner operates over tasks $\mathcal{T}\sim p(\mathcal{T})$ through an $N$-shot trajectory $D=\{(\mathbf{s}_1,\mathbf{a}_1,\dots ,\mathbf{s}_T)\}$: each $\mathcal{D}_k$ is collected by executing $\pi_\theta$ on $\mathcal{T}_k$, inner-loop adaptation yields task-specific $\theta'_k$, and the outer loop updates $\theta$ using $\mathcal{L}_{\mathcal{T}_k}$.
% Algorithm~\ref{algo:main_algo} instantiates this framework: line 3 samples a single task $\mathcal{T}$ (reducing task-induced variance within the batch), lines 9-14 compute $K$ adapted gradients under shared dynamics (yielding the covariance structure in Equation~\eqref{eq:var_miracl_conditional}), and line 15 applies the meta-update using averaged gradient $\widehat{G}_{\mathrm{MIRACL}}$. The diversity mechanism, \blue{which utilises the PSA method in hierarchical composite learning,} is applied to improve variety in the adaptation phase. Subsequently, the agent iterates across these subproblems, then updates the existing meta-policy in the meta-learning phase. The meta-learner operates across multiple tasks $\mathcal{T}$ within a distribution of tasks $p(\mathcal{T})$ through a $N$ shot learning trajectory $D$, where $D=\{(\mathbf{s}_1,\mathbf{a}_1,\dots ,\mathbf{s}_T)\}$. Each $D_k$ is obtained by applying the meta-policy $\pi_{\theta}$ with the generalised meta-parameter $\theta$ on $\mathcal{T}_k$. The meta-parameter $\theta$ is updated in the inner loop by learning the adaptation with gradient descent, resulting in a task-specific parameter $\theta'_k$. Then, $\theta'_k$ is used to generate $D'_k$ in $\mathcal{T}_k$. In the outer loop, $D'_k$ and $\mathcal{L}_{\mathcal{T}_k}$ are employed to update the meta-parameter $\theta$. At the end of the training, the algorithm produces a generalised updated meta-policy $\pi_\theta$ that will be used for the fine-tuning phase.

\begin{algorithm}
\caption{Meta-training Phase of MIRACL}
\label{algo:main_algo}
\begin{algorithmic}[1]
\REQUIRE Task distribution $p(\mathcal{T})$, weight distribution $p(\mathbf{w})$, number of subproblems $K$, step sizes $\alpha$, $\beta$
\STATE \textbf{Initialise:} meta-policy $\pi_\theta$, $\text{archive} \gets \emptyset$
\WHILE{not done}
    \STATE Sample a task $\mathcal{T} \sim p(\mathcal{T})$
    \IF{$\text{archive} = \emptyset$}
        \STATE Generate $K$ normalised weights $\{\mathbf{w}_1, \dots, \mathbf{w}_K\} \sim p(\mathbf{w})$
    \ELSE
        \STATE Generate $K$ weights $\{\mathbf{w}_1, \dots, \mathbf{w}_K\}$ from $\text{archive}$
    \ENDIF
    \FOR{$k = 1$ to $K$} %task decomposition
        \STATE Form subproblems $\mathcal{T}_k$ using $(\mathcal{T}, \mathbf{w}_k)$
        \STATE Rollout $\mathcal{D}_k$ using $\pi_\theta$ in $\mathcal{T}_k$ %adaptation phase
        \STATE Compute $\theta'_k \gets \theta - \alpha \nabla_\theta \mathcal{L}_{\mathcal{T}_k}(\pi_\theta)$
        % \STATE Rollout $\mathcal{D}'_k$ using $\pi_{\theta'_k}$ in $\mathcal{T}_k$
        % \STATE Evaluate reward $\mathbf{r}_k$ from $\mathcal{D}'_k$
        \STATE Evaluate reward $\mathbf{r}_k$ in $\mathcal{T}_k$ using $\pi_{\theta'_k}$
    \ENDFOR
    \STATE Meta-update $\theta \gets \theta - \beta \nabla_\theta \sum_k \mathcal{L}_{\mathcal{T}_k}(\pi_{\theta'_k})$ %meta-learning phase
    \STATE Update $\mathbf{w}$ via $\text{DiversityMechanism}(\mathcal{T}, \mathbf{w}, \mathbf{r}, \text{archive})$
\ENDWHILE
\RETURN Updated meta-policy $\pi_\theta$
\end{algorithmic}
\end{algorithm}

\subsubsection{Diversity mechanism}

In Meta-MORL, each task is scalarised by sampling a preference vector $\mathbf{w}\sim p(\mathbf{w})$ and adapting from a shared meta-policy $\pi_\theta$. While this enables efficient multi-objective learning and cross-task generalisation, it also couples early exploration across subproblems: within a meta-batch, trajectories are generated from the same initial policy before inner-loop adaptation. As a result, the diversity of collected experience is largely determined by the coverage of $p(\mathbf{w})$ and by the capacity of the adaptation step to separate behaviours across weights. When $p(\mathbf{w})$ under-explores the simplex or adaptation is shallow, the meta-policy may concentrate on a narrow subset of trade-offs and generalise poorly to unseen preferences.

We address this limitation by introducing a \emph{diversity mechanism} that actively perturbs the $K$ subproblem weights after each meta-update. Specifically, we employ Pareto Simulated Annealing (PSA) \citep{czyzzak_pareto_1998} to update $\{\mathbf{w}_k\}_{k=1}^K$ using the archive of previously evaluated objective vectors. After normalising objectives to $[0,1]$, we find for each $\mathbf{r}_k$ its nearest archived neighbour $\mathbf{r}'_k$ (Euclidean distance) and adjust each weight component $w_k^j$ according to the PSA rule in Equation~\eqref{eq:psa}. Intuitively, the archive acts as a reference set, and PSA encourages successive subproblems to move away from existing solutions, promoting broader coverage of the PF set.

\begin{equation}
\label{eq:psa}
    w_k^j =
    \begin{cases}
        w_k^j \times (1+\delta), & \textbf{if } r_k^j \geq r_k^{\prime j},\\[4pt]
        w_k^j / (1+\delta), & \textbf{if } r_k^j < r_k^{\prime j}.
    \end{cases}
\end{equation}
\noindent The update is applied component-wise for $j=1,\dots,d$ (with $d$ objectives). For minimisation objectives (e.g., emissions, inequality), we sign-flip rewards so that larger values always indicate better performance before comparison.
% \begin{equation}
% \label{eq:psa}
%     w_k^j =
%     \begin{cases}
%         w_k^j \times (1+\delta) \quad \textbf{if } r_k^j \geq r^{'j}_k, \quad \forall{j}\\
%         w_k^j / (1+\delta) \quad \textbf{if } r_k^j< r^{'j}_k, \quad \forall{j}.
%     \end{cases}
% \end{equation}

\begin{algorithm}[h]
\caption{Diversity Mechanism (with PSA)}
\label{alg:diversity_mechanism}
\begin{algorithmic}[1]
\REQUIRE Task $\mathcal{T}$, weights $\{\mathbf{w}_1, \dots, \mathbf{w}_K\}$, rewards $\{\mathbf{r}_1, \dots, \mathbf{r}_K\}$, archive, PSA steps $S$, PSA rate $\delta$
\FOR{$s = 1$ to $S$}
    \FOR{$k = 1$ to $K$}
        \IF{archive $\neq \emptyset$}
            \STATE Find closest reward $\mathbf{r}'$ in archive to $\mathbf{r}_k$
            \STATE Update $\mathbf{w}_k$ using PSA rule (Equation~\eqref{eq:psa})
            \STATE Normalise $\mathbf{w}_k \leftarrow \mathbf{w}_k / \|\mathbf{w}_k\|_1$, with $\mathbf{w}_k \geq 0$
        \ENDIF
    \ENDFOR
\ENDFOR
\FOR{$k = 1$ to $K$}
    \IF{$\mathbf{r}_k$ is non-dominated w.r.t. archive}
        \STATE Add $\mathbf{r}_k$ to archive
    \ENDIF
\ENDFOR
\RETURN{Updated $\mathbf{w}$, archive}
\end{algorithmic}
\end{algorithm}

% Algorithm~\ref{alg:diversity_mechanism} elucidates how this concept is applied to our proposed algorithm. Tasks $\mathcal{T}$, weights $\mathbf{w}$, rewards $\mathbf{r}$, and archive are rendered from the task decomposition process in the main algorithm. In addition, the PSA steps $S$ and the adjustment factor $\delta$ are also defined in advance. The agent iterates across generated rewards from all weights and applies the PSA technique to each of them based on the archive. All updated non-dominated rewards from each task are used to update the archive.

Algorithm~\ref{alg:diversity_mechanism} applies PSA after task decomposition: given $\mathcal{T}$, $\{\mathbf{w}_k\}_{k=1}^K$, $\{\mathbf{r}_k\}_{k=1}^K$, and the archive (with fixed $S$ and $\delta$), it updates each $\mathbf{w}_k$ using its nearest archived neighbour, then renormalises onto the simplex. Finally, newly obtained non-dominated $\mathbf{r}_k$ are added to the archive.

\subsubsection{Fine-tuning phase}

After meta-learning, the trained meta-policy initialises fine-tuning, where the agent adapts to an unseen task in a few gradient steps. To approximate the PF, we train a set of scalarised policies using randomly sampled weights over the preference simplex, which avoids concentrating training on a small set of fixed trade-offs and remains practical as the number of objectives grows. Each sampled weight defines one scalarised subproblem, and we train each corresponding solution separately with the chosen RL algorithm, for example, PPO.

\begin{figure}[t]
    \centering
    \includegraphics[width=\linewidth, trim={1cm 0cm 0cm 0cm}]{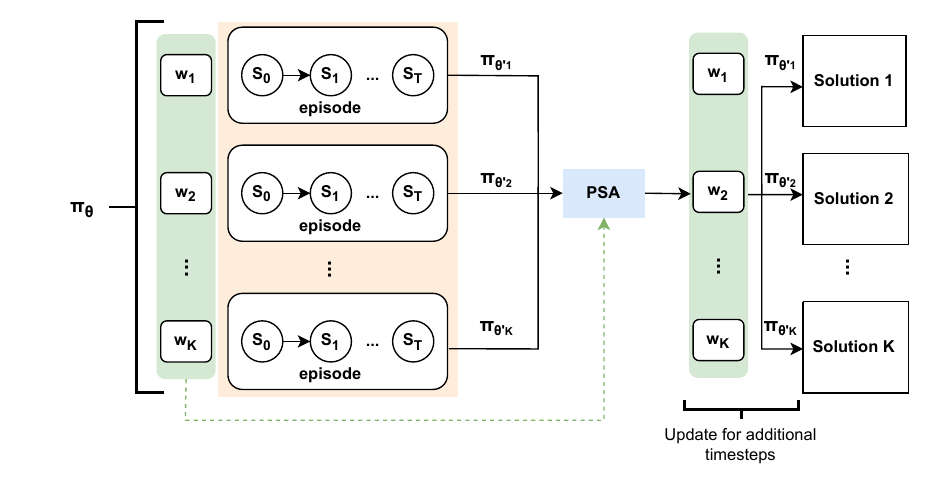}
    \caption{Fine-tuning phase requires only a few shots training since it utilises the meta-policy as a good initialisation.}
    \label{fig:fine-tuning}
\end{figure}

The PSA-based weight diversity mechanism is applied only at the end of fine-tuning rather than throughout training. This approach aligns with meta-learning principles. Since all policies start from the same meta-policy, early PSA application would be ineffective when policies remain similar. Delaying PSA until after fine-tuning allows policies to first specialise toward their assigned weights before targeted refinement, making diversity enhancement more effective. The detailed pseudocode is provided in Appendix C.

\section{Experiments} \label{sec:experiment}
This section outlines the settings for the meta-training and fine-tuning phases in our experiments. We conducted our experiments using Python 3.11.2 on JupyterLab with 128 GB RAM and a 16-core CPU. We train the algorithms on three complexities of the SC environment shown in Table~\ref{tab:sc_environmnet}. Each instantiation is repeated ten times for statistical robustness. SC networks are configured as collections of nodes that are interconnected by edges. Additional SC configuration and environment details are in Appendix D.

\begin{table}[h]
    \centering
    \caption{SC environments simulated in our experiment. Each environment complexity corresponds with the action and observation space dimension, number of nodes, and edges.}
    \label{tab:sc_environmnet}
    \begin{tabularx}{10cm}{lllll}
        \toprule
        Complexities & Action & Observation & Nodes & Edges\\
        \midrule
        Simple & 8 & 20 & 7 & 8\\
        Moderate & 21 & 49 & 13 & 21\\
        Complex & 59 & 131 & 24 & 59\\
        \bottomrule
    \end{tabularx}
\end{table}

\begin{figure*}[t]
    \centering
    \begin{subfigure}[t]{0.32\textwidth}
        \centering
        \includegraphics[width=\linewidth]{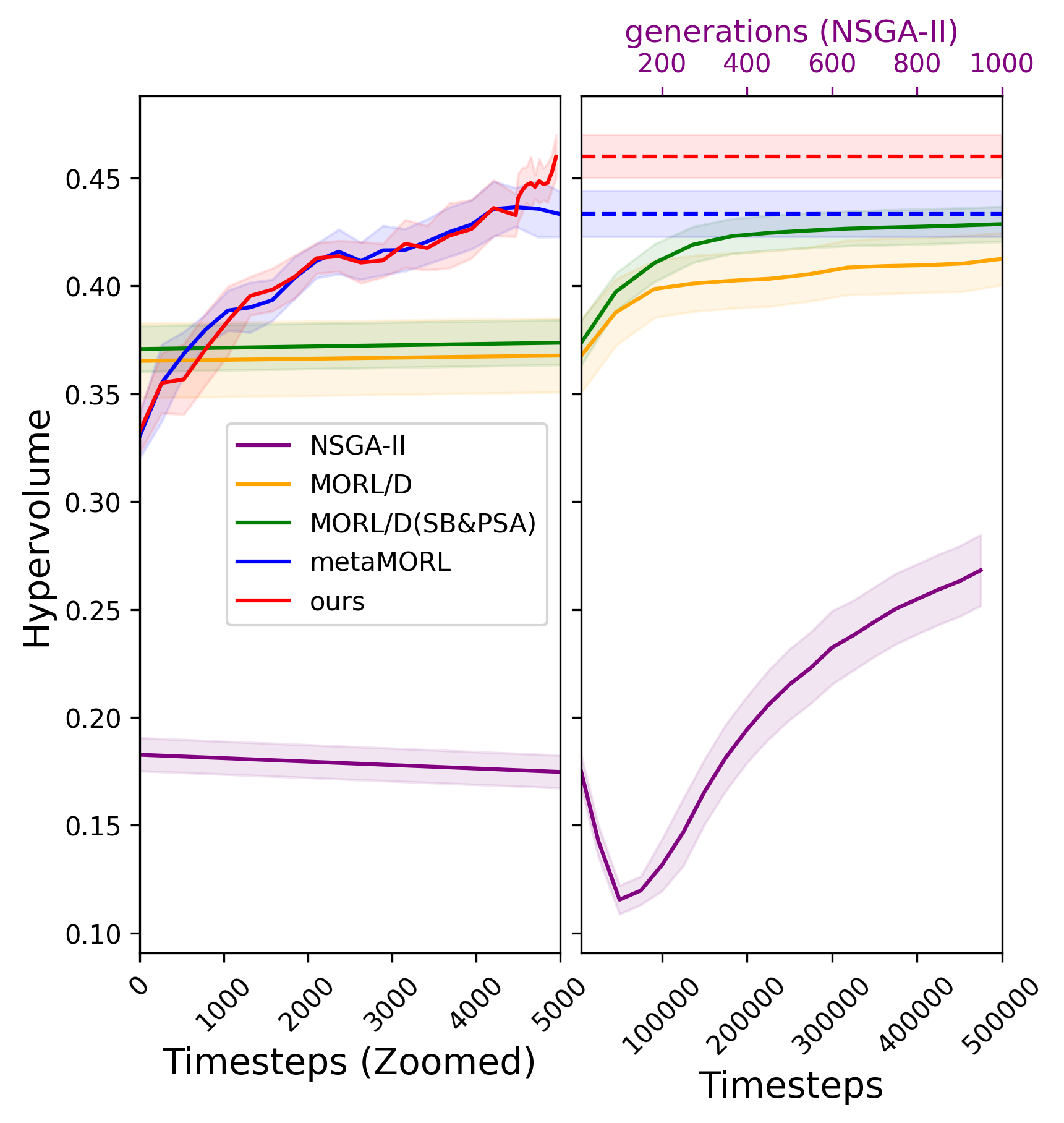}
        \caption{Simple}
        \label{fig:hv_simple}
    \end{subfigure}
    % \hspace{0.04\textwidth}
    \begin{subfigure}[t]{0.32\textwidth}
        \centering
        \includegraphics[width=\linewidth]{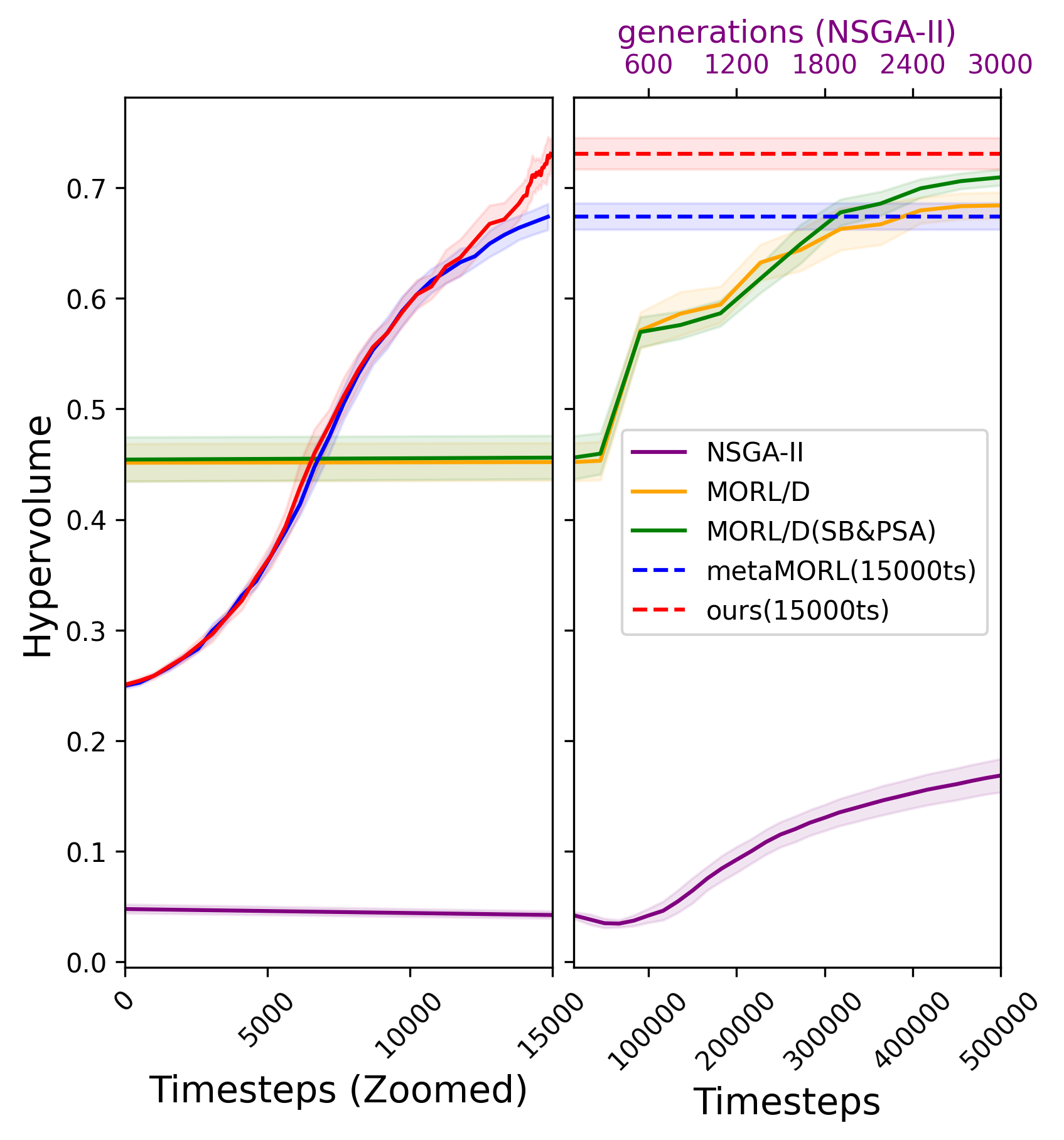}
        \caption{Moderate}
        \label{fig:hv_moderate}
    \end{subfigure}
    % \hspace{0.04\textwidth}
    \begin{subfigure}[t]{0.32\textwidth}
        \centering
        \includegraphics[width=\linewidth]{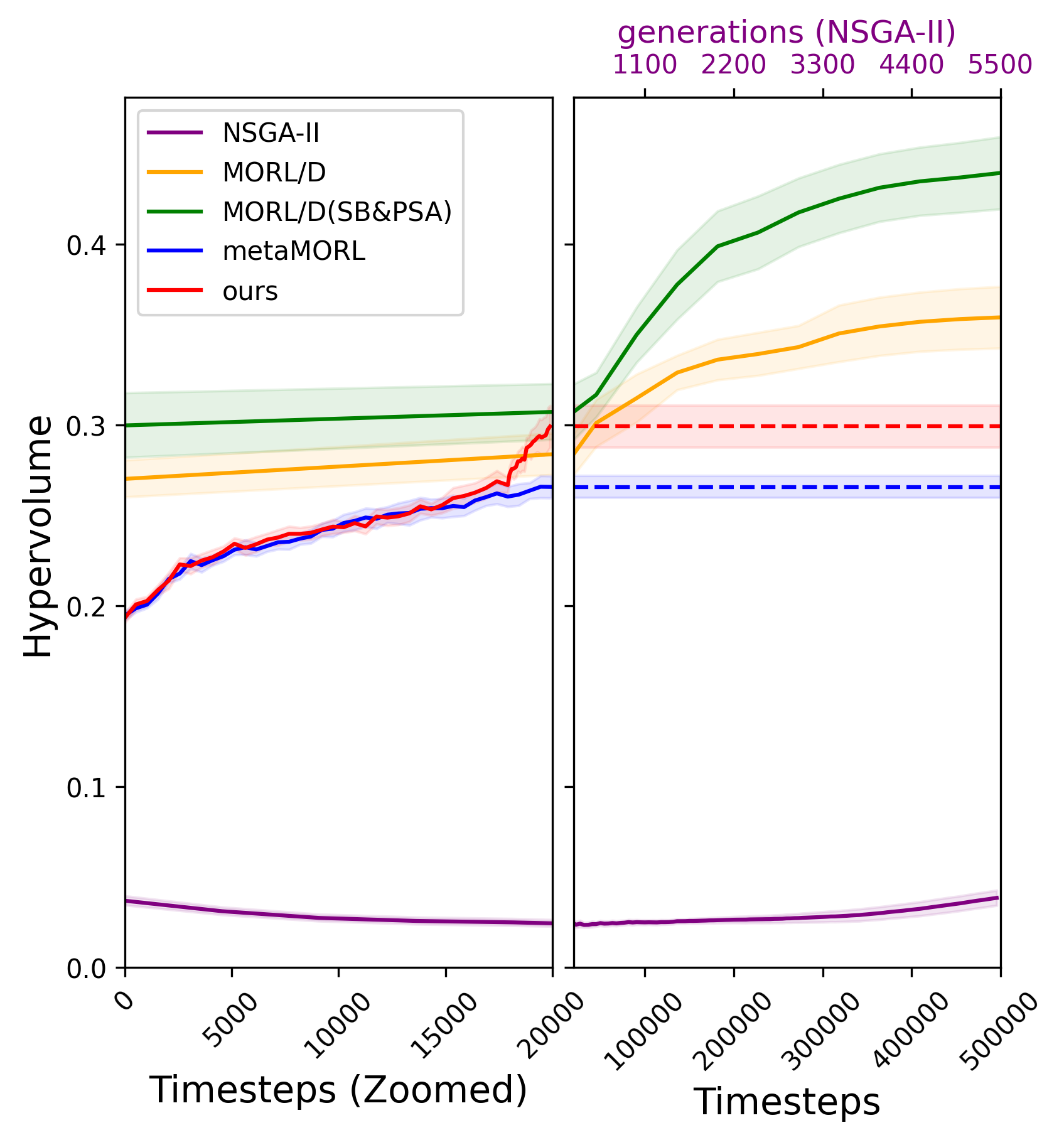}
        \caption{Complex}
        \label{fig:hv_complex}
    \end{subfigure}

    \caption{Normalised hypervolume comparison of MORL/D, MORL/D with SB and PSA, NSGA-II, Meta-MORL, and our proposed methods. Left panels show early time steps (proportional to NSGA-II generations), while right panels show later iterations. MIRACL consistently outperforms all baseline methods in simple (\ref{fig:hv_simple}) and moderate problems (\ref{fig:hv_moderate}), but is exceeded by MORL/D in complex ones (\ref{fig:hv_complex}).}
    \label{fig:comp_rl}
\end{figure*}

\begin{table}[t]
    \centering
    \caption{Runtime comparison (in minutes) between MIRACL and traditional algorithms. Meta-training time (MT) is reported separately as a one-off overhead (using 4 parallel workers), while fine-tuning time (FT) reflects the per-task cost.}
    \label{tab:running_time}
    \begin{tabular}{lrrr}
    \toprule
    Algorithms & Simple & Moderate & Complex \\
    \midrule
    MIRACL(MT) & 8 & 12 & 14 \\
    MIRACL(FT) & 14 & 47 & 77 \\
    MORL/D & 84 & 95 & 111 \\
    NSGA-II & 2 & 10 & 47 \\
    \bottomrule
    \end{tabular}
\end{table}

\subsection{Algorithm Settings}
% \hl{extend a bit to fill in the pages}
% The algorithm settings are divided into meta-training and fine-tuning phases. We use PPO as the RL backbone and report results over multiple seeds. Unless stated otherwise, weights are sampled from Dirichlet$(1,1,1)$, and we solve $K=10$ scalarised subproblems per task. Meta-learned methods are evaluated in a few-shot fine-tuning regime on unseen tasks, while non-meta baselines train from scratch under the same evaluation protocol. Full meta-training, fine-tuning hyperparameters and all detailed algorithm settings are provided in Appendix~\ref{sec:detailed_algo_setting}.

% \section{Detailed Algorithm Settings} \label{sec:detailed_algo_setting}
We report meta-training and fine-tuning settings for MIRACL and baselines, with detailed hyperparameters provided in Appendix E. Minimisation objectives (emissions and inequality) are sign-flipped, and all objectives are normalised to $[0,1]$ prior to scalarisation and nearest-neighbour comparisons.

\subsection{Meta-training settings}
We implement MAML~\citep{finn_model-agnostic_2017} with PPO~\citep{schulman_proximal_2017} in Ray RLlib 2.3.1~\citep{liang_rllib_2017}, extended to support Meta-MORL and MIRACL. Meta-training runs for $10^6$ environment steps on randomly generated SC tasks. In each meta-iteration, a task is sampled and decomposed into subproblems. Then, inner-loop adaptation is performed, and a meta-update is applied to the shared policy. We use four inner adaptation steps for simple problems and eight for moderate and complex cases, balancing fast adaptation and task-specific learning. A meta-environment wrapper enables dynamic task sampling to reflect realistic task variation.

To promote diversity, MIRACL applies 10 PSA steps after each meta-update. We sample weights $\mathbf{w}$ uniformly from the 3D simplex using Dirichlet$(1,1,1)$ and solve $K=10$ scalarised subproblems per meta-iteration with a fixed PSA rate $\delta=0.05$. We use linear scalarisation for its simplicity and smooth optimisation signal when training many subproblems in high-dimensional combinatorial settings. Noting that it may under-represent non-convex PF regions, we therefore include a Linear vs.~Tchebycheff ablation under the same budget and protocol. MIRACL maintains a global archive of non-dominated (normalised) reward vectors across tasks, which guides PSA weight updates. Training stability is improved via PPO clipping, Kullback--Leibler regularisation, and generalised advantage estimation.

\subsection{Fine-tuning settings}
Fine-tuning uses the Messiah SC simulator~\citep{rachman_reinforcement_2026} for 5{,}000, 15{,}000, and 20{,}000 steps on simple, moderate, and complex tasks, respectively, which is substantially fewer than training-from-scratch RL baselines (typically $\sim$500{,}000 steps in our SC setup). The number of shots matches meta-training, and observation/action spaces are normalised.

As PPO is single-objective, we scalarise vector rewards with simplex-projected weights. For each task, we train 21 weighted agents ($K=21$) across 10 seeds (210 runs) and obtain 10 PF approximation sets via non-dominated sorting~\citep{singh_multi-objective_2025}. We use generalised advantage estimation with advantage normalisation to reduce variance, and a standard multi-layer perceptron policy architecture. Experiments are conducted on three SC environments of increasing network complexity (simple, moderate, complex).

\begin{table*}[t]
    \centering
    \caption{Performance comparison between Meta-MORL and MIRACL. MIRACL is evaluated with PSA applied only in meta-training (MT) or in both meta-training and fine-tuning (MT\&FT). We report hypervolume, sparsity, and EUM, with $\Delta$ indicating the percentage change relative to Meta-MORL. Statistical significance is assessed via a Kruskal--Wallis test followed by Dunn's post hoc tests with Bonferroni correction, where *, **, and *** denote $p<0.05$, $p<0.01$, and $p<0.001$, respectively.}
    \label{tab:ablation}
    \begin{tabularx}{\linewidth}{lllllll}
    \toprule
    Problems & Aspects & Meta-MORL & MT & MT\&FT & $\Delta$ MT & $\Delta$ MT\&FT\\
    \midrule
    \multirow{3}{*}{Simple} & Hypervolume & 0.4333 & 0.4460 & 0.4600 & \textbf{2.92\%} & \textbf{6.16\%}** \\
    & Sparsity & 0.0056 & 0.0052 & 0.0057 & -8.03\% & \textbf{1.36}\% \\\vspace{+2mm}
    & EUM & 0.7075 & 0.7398 & 0.7436 & \textbf{4.57\%} & \textbf{5.10\%}** \\
    \multirow{3}{*}{Moderate} & Hypervolume & 0.6737 & 0.6853 & 0.7306 & \textbf{1.73\%} & \textbf{8.45\%}*** \\
    & Sparsity & 0.0015 & 0.0025 & 0.0013 & \textbf{71.75}\%* & -7.78\% \\\vspace{+2mm}
    & EUM & 0.8387 & 0.8538 & 0.8394 & \textbf{2.75\%}* & \textbf{1.03\%} \\
    \multirow{3}{*}{Complex} & Hypervolume & 0.2658 & 0.2668 & 0.2993 & \textbf{0.38\%} & \textbf{12.62\%}*** \\
    & Sparsity & 0.0022 & 0.0016 & 0.0032 & -25.77\% & \textbf{44.78\%}* \\
    & EUM & 0.6118 & 0.6176 & 0.6236 & -0.35\% & \textbf{1.79\%} \\
    \bottomrule
    \end{tabularx}%
\end{table*}

\subsection{Results and Discussion}

We evaluate MIRACL on multiple SC settings, comparing its fine-tuning performance with Meta-MORL and standard MORL baselines~\citep{rachman_reinforcement_2026}: MORL/D, its shared-buffer (SB) variant that shares trajectories across policies, and the classical NSGA-II. All methods are tested on similar unseen tasks. Performance is measured by normalised hypervolume~\citep{goos_multiobjective_1998} with reference point (0,0,0) after sign-flipping minimisation objectives; the expected utility metric (EUM)~\citep{hayes_practical_2022},
$\text{EUM} = \frac{1}{|\mathcal{W}|}\sum_{\mathbf{w}\in\mathcal{W}} \max_{\pi\in\Pi} \mathbf{w}^\top \mathbf{r}(\pi)$,
using weight vectors $\mathbf{w}$ uniformly sampled from the 3-dimensional simplex; and sparsity, the average squared distance between consecutive Pareto solutions used as a diversity indicator~\citep{pmlr-v119-xu20h}. Higher hypervolume and EUM indicate better performance, while lower sparsity corresponds to a denser solution set.

\subsubsection{Traditional method comparison}
% \blue{Figure~\ref{fig:comp_rl} shows that MIRACL achieves about 10\% and 6\% higher hypervolume than baselines in simple and moderate tasks, respectively, with strong statistical significance. In complex settings, MIRACL lags behind MORL/D by roughly 20\% but still converges faster and generalises better than NSGA-II. This gap reflects the difficulty of few-shot adaptation in highly complex environments rather than a fundamental limitation, and can be mitigated through longer fine-tuning or repeated PSA at higher computational cost. NSGA-II consistently performs worst, especially in complex cases. Overall, MIRACL performs strongly in simple and moderate tasks and remains competitive in harder settings despite using far fewer time steps.}

Figure~\ref{fig:comp_rl} shows that MIRACL achieves about 10\% and 6\% higher normalised hypervolume than the baselines on the simple and moderate tasks, respectively, while using substantially fewer fine-tuning time steps. In the complex task, MIRACL is roughly 20\% below MORL/D but converges faster and generalises better than NSGA-II, which performs worst overall. Across 10 runs per method per setting, Kruskal--Wallis detects significant differences ($p<0.001$). Dunn's post hoc tests with Bonferroni correction confirm MIRACL outperforms NSGA-II in all settings ($p<0.001$), and also MORL/D in the simple task and Meta-MORL in the moderate task ($p<0.05$). Where differences are significant, MIRACL attains the higher hypervolume. In the complex task, the deficit to MORL/D is not statistically significant, suggesting competitiveness under increased complexity. The remaining gap likely stems from the few-shot adaptation difficulty in highly complex environments. It can be reduced through longer fine-tuning or repeated PSA to allow the solutions to spread more, covering a larger hypervolume area, at a higher computational cost.

% \blue{Table~\ref{tab:running_time}} reports wall-clock time comparisons with the baselines (Meta-MORL is omitted as it is similar to MIRACL). MIRACL pays a one-off meta-training (with 4 parallel rollout workers) cost over a task distribution, then reuses the meta-policy to initialise per-task fine-tuning. Even including this upfront overhead, MIRACL is substantially faster than MORL/D across all problem complexities, with runtime increasing naturally with problem size. NSGA-II is the least expensive but delivers weaker performance on the more complex problems, indicating that MIRACL provides a favourable upfront, per-task runtime trade-off.

% \blue{Table~\ref{tab:running_time}} reports wall-clock runtimes (Meta-MORL omitted as it approximates MIRACL) on the computation resource mentioned in section~\ref{sec:experiment}. MIRACL pays a one-off meta-training cost with 4 parallel rollout workers (aggregate steps) and reuses the meta-policy for per-task fine-tuning. Even so, it is substantially faster than MORL/D across all complexities. NSGA-II is cheapest but degrades on harder problems, suggesting MIRACL offers a favourable upfront vs.~per-task runtime trade-off. We emphasise that meta-learning targets \emph{per-task} sample efficiency: MIRACL is evaluated in a few-shot fine-tuning regime by design, whereas MORL/D and NSGA-II require training from scratch.

Table~\ref{tab:running_time} reports wall-clock runtimes on the resource in Section~\ref{sec:experiment} (Meta-MORL omitted as its runtime is comparable to MIRACL). MIRACL incurs a one-off meta-training cost with four parallel rollout workers, then reuses the meta-policy for fast per-task fine-tuning. Despite this, it is consistently faster than MORL/D across all complexities. NSGA-II is the cheapest but degrades on harder instances, highlighting MIRACL's favourable upfront--per-task trade-off. Meta-learning targets \emph{per-task} sample efficiency: MIRACL is evaluated in a few-shot fine-tuning regime, whereas MORL/D and NSGA-II train from scratch.

\begin{figure*}[t]
    \centering
    \begin{subfigure}[t]{0.30\textwidth}
        \centering
        \includegraphics[width=\linewidth]{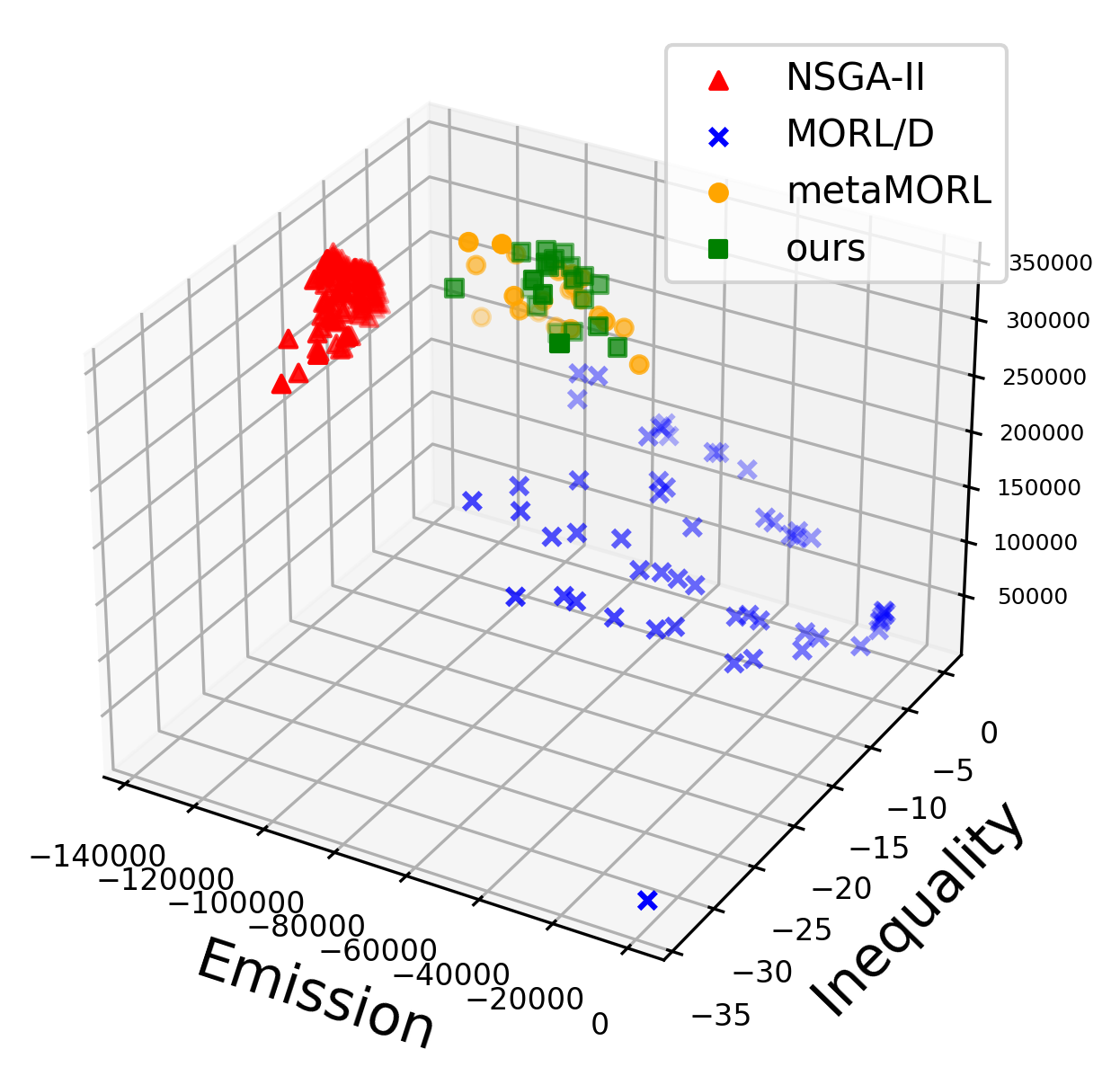}
        \caption{Simple}
    \end{subfigure}
    \begin{subfigure}[t]{0.30\textwidth}
        \centering
        \includegraphics[width=\linewidth]{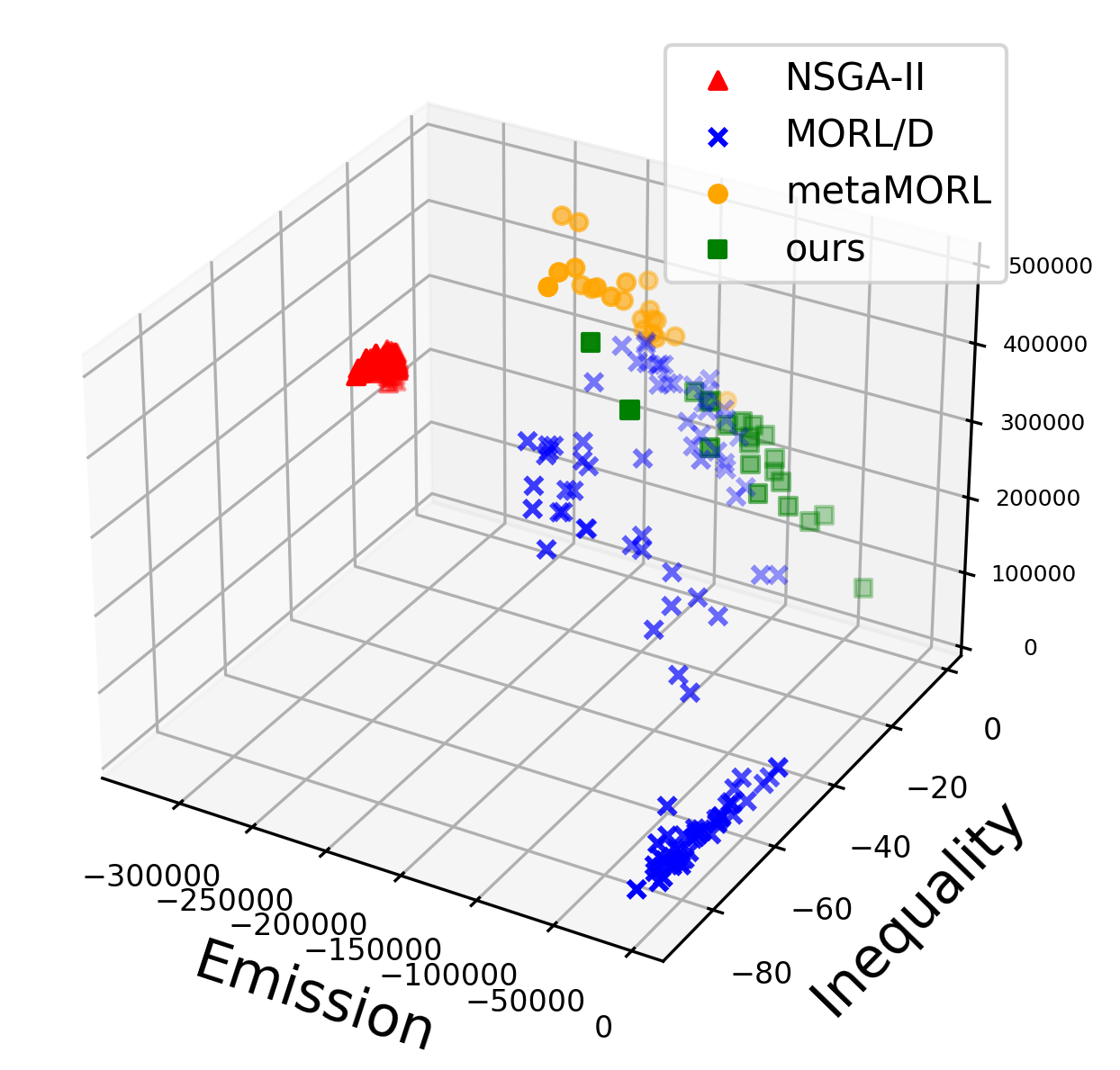}
        \caption{Moderate}
    \end{subfigure}
    \begin{subfigure}[t]{0.30\textwidth}
        \centering
        \includegraphics[width=\linewidth]{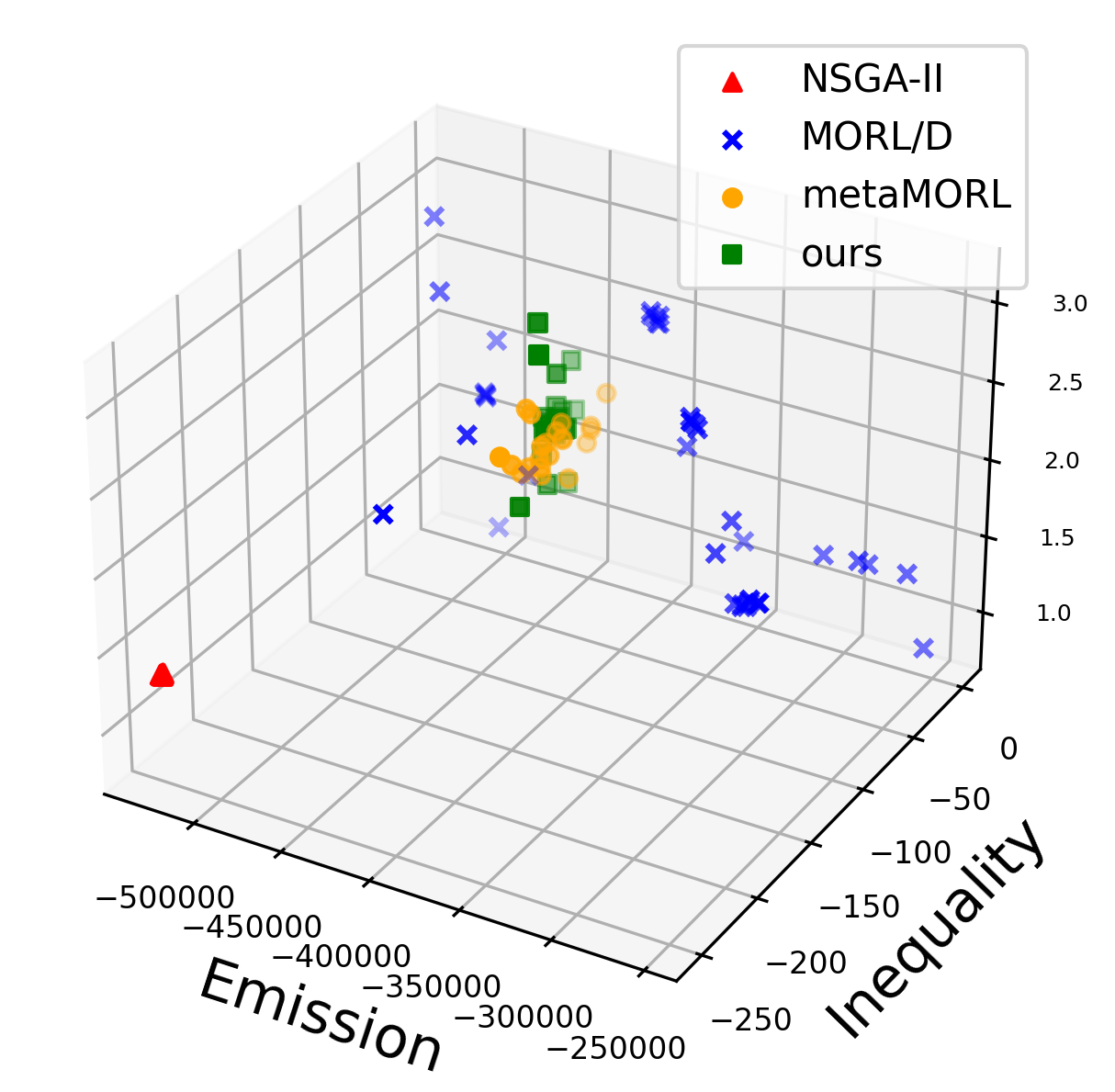}
        \caption{Complex}
    \end{subfigure}

    \caption{PF approximation sets of NSGA-II, MORL/D, Meta-MORL, and MIRACL across simple, moderate, and complex SC problems. Meta-learning-based methods produce more diverse solutions than NSGA-II, yet, more concentrated than MORL/D. The concentration increases with the problem complexity.}
    \label{fig:comp_pf}
\end{figure*}

\subsubsection{Ablation studies}

We conduct some ablations to validate MIRACL's design choices and assess component contribution. Table~\ref{tab:ablation} focuses on the diversity mechanism, showing that PSA is most beneficial for improving solution quality. Using PSA during meta-training (MT) yields consistent hypervolume gains across all problem complexities, consistent with broader exposure to diverse trade-offs during inner-loop adaptation and improved generalisation. Applying PSA again after fine-tuning (MT\&FT) further increases hypervolume and produces the greatest improvements, particularly as complexity grows. These hypervolume gains are statistically significant under Kruskal--Wallis with Bonferroni-corrected Dunn post hoc tests. EUM generally improves in tandem, while sparsity effects are mixed, suggesting PSA primarily prioritises non-dominated solution quality and may trade off frontier spacing depending on when it is applied. Moreover, we conduct ablation on PSA Parameters ($K, S, \delta$) and scalarisation methods (including Tchebycheff~\citep{van_moffaert_scalarized_2013}), which is presented in Appendix F.2.

\subsubsection{Algorithm solutions analysis}

We visualise PF approximations using best-performing runs for each algorithm, as shown in Figure~\ref{fig:comp_pf}. Meta-learning methods produce competitive solutions with moderate concentration and low sparsity, indicating stability but limited diversity. MIRACL converges to high-quality regions, achieving high profit, low emissions, and low inequality. In simpler problems, it prioritises profit at the cost of higher emissions while maintaining lower inequality than MORL/D. A strong profit–emissions trade-off emerges, while inequality shows weaker correlation with other objectives. In complex settings, meta-learning methods yield more concentrated PFs due to few-shot adaptation constraints. However, MIRACL spans a broader range of profit outcomes with moderate emissions and inequality, outperforming NSGA-II in both diversity and optimality, demonstrating the value of PSA-based diversity mechanisms for better PF exploration. We further examine the operational behaviour and found that MIRACL achieves the most consistent production and inventory profiles over time, as discussed in detail in Appendix F.4.

\subsection{Domain-Agnostic MIRACL}

\begin{table}[t]
    \centering
    \caption{Cross-domain benchmarks (mean over 10 runs). The asterisk (*) denote statistically significantly different results using the Wilcoxon test on $p<0.05$, respectively against Meta-MORL.}
    \label{tab:generalisation}
    \begin{tabularx}{10 cm}{@{}llll@{}}
        \toprule
        Algorithm & Hypervolume & Sparsity & EUM \\
        \midrule
        \multicolumn{4}{l}{\textit{mo-hopper-v4}} \\
        \quad MIRACL    & 0.1108* & 0.0234* & 0.4003* \\
        \quad \textbf{Meta-MORL} & 0.0756 & 0.0047 & 0.3718 \\
        \midrule
        \multicolumn{4}{l}{\textit{mo-halfcheetah-v4}} \\
        \quad MIRACL    & 0.7002 & 0.0143 & 0.6910 \\
        \quad \textbf{Meta-MORL} & 0.6556 & 0.0114 & 0.6655 \\
        \midrule
        \multicolumn{4}{l}{\textit{resource-gathering-v0}} \\
        \quad MIRACL    & 0.7000 & 0.2051 & 0.7288 \\
        \quad \textbf{Meta-MORL} & 0.7000 & 0.1927 & 0.7727 \\
        \bottomrule
    \end{tabularx}
\end{table}

% We evaluate MIRACL on three MO-Gymnasium benchmarks~\cite{felten_toolkit_2023}: \textit{mo-hopper-v4}, \textit{mo-halfcheetah-v4} (continuous control with unbounded rewards), and \textit{resource-gathering-v0} (discrete with bounded rewards). Across ten runs (Table~\ref{tab:generalisation}), MIRACL improves hypervolume and EUM over Meta-MORL on both continuous tasks, while also producing a more spread solution set (higher sparsity). On \textit{resource-gathering-v0}, both methods match hypervolume, but Meta-MORL attains higher EUM with similar sparsity. Overall, MIRACL generalises across diverse MORL settings, with the clearest benefits in continuous, unbounded-reward environments.

We evaluate MIRACL on three MO-Gymnasium benchmarks~\citep{felten_toolkit_2023}: \textit{mo-hopper-v4}, \textit{mo-halfcheetah-v4} (continuous, unbounded rewards), and \textit{resource-gathering-v0} (discrete, bounded rewards). As shown in Table~\ref{tab:generalisation}, MIRACL demonstrates strong generalisation capabilities, particularly in continuous control tasks. It significantly outperforms Meta-MORL ($p < 0.05$) across all metrics in \textit{mo-hopper-v4} and achieves higher hypervolume and EUM in \textit{mo-halfcheetah-v4}. In the discrete \textit{resource-gathering-v0} task, MIRACL remains competitive, matching Meta-MORL's hypervolume with comparable EUM and sparsity. These results confirm that MIRACL's can also extend beyond SC environments.

\section{Conclusions and Future Works}

Our results show that meta-learning methods, including MIRACL, perform well in simple and moderate tasks with significantly fewer training steps than traditional RL. In complex scenarios, performance remains comparable. While meta-learning yields denser and more robust solution sets, this may reduce diversity. Integration of PSA-based diversity into MIRACL fine-tuning improves hypervolume, EUM, and sparsity, especially under higher task complexity. PF approximations show that both meta-learning-based and traditional MORL solutions often share solution areas. However, meta-learning achieves more clustered sets, particularly in complex tasks, although not as severe as NSGA-II. 
% This is due to the shared meta-policy and limited fine-tuning steps, which limit exploration and reduce diversity.
As tasks grow more complex, it is crucial to balance rapid adaptation with adequate exploration to improve exploration in meta-learning, such as by combining different meta-policy training methods.

% Use the style of numbering in square brackets.
% If nothing is used, default style will be taken.
%\begin{enumerate}[a)]
%\item 
%\item 
%\item 
%\end{enumerate}  

%\begin{itemize}
%\item 
%\item 
%\item 
%\end{itemize}  

%\begin{description}
%\item[]
%\item[] 
%\item[] 
%\end{description}  

% Uncomment and use as the case may be
%\begin{theorem} 
%\end{theorem}

% Uncomment and use as the case may be
%\begin{lemma} 
%\end{lemma}


\vspace{20pt}

\noindent\textbf{Acknowledgements}
The authors acknowledge the funding granted by the Economic and Social Research Council of UK Research and Innovation, the University of Manchester, and Peak AI, Ltd, with grant reference number ES/T002085/1. Research is carried out using computational resources from The University of Manchester and Peak AI, Ltd. We also acknowledge that the Peak AI, Ltd. Data Science team is involved in \textit{Messiah} development.
%\clearpage
%\newpage
% \bibliographystyle{cas-model2-names-no-url}

% \bibliography{references}



\end{document}

\appendix
\section*{SUPPLEMENTARY MATERIAL OF MIRACL: A Diverse Meta-Reinforcement Learning for Multi-Objective Multi-Echelon Combinatorial Supply Chain Optimisation}

\section{Notations} \label{sec:notation}
\begin{table}[H]
    \centering
    \label{tab:parameter_definition}
    \caption{The nodes $i$, $j$, and $k$ symbolise the SC network facilities, with period $t$ as the daily interval. The variables $\varsigma$, $\psi$, and $\tau$ signify inventory, production, and transport activities.}
    \begin{tabularx}{\linewidth}{lX}
        \toprule
        Notations & Definition\\
        \midrule
        $q^\tau_{tij}$ & Transport quantity at period $t$, from node $i$ to $j$\\
        $q^\psi_{tk}$ & Manufacturing quantity at period $t$, at manufacturer $k$\\
        $Prof$ & Total profit of all periods\\
        $E$ & Total emission of all periods\\
        $F$ & Total SL inequality measure of all periods\\
        $SL_y$ & SL at node $y$ of all periods\\
        $Rev_t$ & Revenue at period $t$\\
        $PC_t$ & Production cost at period $t$\\
        $TC_t$ & Transport cost at period $t$\\
        $IC_t$ & Inventory cost at period $t$\\
        $L$ & Transport lead time\\
        $I_{tj}$ & Inventory unit at period $t$, node $j$\\
        $c_j^\varsigma$ & Inventory cost per unit at node $j$\\
        $e^{\varsigma}_j$ & Emission per unit resulted from inventory at node $j$\\
        $c_k^\psi$ & Manufacturing cost per unit at node $k$\\
        $v_k^\psi$ & Yield ratio from manufacturing process at node $k$\\
        $e^{\psi}_k$ & Emission per unit resulted from the manufacturing process at node $k$\\
        $c_{ij}^\tau$ & Transport cost per unit from node $i$ to $j$\\
        $e^{\tau}_{ij}$ & Emission per unit resulted from product transport from node $i$ to $j$\\
        $\mathbf{Q_{tj}}$ & The outstanding order at node $j$\\
        $CE_t$ & The accumulated emission at period $t$\\
        $AF_t$ & The average service level inequality in period $t$\\
        $d_{tz}$ & Demand at period $t$, market $z$\\
        $Cap$ & Transport capacity\\
        \bottomrule     
    \end{tabularx}
\end{table}

\section{Problem Definition: MOMDP-Based Formulation} \label{sec:sc_problem}

The supply chain (SC) problem simulated in this paper adopts the SC optimisation problem based on the multi-objective Markov decision process (MOMDP) as referred to in the main text. We extended the problem to simpler and more complex SC networks for our experiments. $N^s, N^m, N^d, \text{ and } N^r$ represent sets of suppliers, manufacturers, distributors, and retailers respectively. For completeness purposes, we restate the formulation here.

\subsection{State Transition Functions}
State transition describes the shift of functions from the state at period $(t-1)$ to period $t$, as defined by Equations~\eqref{eq:rl_inv_transition} to~\eqref{eq:rl_avg_ineq}:

\begin{equation}
    \begin{split}
    \label{eq:rl_inv_transition}
    I_{tj} &= I_{(t-1)j} + \sum_{i \in (N^{s} \cup N^{m} \cup N^{d})} q_{(t-L)ij}^{\tau} - \sum_{k \in (N^{d} \cup N^{r} \cup N^{z})} q_{tjk}^{\tau}, \quad \forall j \in (N^{m} \cup N^{d} \cup N^{r}), \text{ where } i < j < k
    \end{split}
\end{equation}

\begin{equation}
    \begin{split}
    \label{eq:rl_order_vector}
    \mathbf{Q}_{tj} &= \big\{ q^{\tau}_{(t-L+1)ij}, \ldots, q^{\tau}_{tij} \,\big|\, \forall j \in (N^{m} \cup N^{d} \cup N^{r}), i \in (N^{s} \cup N^{m} \cup N^{d}), \quad \text{such that } i < j \big\}
    \end{split}
\end{equation}

\begin{equation}
    \label{eq:rl_cum_emission}
    CE_t = CE_{(t-1)} + E_t
\end{equation}

\begin{equation}
    \label{eq:rl_avg_ineq}
    AF_t = \frac{AF_{t-1} \cdot (t-1) + F_t}{t}.
\end{equation}

\subsection{State Space (S)}
At each timestep, the agents' observation of the simulation environment is described by the state. The state at period $t$ is defined as follows:
\begin{equation}
\begin{split}
    \label{eq:rl_state_vector}
    \mathbf{S_t} &= \{ I_{(t-1)j}, \mathbf{Q}_{(t-1)j}, CE_{t-1}, AF_{t-1} \, | \, 0 \leq I_{(t-1)j}, \forall j \in (N^{m} \cup N^{d} \cup N^{r}) \}.
\end{split}
\end{equation}

\subsection{Action Space (A)}
In our SC environment, the actions at period $t$ are given by:
\begin{equation}
    \label{eq:rl_action_vector}
    \begin{split}
    \mathbf{A_t} = & \{ q_{tij}^{\tau}, q_{tk}^{\psi} \,|\, 0 \leq q_{tij}^{\tau} \leq Cap, \quad 0 \leq q_{tk}^{\psi}, \forall i \in (N^{s} \cup N^{m} \cup N^{d}), \, j \in (N^{m} \cup N^{d} \cup N^{r}), k \in N^{m},\\
    & \text{ where } i < j \}.
    \end{split}
\end{equation}
\vspace{-10pt}
\subsection{Rewards (R)}
The problem has three goals: maximise profit while minimising greenhouse gas (GHG) emissions and service level (SL) inequality. Multi-objective reinforcement learning (MORL) agents maximise cumulative rewards with negative values for GHG emissions and SL inequality, resulting in the reward vector $\mathbf{R_t}=\{\mathit{Prof_t}, -\mathit{E_t}, -\mathit{F_t}\}$. Each objective is computed as follows:
\begin{equation}
    \label{eq:ga_profit}
    \begin{split}
    Prof_t &= \left( \sum_{i \in N^d} \sum_{j \in N^r} q_{(t-L)i j}^\tau \cdot \text{price} \right) - \left( \sum_{k \in N^m} \frac{ q_{tk}^\psi \cdot c^\psi_k}{v^\psi_k} \right) - \left( \sum_{i \in (N^{s} \cup N^{m} \cup N^{d})} \sum_{j \in (N^{m} \cup N^{d} \cup N^{r})} q^\tau_{tij} \cdot c^\tau_{ij} \cdot L \right) \\
    & \quad - \left( \sum_{j \in (N^{m} \cup N^{d} \cup N^{r})} I_{tj} \cdot c^\varsigma_j \right), \quad \text{where} \quad i < j,
    \end{split}
\end{equation}

\begin{equation}
    \label{eq:ga_emission}
    \begin{split}
        E_t =& \sum_{j \in (N^{m} \cup N^{d} \cup N^{r})} \left( I_{tj} \cdot e^\varsigma_j \right) + \sum_{k \in N^{m}} \left( q^\psi_{tk} \cdot e^\psi_k \right) + \sum_{i \in (N^{s} \cup N^{m} \cup N^{d})} \sum_{j \in (N^{m} \cup N^{d} \cup N^{r})} \left( q_{ij}^\tau \cdot e_{ij}^\tau \cdot L \right),\\ &\text{where} \quad i < j,
    \end{split}
\end{equation}

\begin{equation}
    \label{eq:ga_ineq}
    \begin{split}
    F_t &= \, \frac{1}{2} \sum_{y \in N^r} \sum_{\substack{y' \in N^r \\ y' \neq y}} \left| SL_{ty} - SL_{ty'} \right|.
    \end{split}
\end{equation}

While this model reference operates on deterministic cases, our SC environments involve uncertainties arising from lead time, cost, emissions, and market demand.

\section{Fine-Tuning Algorithm} \label{sec:fine_tuning_algo}
The pseudocode of the fine-tuning mechanism is given in Algorithm~\ref{algo:fine_tuning}.

\begin{algorithm}[H]
\caption{Fine-tuning Phase of MIRACL}
\label{algo:fine_tuning}
\begin{algorithmic}[1]
\REQUIRE Trained meta-policy $\pi_\theta$, unseen task $\mathcal{T}_{\text{new}} \sim p(\mathcal{T})$,
         number of solutions $K$, fine-tuning steps $T$, PSA fine-tuning steps $T_{\text{add}}$
\STATE \textbf{Initialise:} archive $PF \gets \emptyset$
\STATE Generate $K$ normalised weights $\{\mathbf{w}_1, \dots, \mathbf{w}_K\}$ on the simplex
\FOR{$k = 1$ to $K$}
    \STATE Form subproblem $\mathcal{T}_k$ using $(\mathcal{T}_{\text{new}}, \mathbf{w}_k)$
    \STATE Rollout $\mathcal{D}_k$ using $\pi_\theta$ in $\mathcal{T}_k$ for $T$ steps
    \STATE Compute adapted parameters $\theta'_k$ from $\mathcal{D}_k$
    \STATE Evaluate reward $\mathbf{r}_k$ in $\mathcal{T}_k$ using $\pi_{\theta'_k}$
\ENDFOR
\STATE Update $\mathbf{w}, PF \gets \text{DiversityMechanism}(\mathcal{T}_{\text{new}}, \mathbf{w}, \mathbf{r}, PF)$
\FOR{$k = 1$ to $K$}
    \STATE Form updated subproblem $\mathcal{T}_k$ using current $\mathbf{w}_k$
    \STATE Rollout additional $D_k$ in $\mathcal{T}_k$ and compute additional updates on $\theta'_k$ for $T_{\text{add}}$ steps
    \STATE Evaluate vector reward $\mathbf{r}_k$ in $\mathcal{T}_k$ using $\pi_{\theta'_k}$
\ENDFOR
\STATE \textbf{Return:} Fine-tuned policies $\{\pi_{\theta'_k}\}_{k=1}^K$, weights $\{\mathbf{w}_k\}_{k=1}^K$, updated $PF$
\end{algorithmic}
\end{algorithm}

\section{Simulated Environment} \label{sec:sc_setting}
Our experiment employs three SC environments: simple, moderate, and complex networks. SC networks are node collections linked by edges, with parameters set at nodes and edges. Each network is 'fully connected', linking all nodes across consecutive layers. Material supply from suppliers is treated as unlimited. Market demand is unstable: markets one and two fluctuate as normal distributions ($\mu=150, \sigma=60$ and $\mu=100, \sigma=40$) and markets three, four, and five as Poisson distributions ($\lambda=200, 100, 150$). Seasonal demand shifts through sinusoidal-modulated distributions. Simple SC serves markets 1 and 2, moderate SC covers markets 1-3, and complex SC involves all markets.

For simple SC markets, a fixed mean price of $20$ is applied, while moderate and complex SC markets encounter variable mean prices: $[20, 21, 20.5]$ and $[100, 101, 105, 103, 104]$. Unfulfilled demand is regarded as demand loss. This problem is structured within the MOMDP framework with a finite time horizon $T=100$ and a lead time range of $L=\{1,3\}$ days. The transportation capacity ($Cap$) is set at 200, while the manufacturing capacity is not restricted.

To simulate uncertainties originating mainly from demand, cost, price structure, and lead time, we randomise the parameters during environment generations. For example, in addition to the fluctuation of demand along the period horizon, we also introduce randomised mean and standard deviation values within a range of $\{90\%, 110\%\}$ of set values. A similar range is also applied to the costs and prices, subject to the task arrangement scenarios applied. During fine-tuning, new unseen tasks are given in the Table~\ref{tab:para_value_simple} to~\ref{tab:trans_value_complex}.

\begin{table}[H]
    \centering
    \small
    \caption{\textbf{Simple SC.} Node parameter values that correspond to the facilities in Simple SC environment.}
    \label{tab:para_value_simple}
    \begin{tabularx}{10 cm}{ccccccc}
        \toprule
        Node & $I_{0j}$ & $c_j^\varsigma$ & $e^\varsigma_j$ & $c_k^\psi$ & $v_k^\psi$ & $e^\psi_k$ \\
        \midrule
        3 & 380 & 0.1100 & 0.0002 & 2.0000 & 1.0000 & 5.0126\\
        4 & 350 & 0.1300 & 0.0002 & 2.2000 & 1.0000 & 4.5754\\
        5 & 400 & 0.1200 & 0.0002 & NA & NA & NA\\
        6 & 80  & 0.1500 & 0.0002 & NA & NA & NA\\
        \bottomrule      
    \end{tabularx}
\end{table}

\begin{table}[H]
    \centering
    \small
    \caption{\small Transport parameter values that correspond to product delivery.}
    \label{tab:trans_value_simple}
    \begin{tabularx}{7 cm}{cccc}
        \toprule
        From Node & To Node & $c^\tau_{ij}$ & $e^\tau_{ij}$ \\
        \midrule
        \multirow{2}{*}{1} & 2 & 0.2200 & 0.1258\\
        & 3 & 0.6900 & 0.3947\\
        \hline
        \multirow{2}{*}{2} & 4 & 1.0550 & 0.6035\\
        & 5 & 0.4300 & 0.2460\\
        \hline
        \multirow{2}{*}{3} & 4 & 0.4850 & 0.2774\\
        & 5 & 0.7500 & 0.4290\\
        \bottomrule
    \end{tabularx}
\end{table}

\begin{table}[H]
\centering
    \small
    \caption{\small \textbf{Moderate SC.} Node parameter values that correspond to the facilities in Moderate SC environment.}
    \label{tab:para_value_moderate}
    \begin{tabularx}{10cm}{ccccccc}
        \toprule
        Node & $I_{0j}$ & $c_j^\varsigma$ & $e^\varsigma_j$ & $c_k^\psi$ & $v_k^\psi$ & $e^\psi_k$ \\
        \midrule
        3 & 380 & 0.1100 & 0.0002 & 2.0000 & 1.0000 & 5.0126\\
        4 & 350 & 0.1300 & 0.0002 & 2.2000 & 1.0000 & 4.5754\\
        5 & 400 & 0.1200 & 0.0002 & 2.3000 & 1.0000 & 5.4491\\
        6 & 80 & 0.1500 & 0.0002 & NA & NA & NA\\
        7 & 110 & 0.2000 & 0.0002 & NA & NA & NA\\
        8 & 100 & 0.2500 & 0.0002 & NA & NA & NA\\
        9 & 80 & 0.3000 & 0.0002 & NA & NA & NA\\
        10 & 120 & 0.2000 & 0.0002 & NA & NA & NA\\
        \bottomrule      
    \end{tabularx}
\end{table}

\begin{table}[t]
\centering
    \small
    \caption{\small Transport parameter values that correspond to product delivery.}
    \label{tab:trans_value_moderate}
    \begin{tabularx}{7cm}{cccc}
        \toprule
        From Node & To Node & $c^\tau_{ij}$ & $e^\tau_{ij}$ \\
        \midrule
        \multirow{3}{*}{1} & 3 & 0.2200 & 0.1258\\
        & 4 & 0.6900 & 0.3947\\
        & 5 & 0.5650 & 0.3232\\
        \hline
        \multirow{3}{*}{2} & 3 & 1.0550 & 0.6035\\
        & 4 & 0.6500 & 0.3718\\
        & 5 & 0.6300 & 0.3604\\
        \hline
        \multirow{2}{*}{3} & 6 & 0.0750 & 0.0429\\
        & 7 & 0.4300 & 0.2460\\
        \hline
        \multirow{2}{*}{4} & 6 & 0.6300 & 0.3604\\
        & 7 & 0.2300 & 0.1316\\
        \hline
        \multirow{2}{*}{5} & 6 & 0.4950 & 0.2831\\
        & 7 & 0.0750 & 0.0429\\
        \hline
        \multirow{3}{*}{6} & 8 & 1.0950 & 0.6263 \\
        & 9 & 0.6250 & 0.3575\\
        & 10 & 0.9500 & 0.5434\\
        \hline
        \multirow{3}{*}{7} & 8 & 1.6400 & 0.9381\\
        & 9 & 1.1600 & 0.6635\\
        & 10 & 0.5800 & 0.3318\\
        \bottomrule
    \end{tabularx}
\end{table}

\begin{table}[H]
    \centering
    \small    
    \caption{\small \textbf{Complex SC.} Node parameter values that correspond to the facilities in Complex SC environment.}
    \label{tab:para_value_complex}
    \begin{tabularx}{10cm}{ccccccc}
        \toprule
        Node & $I_{0j}$ & $c_j^\varsigma$ & $e^\varsigma_j$ & $c_k^\psi$ & $v_k^\psi$ & $e^\psi_k$ \\
        \midrule
        4 & 155 & 0.2300 & 0.0002 & 2.0000 & 1.0000 & 5.0126\\
        5 & 267 & 0.3500 & 0.0002 & 2.2000 & 1.0000 & 4.5754\\
        6 & 342 & 0.2200 & 0.0002 & 2.1000 & 1.0000 & 5.4491\\
        7 & 211 & 0.1100 & 0.0002 & 2.0000 & 1.0000 & 6.1232\\
        8 & 162 & 0.2900 & 0.0002 & 2.3000 & 1.0000 & 5.5157\\
        9 & 195 & 0.3700 & 0.0002 & NA & NA & NA\\
        10 & 333 & 0.1100 & 0.0002 & NA & NA & NA\\
        11 & 96 & 0.3600 & 0.0002 & NA & NA & NA\\
        12 & 285 & 0.3300 & 0.0002 & NA & NA & NA\\
        13 & 68 & 0.2600 & 0.0002 & NA & NA & NA\\
        14 & 379 & 0.3000 & 0.0002 & NA & NA & NA\\
        15 & 344 & 0.1700 & 0.0002 & NA & NA & NA\\
        16 & 66 & 0.2900 & 0.0002 & NA & NA & NA\\
        17 & 356 & 0.2700 & 0.0002 & NA & NA & NA\\
        18 & 382 & 0.2300 & 0.0002 & NA & NA & NA\\
        19 & 362 & 0.3700 & 0.0002 & NA & NA & NA\\
        \bottomrule      
    \end{tabularx}
\end{table}

\begin{table}[H]
\centering
\small
\caption{\small Transport parameter values that correspond to product delivery.}
\label{tab:trans_value_complex}

\begin{minipage}[t]{0.48\linewidth}
\centering
\begin{tabular}{cccc}
\toprule
From Node & To Node & $c^\tau_{ij}$ & $e^\tau_{ij}$ \\
\midrule
\multirow{5}{*}{1} & 4 & 0.5350 & 0.3060\\
 & 5 & 0.2650 & 0.1516\\
 & 6 & 1.8450 & 1.0553\\
 & 7 & 1.6000 & 0.9152\\
 & 8 & 1.4400 & 0.8237\\
\midrule
\multirow{5}{*}{2} & 4 & 0.3600 & 0.2059\\
 & 5 & 0.2950 & 0.1687\\
 & 6 & 1.2350 & 0.7064\\
 & 7 & 0.6250 & 0.3575\\
 & 8 & 1.8550 & 1.0611\\
\midrule
\multirow{5}{*}{3} & 4 & 0.6000 & 0.3432\\
 & 5 & 0.1750 & 0.1001\\
 & 6 & 0.7450 & 0.4261\\
 & 7 & 1.3300 & 0.7608\\
 & 8 & 0.1700 & 0.0972\\
\midrule
\multirow{3}{*}{4} & 9 & 1.9900 & 1.1383\\
 & 10 & 0.3400 & 0.1945\\
 & 11 & 0.8100 & 0.4633\\
\midrule
\multirow{3}{*}{5} & 9 & 1.5150 & 0.8666\\
 & 10 & 0.6600 & 0.3775\\
 & 11 & 0.6450 & 0.3689\\
\midrule
\multirow{3}{*}{6} & 9 & 1.6950 & 0.9695\\
 & 10 & 1.5800 & 0.9038\\
 & 11 & 0.8150 & 0.4662\\
\bottomrule
\end{tabular}
\end{minipage}\hfill
\begin{minipage}[t]{0.48\linewidth}
\centering
\begin{tabular}{cccc}
\toprule
From Node & To Node & $c^\tau_{ij}$ & $e^\tau_{ij}$ \\
\midrule
\multirow{3}{*}{7} & 9 & 1.6150 & 0.9238\\
 & 10 & 1.2600 & 0.7207\\
 & 11 & 0.6750 & 0.3861\\
\midrule
\multirow{3}{*}{8} & 9 & 1.0300 & 0.5892\\
 & 10 & 1.0900 & 0.6235\\
 & 11 & 1.6300 & 0.9324\\
\midrule
\multirow{3}{*}{9} & 12 & 1.9650 & 1.1240 \\
 & 13 & 1.9250 & 1.1011\\
 & 14 & 1.6200 & 0.9266\\
\midrule
\multirow{3}{*}{10} & 8 & 1.4900 & 0.8523\\
 & 9 & 1.9600 & 1.1211\\
 & 10 & 0.6350 & 0.3632\\
\midrule
\multirow{3}{*}{11} & 8 & 1.8700 & 1.0696\\
 & 9 & 0.2000 & 0.1144\\
 & 10 & 1.8550 & 1.0611\\
\midrule
\multirow{5}{*}{12} & 15 & 1.9450 & 1.1125\\
 & 16 & 0.9650 & 0.5520\\
 & 17 & 1.9050 & 1.0897\\
 & 18 & 0.9000 & 0.5148\\
 & 19 & 0.6900 & 0.3947\\
\midrule
\multirow{5}{*}{13} & 15 & 0.8050 & 0.4605\\
 & 16 & 1.0650 & 0.6092\\
 & 17 & 1.8400 & 1.0525\\
 & 18 & 0.8300 & 0.4748\\
 & 19 & 1.8850 & 1.0782\\
\midrule
\multirow{5}{*}{14} & 15 & 1.6600 & 0.9495\\
 & 16 & 1.5100 & 0.8637\\
 & 17 & 0.5900 & 0.3375\\
 & 18 & 0.4000 & 0.2288\\
 & 19 & 1.3950 & 0.7979\\
\bottomrule
\end{tabular}
\end{minipage}

\end{table}

\section{Detailed Algorithm Settings} \label{sec:detailed_algo_setting}
This section provides the detailed hyperparameters used in our experiment.

\subsection{Meta-training settings}
Table~\ref{tab:hyperparameters_ml} and~\ref{tab:hyperparameters_ppo} show the algorithm settings we used for MIRACL and Meta-MORL during our experiments.
\begin{table}[H]
    \centering
    \caption{Hyperparameters for MAML-based meta-training.}
    \label{tab:hyperparameters_ml}
    \begin{tabularx}{7cm}{lr}
        \toprule
        Hyperparameter & Value \\
        \midrule
        Inner-loop learning rate $\alpha$ & 0.003 \\
        Outer-loop learning rate $\beta$ & 0.001 \\
        PPO steps per meta-update & 10 \\
        Adaptation steps & 4--8 \\
        Network architecture & [64, 64] \\
        Discount factor $\gamma$ & 0.99 \\
        GAE weight & 1.0 \\
        VF loss coefficient & 0.5 \\
        Clipping parameter & 0.3 \\
        KL target & 0.01 \\
        KL coefficient & 0.001 \\
        Batch size & 32 \\
        Fragment length & 32 \\
        Rollout workers & 4 \\
        Evaluation interval & 100 \\
        \bottomrule
    \end{tabularx}
\end{table}

\subsection{Fine-tuning settings}
% In fine-tuning, the SC simulator Messiah~\cite{singh_multi-objective_2025} is used in 5,000, 15,000, and 20,000 time steps for simple, moderate, and complex tasks, respectively. \blue{These time steps are much fewer than used by traditional RL methods, which typically need around 500,000 time steps in our SC problem.} The number of shots matches meta-training. Observation and action spaces are normalised as recommended.

% Since PPO is inherently single-objective, we apply reward scalarisation using \blue{simplex-projected weights. Following standard practice in MORL, minimisation objectives (emissions and inequality) are converted to maximisation by sign-flipping and all objectives are then normalised to the range $[0,1]$ before nearest-neighbour comparison and scalarisation.} For each instantiation, we train 21 weighted agents \blue{($K=21$) across 10 seeds (210 total).} Then, we derive 10 PF approximation sets using non-dominated sorting~\cite{singh_multi-objective_2025}. We use \blue{generalised advantage estimation, a technique that enhances estimate accuracy by lowering variability through averaging the gaps between predicted and actual outcomes over time,} in order to reduce variance while preserving control over bias. Normalising advantages further stabilises training. The agent uses a standard multi-layer perceptron architecture. The algorithms are trained in three SC environments with various network complexities: simple, moderate, and complex.

\begin{table}[H]
    \centering
    \caption{PPO hyperparameters used in fine-tuning.}
    \label{tab:hyperparameters_ppo}
    \begin{tabularx}{7cm}{lr}
    \toprule
    Hyperparameter & Value \\
    \midrule
    Learning rate & $3 \times 10^{-4}$ \\
    Steps per update & 2048 \\
    Minibatch size & 64 \\
    Epochs per update & 10 \\
    Discount factor $\gamma$ & 0.99 \\
    GAE weight $\lambda$ & 0.95 \\
    Clipping range & 0.2 \\
    Entropy coefficient & 0 \\
    VF loss coefficient & 0.5 \\
    Max gradient norm & 0.5 \\
    Episode length & 100 \\
    Total timesteps & $5 \times 10^5$ \\
    Activation & Tanh \\
    \bottomrule
    \end{tabularx}
\end{table}

\subsection{MORL with decomposition (MORL/D) setup}

Table~\ref{tab:hyperparameter_MORLD} shows the setup used for the MORL/D algorithm as one of the baselines.
\begin{table}[H]
    \centering
    \caption{Hyperparameters of MORL/D Algorithm with MOSAC Policy}
    \label{tab:hyperparameter_MORLD}
    \begin{tabularx}{7cm}{lr}
        \toprule
        Hyperparameters & Values \\
        \midrule
        \textbf{MORL/D} & \\
        Discount factor & 0.995 \\
        Population size & 6 \\
        Exchange every & $5 \times 10^4$ \\
        Total time steps & $5 \times 10^5$ \\
        Neighbourhood size & 1 \\
        Update passes & 10 \\
        Initial weight dist. & uniform \\
        \midrule
        \textbf{MOSAC} & \\
        Buffer size & $10^6$ \\
        Discount factor & 0.99 \\
        Target smoothing coef. & 0.005 \\
        Batch size & 128 \\
        Steps before learning & $10^3$ \\
        Hidden neurons & [256; 256] \\
        Actor LR & $3 \times 10^{-4}$ \\
        Critic LR & $10^{-3}$ \\
        Actor training freq. & 2 \\
        Target training freq. & 1 \\
        Activation function & ReLU \\
        \bottomrule
    \end{tabularx}
\end{table}

\subsection{Non-dominated sorting genetic algorithm II (NSGA-II) setup}
Table~\ref{tab:nsga_parameters} shows the setup used in NSGA-II as one of the baselines.
\begin{table}[H]
    \centering
    \caption{Hyperparameters of NSGA-II to solve the SC problem across three network complexities: simple, moderate, and complex.}
    \label{tab:nsga_parameters}
    \begin{tabularx}{7cm}{lr}
        \toprule
        Hyperparameters & Values\\
        \midrule
        Population size & 300\\
        Offspring number & 30\\
        Cross-over operator & binary (SBX)\\
        Cross-over probability & 90\%\\
        Cross-over $\eta$ & 15\\
        Mutation method & polynomial\\
        Mutation $\eta$ & 20\\
        \bottomrule
    \end{tabularx}
\end{table}

\section{Performance Details}
\subsection{Traditional Method Comparison: Extended Analysis} \label{sec:extended_traditional_comp}

% A Kruskal-Wallis test in normalised hypervolume reveals statistically significant differences \blue{10 runs per method for each problem setting} ($p < 0.001$), indicating that at least one method performs differently. Dunn’s post hoc analysis further shows that MIRACL performs significantly better than NSGA-II across all problem settings ($p < 0.001$). Furthermore, MIRACL significantly outperforms MORL/D in the simple problem and Meta-MORL in the moderate problem ($p < 0.05$). \blue{All post hoc $p$-values are Bonferroni-corrected to account for multiple comparisons}. Notably, in all cases where MIRACL shows significant differences, it achieves a higher hypervolume. In the complex problem, although MIRACL is outperformed by MORL/D, the difference is not statistically significant, suggesting that MIRACL can remain competitive with traditional RL methods even as problem complexity increases, while still operating with substantially fewer time steps.

% Additional metrics in Table~\ref{fig:comp_rl_2} show that MIRACL and other meta-learning methods produce denser solution sets along the PF, indicating robustness but potentially reduced diversity. The EUM mirrors the hypervolume trend: MIRACL outperforms MORL/D in simpler tasks but lags behind in complex ones. This highlights a trade-off between early generalisation and adaptive precision as problem complexity increases.

\begin{figure*}
    \centering
    % First row
    \begin{subfigure}{0.32\linewidth}
        \includegraphics[width=\linewidth]{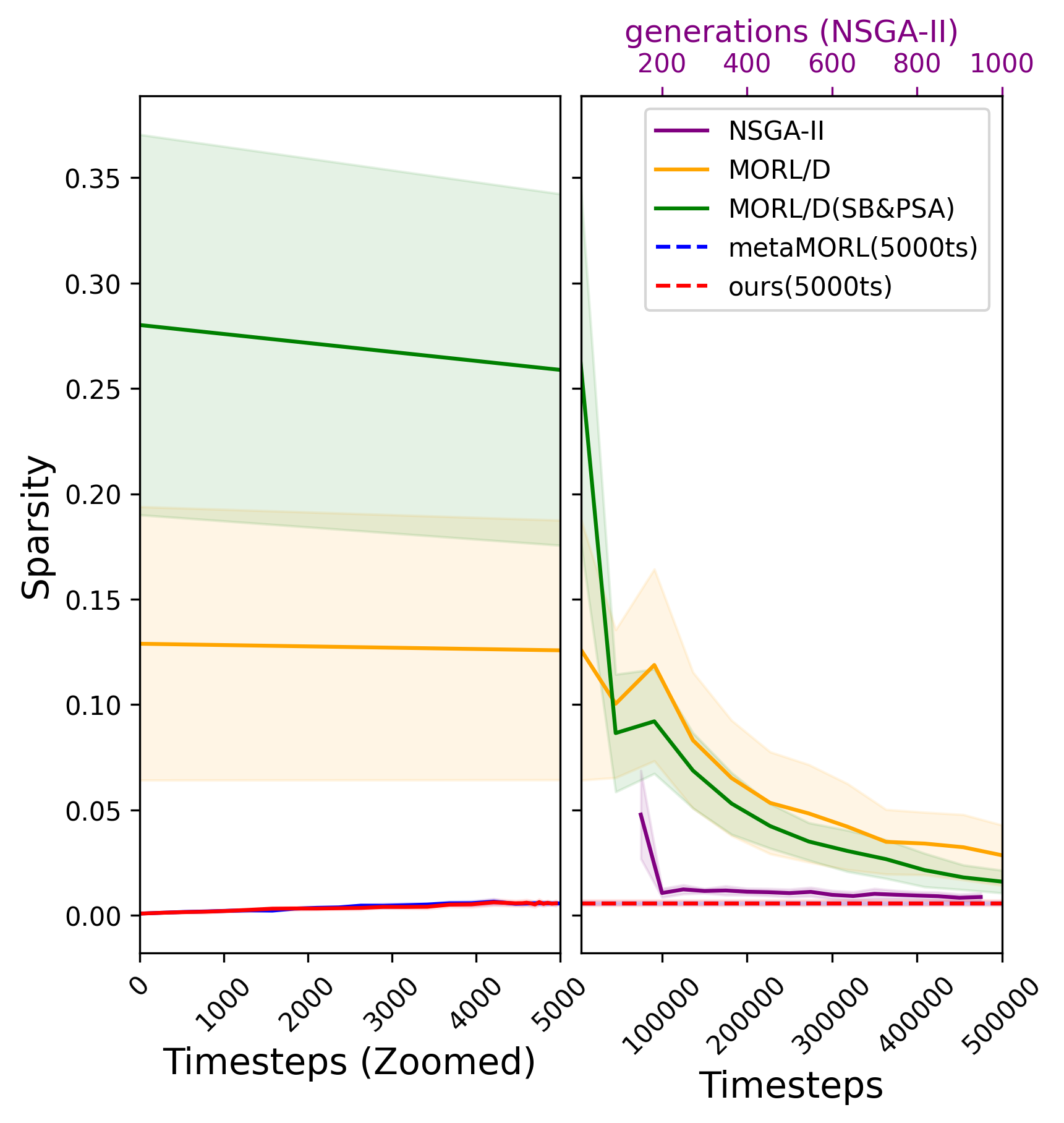}
        \caption{Sparsity($\downarrow$)-Simple}
        \label{fig:spar_comp_rl_simple}
    \end{subfigure}
    \hfill
    \begin{subfigure}{0.32\linewidth}
        \includegraphics[width=\linewidth]{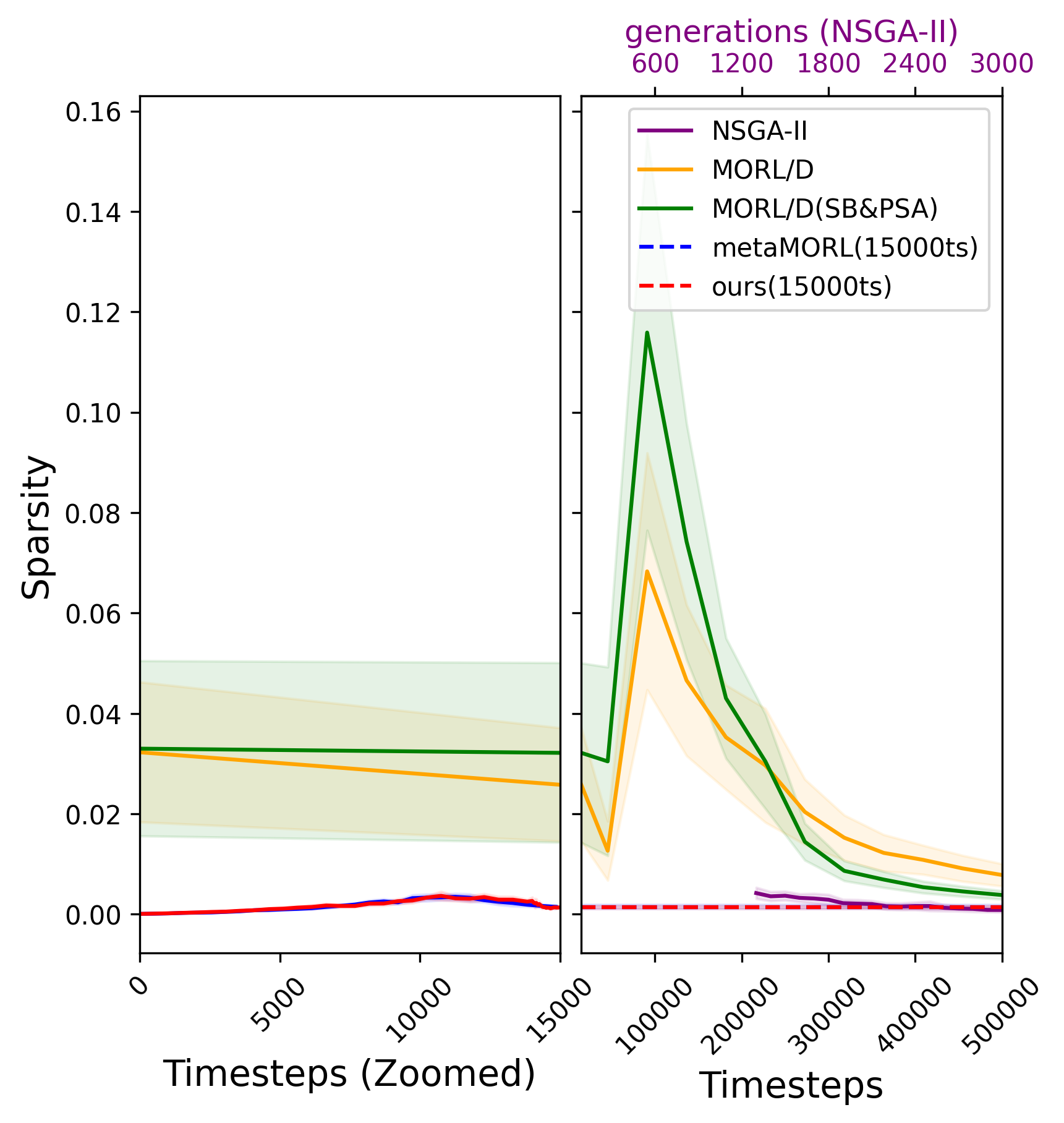}
        \caption{Sparsity($\downarrow$)-Moderate}
        \label{fig:spar_comp_rl_moderate}
    \end{subfigure}
    \hfill
    \begin{subfigure}{0.32\linewidth}
        \includegraphics[width=\linewidth]{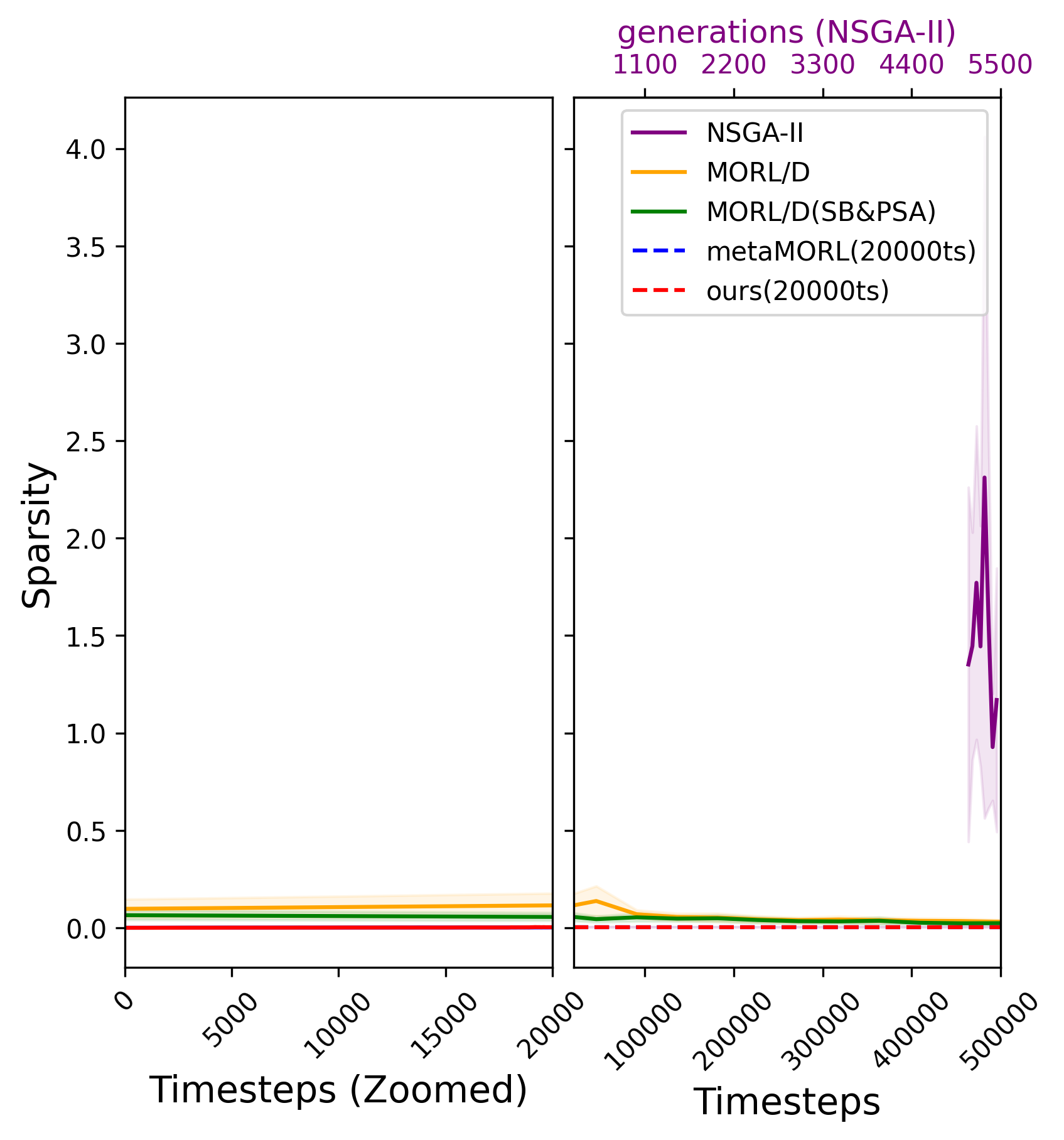}
        \caption{Sparsity($\downarrow$)-Complex}
        \label{fig:spar_comp_rl_complex}
    \end{subfigure}
    
    \vspace{0.3cm} % spacing between rows

    % Second row
    \begin{subfigure}{0.32\linewidth}
        \includegraphics[width=\linewidth]{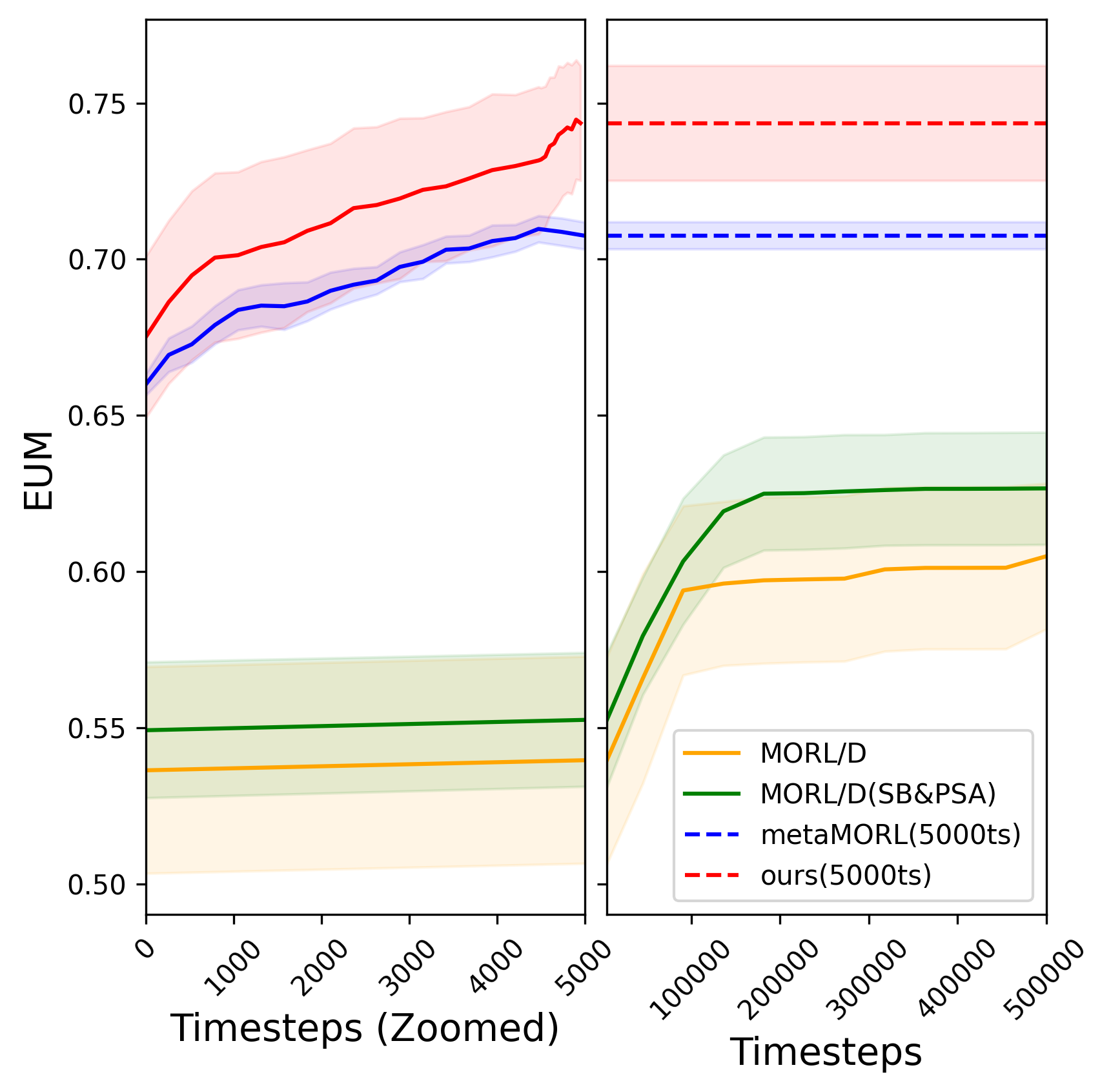}
        \caption{EUM($\uparrow$)-Simple}
        \label{fig:eum_comp_rl_simple}
    \end{subfigure}
    \hfill
    \begin{subfigure}{0.32\linewidth}
        \includegraphics[width=\linewidth]{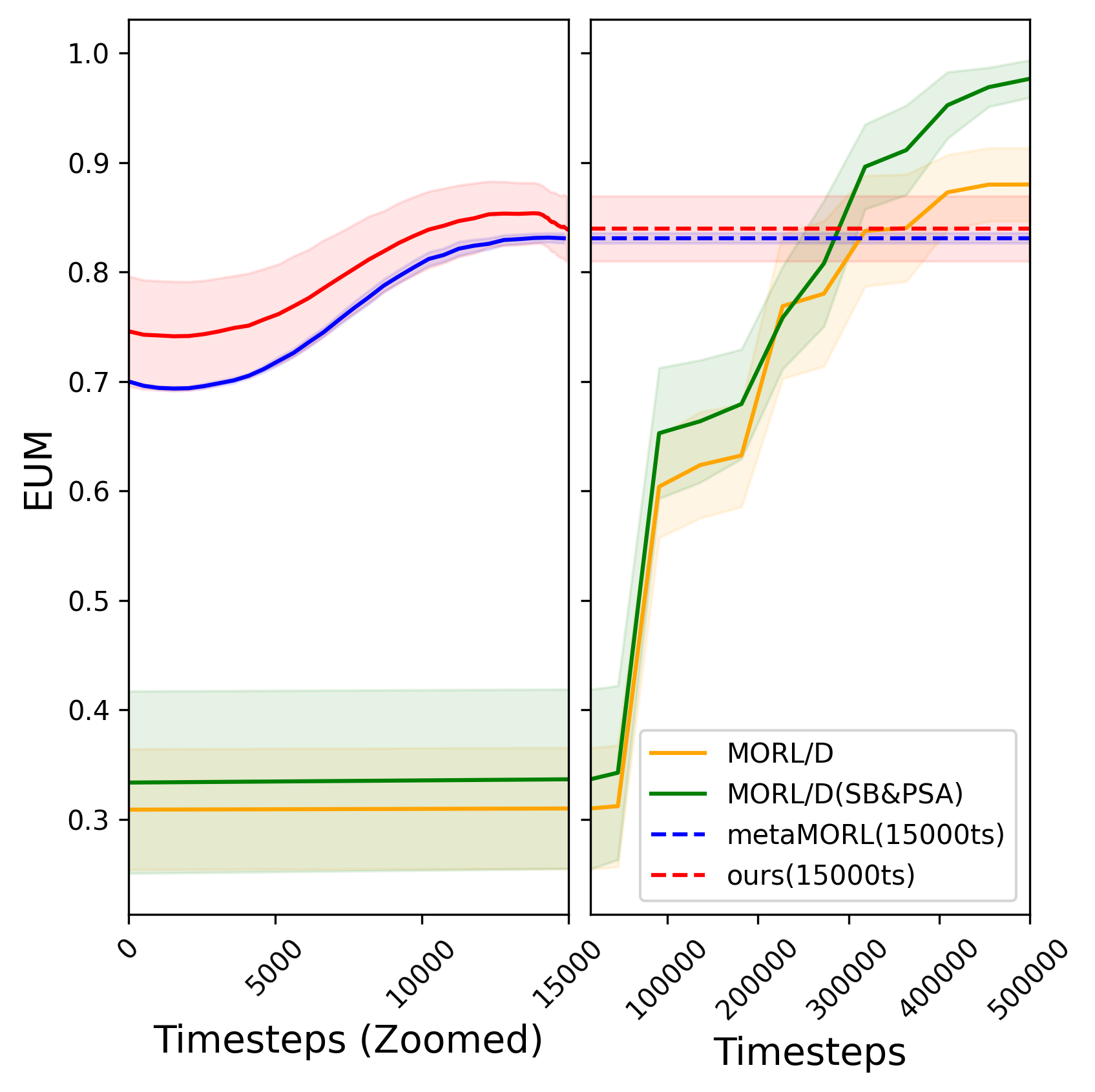}
        \caption{EUM($\uparrow$)-Moderate}
        \label{fig:eum_comp_rl_moderate}
    \end{subfigure}
    \hfill
    \begin{subfigure}{0.32\linewidth}
        \includegraphics[width=\linewidth]{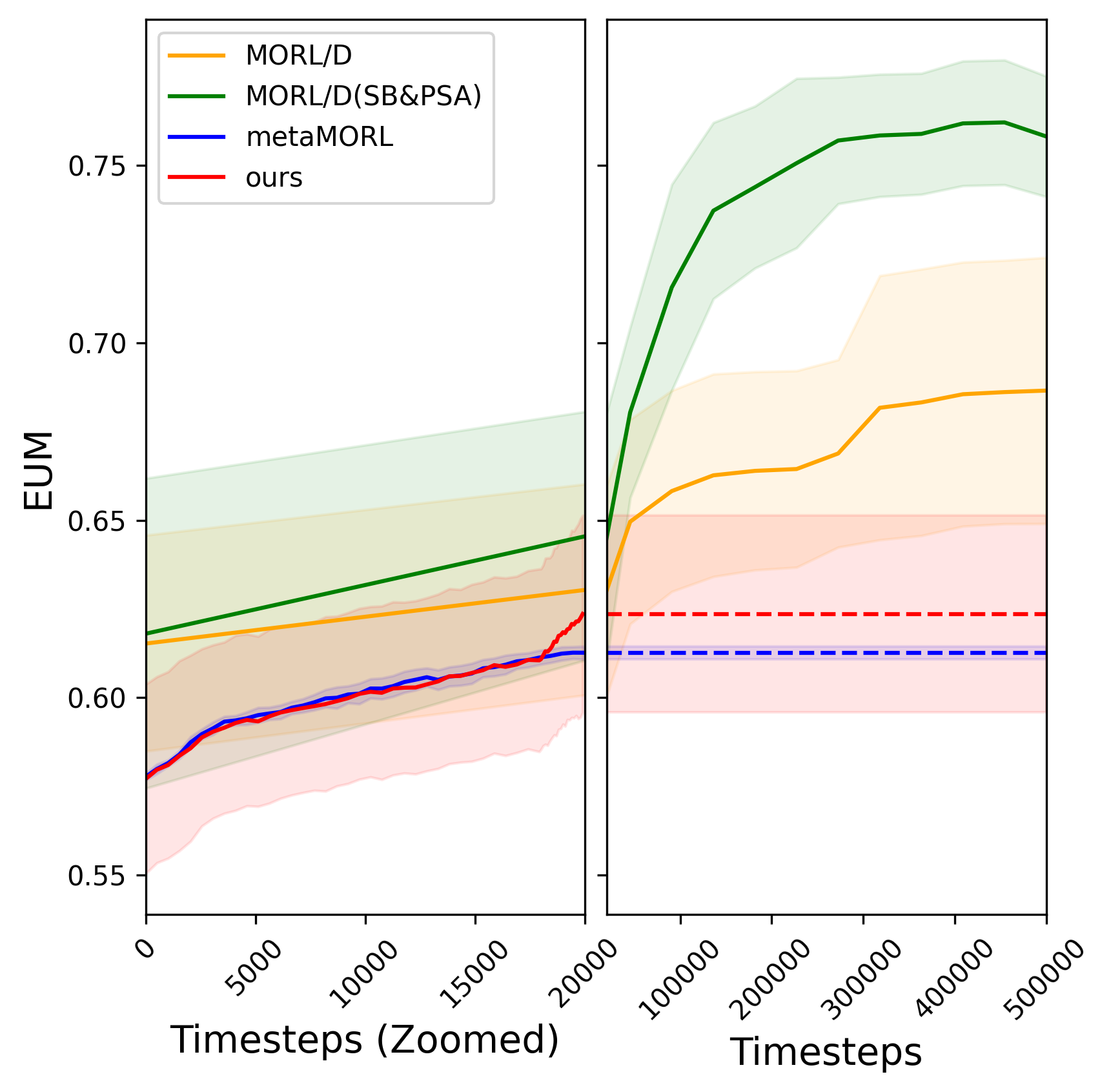}
        \caption{EUM($\uparrow$)-Complex}
        \label{fig:eum_comp_rl_complex}
    \end{subfigure}
    
    \caption{Sparsity and EUM of NSGA-II, MORL/D, MORL/D with SB and PSA, Meta-MORL, and our proposed algorithms. The left parts are zoomed in on the early timesteps with meta-learning, while the right parts show MORL/D timesteps and NSGA-II generations. The EUM reveal that meta-learning methods outperform MORL/D in simpler problems but increasingly struggle with more complex ones. Normalised sparsity consistently shows the lowest values for meta-learning-based methods across all complexities.}
    \label{fig:comp_rl_2}
\end{figure*}

Figure~\ref{fig:comp_rl_2} presents a comparison of sparsity and EUM between the proposed MIRACL, Meta-MORL approach, and traditional RL methods.

\subsection{Ablation Study} \label{sec:additional_ablation}
\begin{table}[h]
    \centering
    \caption{Ablation study for MIRACL on simple SC (mean over 10 runs), where *, **, *** denote statistically significant differences: $p<0.05$, $p<0.01$, and $p<0.001$, respectively. The Kruskal--Wallis followed with Dunn's post hoc (Bonferroni-corrected) is used for data sets $> 2$, and the Wilcoxon test for data sets $=2$. Bold text represents the baseline value of each parameter.}
    \label{tab:ablation_combined}
    \setlength{\tabcolsep}{3.5pt}
    \renewcommand{\arraystretch}{1.05}
    \begin{tabularx}{10 cm}{l l l l}
        \toprule
        Setting & Hypervolume & Sparsity & EUM \\
        \midrule

        \multicolumn{4}{l}{\textit{Number of subproblems $K$}} \\
        $K=3$  & 0.3832*** & 0.0088 & 0.7156 \\
        $K=5$  & 0.4003* & 0.0064 & 0.7146 \\
        $\mathbf{K=10}$ & 0.4387 & 0.0076 & 0.7141 \\
        \midrule

        \multicolumn{4}{l}{\textit{PSA steps $S$}} \\
        $S=3$  & 0.4371 & 0.0060 & 0.7055** \\
        $S=5$  & 0.4487 & 0.0070 & 0.7108 \\
        $\mathbf{S=10}$ & 0.4592 & 0.0063 & 0.7155 \\
        \midrule

        \multicolumn{4}{l}{\textit{PSA rate $\delta$}} \\
        $\delta=0.03$ & 0.4550 & 0.0060 & 0.7156 \\
        $\delta=0.05$ & 0.4512 & 0.0071 & 0.7150 \\
        $\mathbf{\delta=0.10}$ & 0.4564 & 0.0071 & 0.7122 \\
        \midrule

        \multicolumn{4}{l}{\textit{Scalarisation methods}} \\
        \textbf{Linear} & 0.4600 & 0.0057 & 0.7436 \\
        Tchebycheff & 0.4630 & 0.0081*** & 0.7148*** \\
        \bottomrule
    \end{tabularx}
    \vspace{1mm}
    % {\footnotesize{*} Sparsity measures spacing (too high: gaps; too low: redundancy)}
\end{table}

Table~\ref{tab:ablation_combined} suggests three practical configuration insights for MIRACL on simple SC. First, too few subproblems degrades the learned frontier, so increasing $K$ matters more than marginal tuning elsewhere. Second, reducing PSA depth harms EUM more than hypervolume, indicating PSA is key for steering the archive toward solutions preferred under the evaluation weight distribution, not merely enlarging the dominated region. Third, Tchebycheff yields a more dispersed frontier but lower utility, whereas linear scalarisation gives the strongest preference-weighted performance. Overall, MIRACL is relatively insensitive to $\delta$, while $K$ and PSA depth $S$ are the main levers for coverage and preference-aligned performances.

\subsection{PF Approximate}
\begin{figure*}[h]
    \centering
    \begin{subfigure}{0.32\textwidth}
        \includegraphics[width=\textwidth]{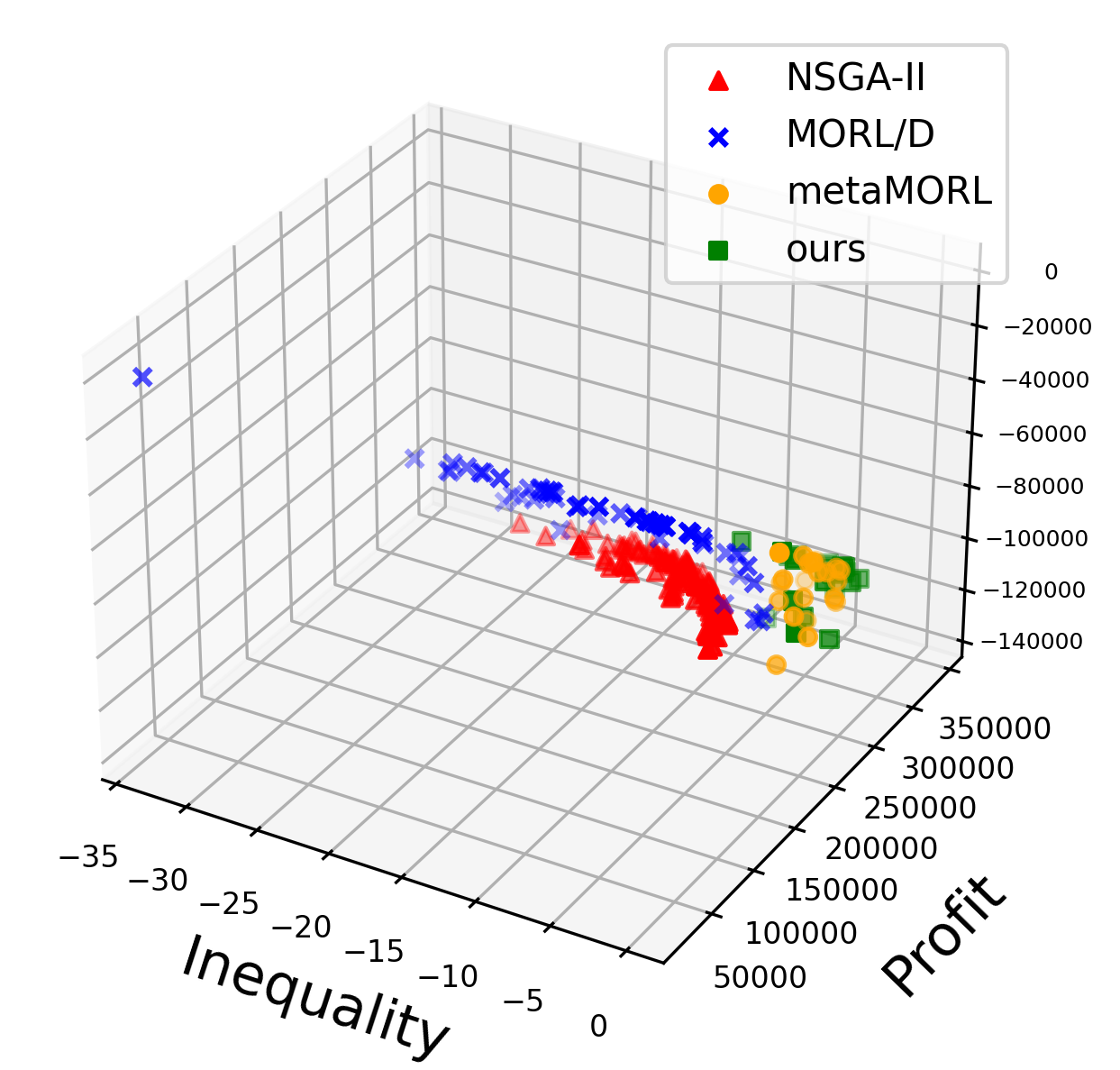}
        \caption{PF Simple - view 2}
        \label{fig:3D-2_comp_pf_simple}
    \end{subfigure}
    \hfill
        \begin{subfigure}{0.32\textwidth}
        \includegraphics[width=\textwidth]{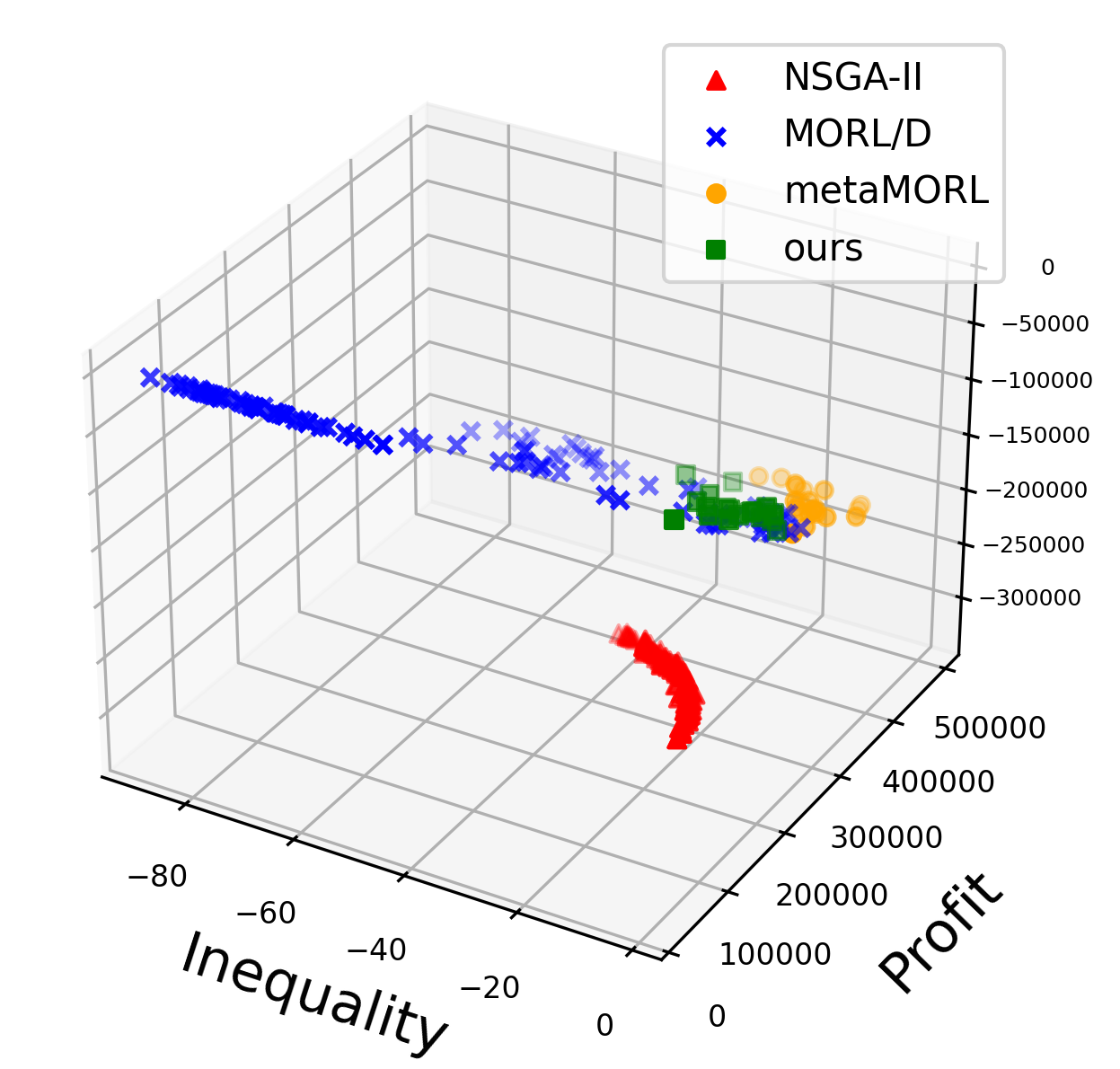}
        \caption{PF Moderate - view 2}
        \label{fig:3D-2_comp_pf_moderate}
    \end{subfigure}
    \hfill
        \begin{subfigure}{0.32\textwidth}
        \includegraphics[width=\textwidth]{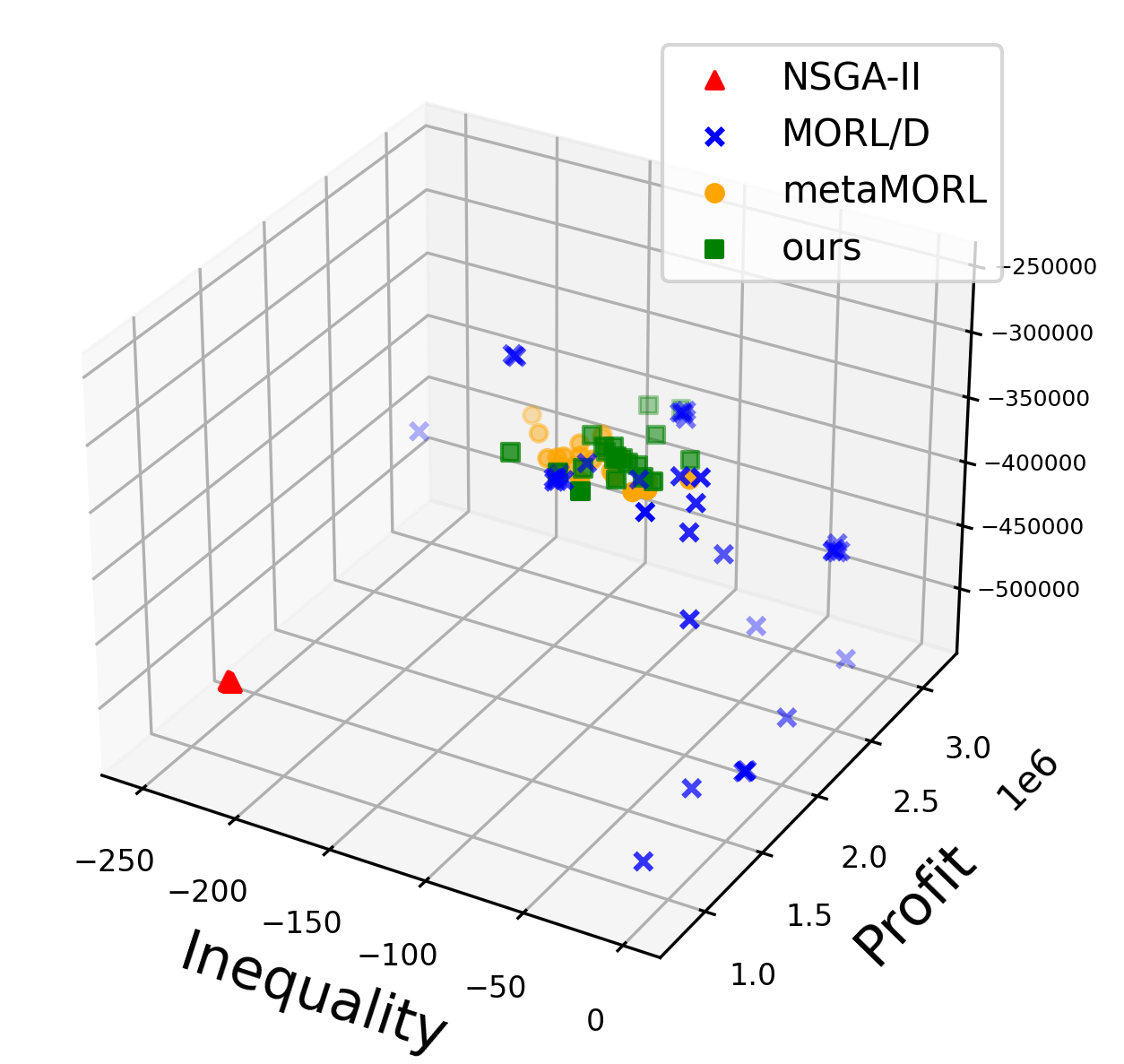}
        \caption{PF Complex - view 2}
        \label{fig:3D-2_comp_pf_complex}
    \end{subfigure}
    
    \begin{subfigure}{0.32\textwidth}
        \includegraphics[width=\textwidth]{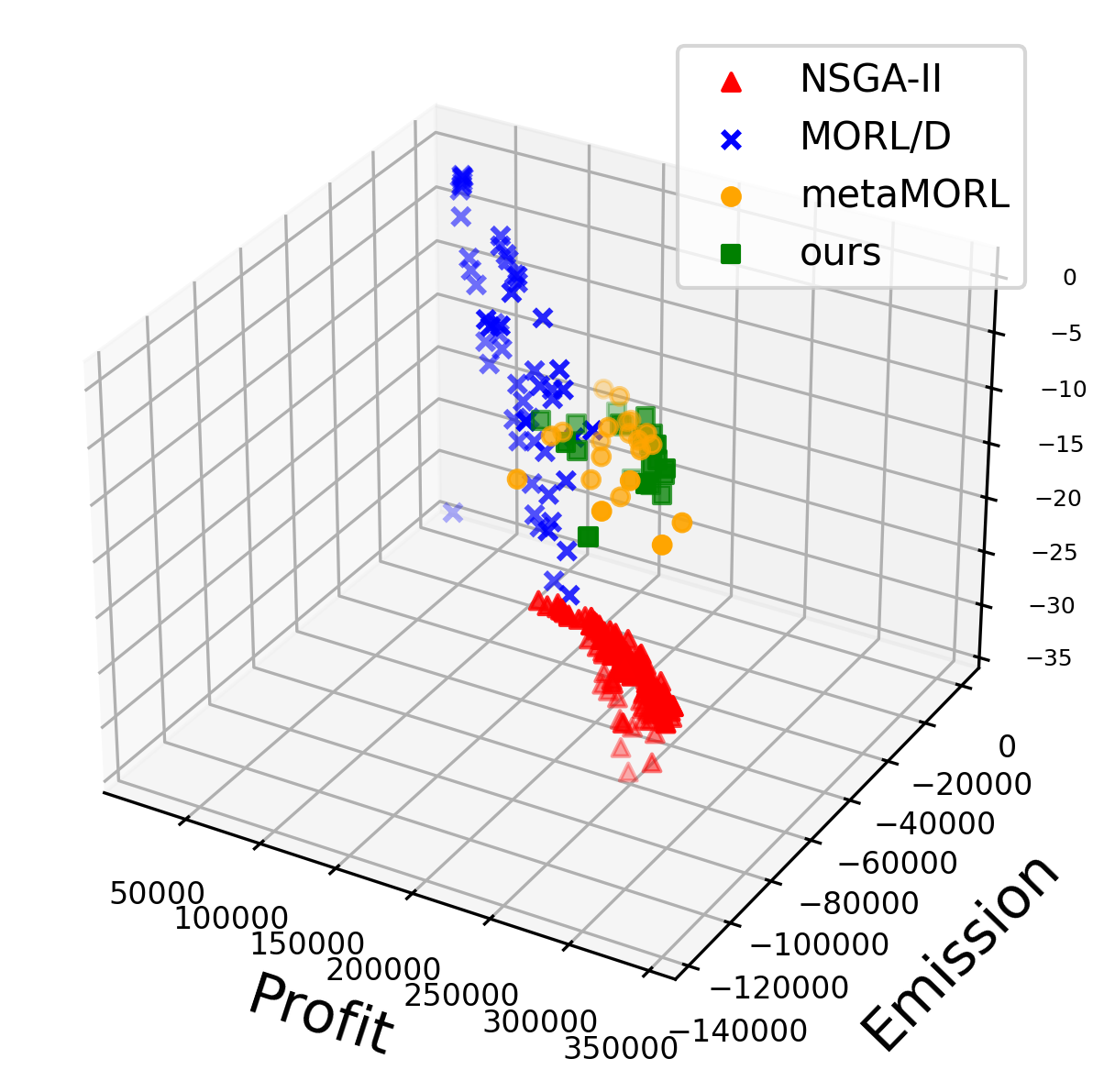}
        \caption{PF Simple - view 3}
        \label{fig:3D-3_comp_pf_simple}
    \end{subfigure}
    \hfill
    \begin{subfigure}{0.32\textwidth}
        \includegraphics[width=\textwidth]{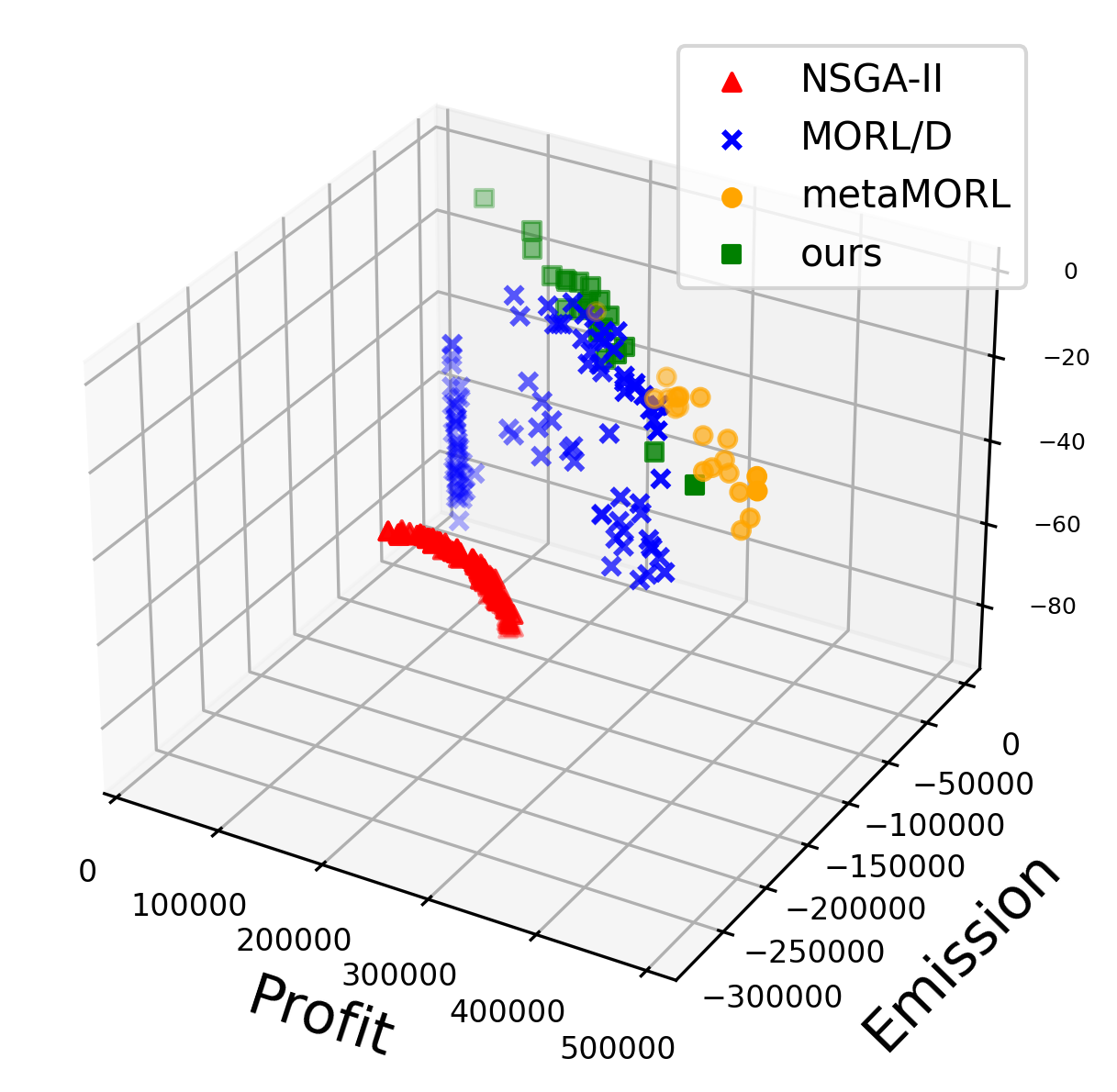}
        \caption{PF Moderate - view 3}
        \label{fig:3D-3_comp_pf_moderate}
    \end{subfigure}
    \hfill
    \begin{subfigure}{0.32\textwidth}
        \includegraphics[width=\textwidth]{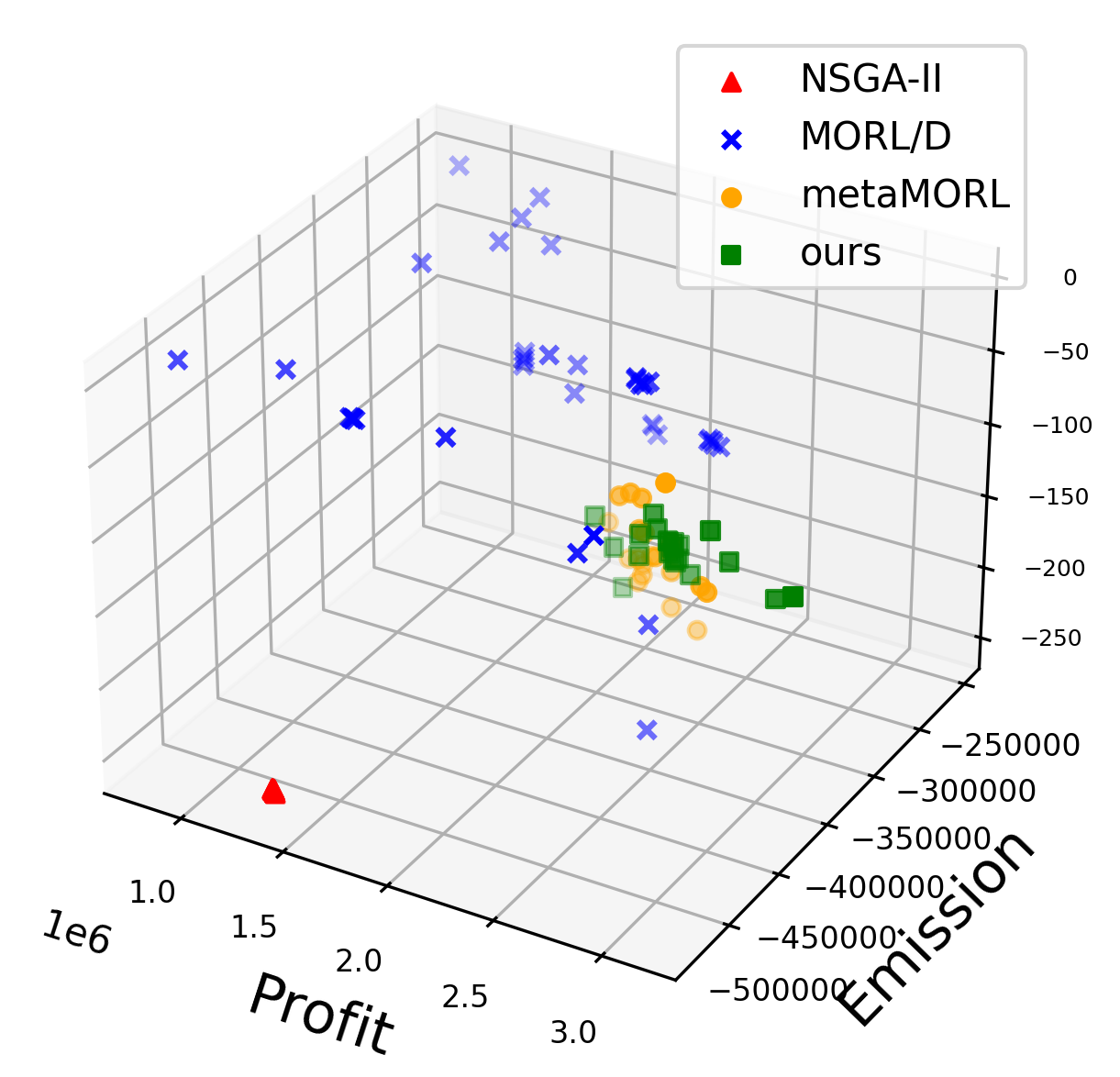}
        \caption{PF Complex - view 3}
        \label{fig:3D-3_comp_pf_complex}
    \end{subfigure}
    \caption{PF approximation sets between algorithms in all problem complexities. In general, meta-learning-based algorithms reside in a narrower solution space. The solutions consistently spread in the range of moderate to high values of profit and GHG emissions, while low to moderate values of SL inequality in all problem complexities.}
    \label{fig:comp_pf_2}
\end{figure*}

Figure~\ref{fig:comp_pf_2} shows the resulting solutions for each method from different points of view to complement the main text.

\subsection{Operational Behaviour} \label{sec:operational_behaviour}
\begin{figure*}[h]
    \centering
    \begin{subfigure}{0.32\textwidth}
        \includegraphics[width=\textwidth]{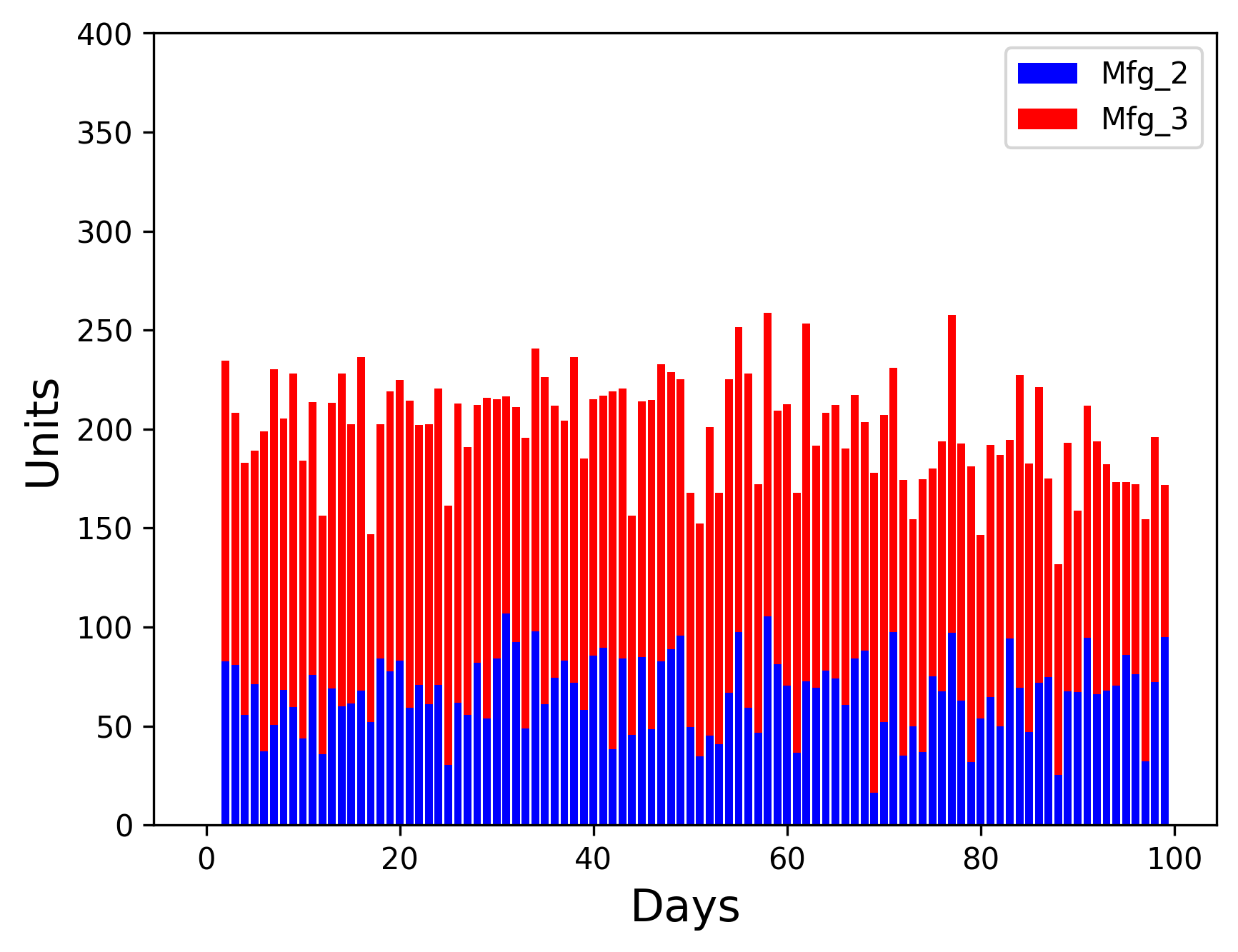}
        \caption{MORL/D - Simple}
        \label{fig:mfg_morld_simple}
    \end{subfigure}
    \hfill
    \begin{subfigure}{0.32\textwidth}
        \includegraphics[width=\textwidth]{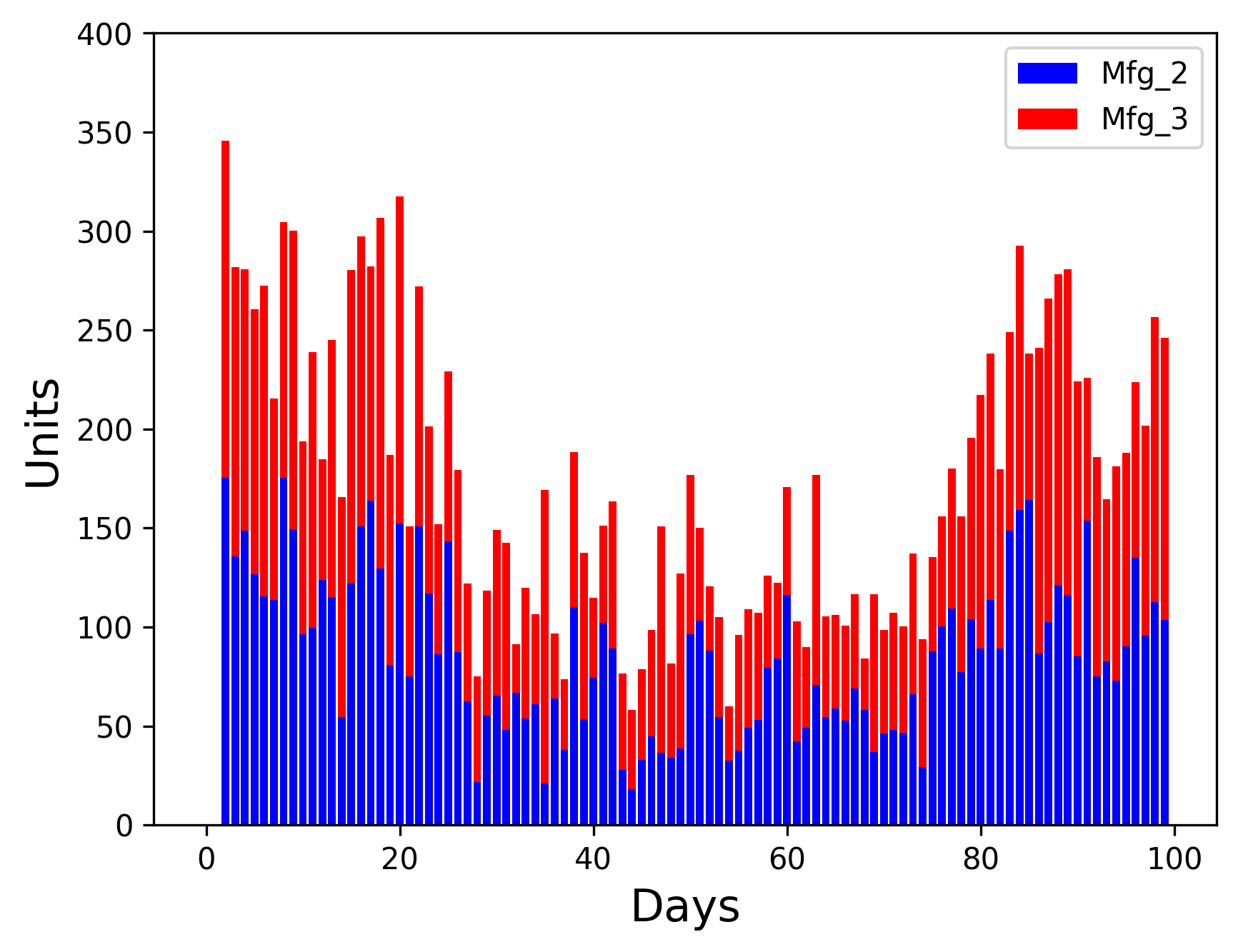}
        \caption{NSGA-II - Simple}
        \label{fig:mfg_nsga2_simple}
    \end{subfigure}
    \hfill
    \begin{subfigure}{0.32\textwidth}
        \includegraphics[width=\textwidth]{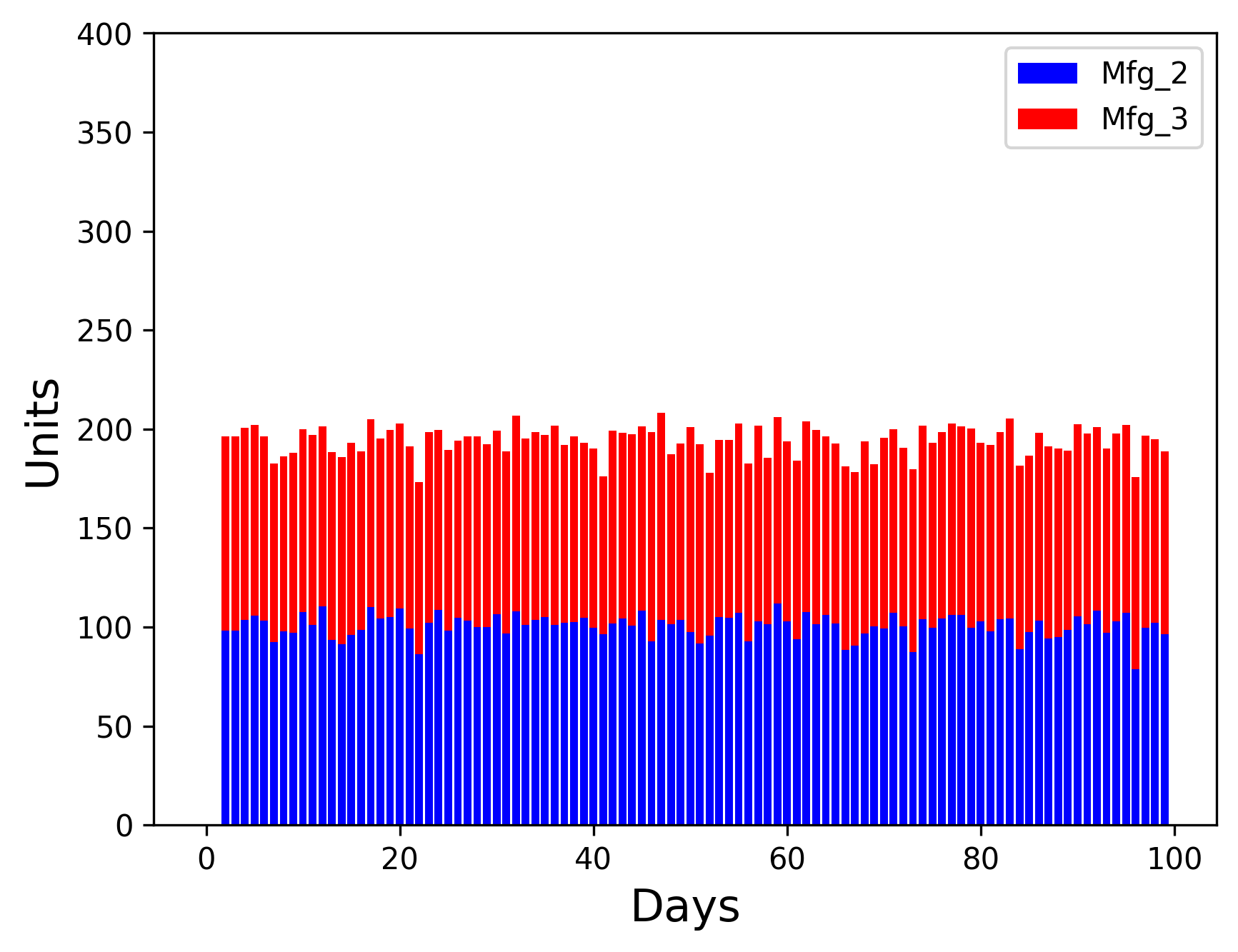}
        \caption{Ours - Simple}
        \label{fig:mfg_ours_simple}
    \end{subfigure}

    \begin{subfigure}{0.32\textwidth}
        \includegraphics[width=\textwidth]{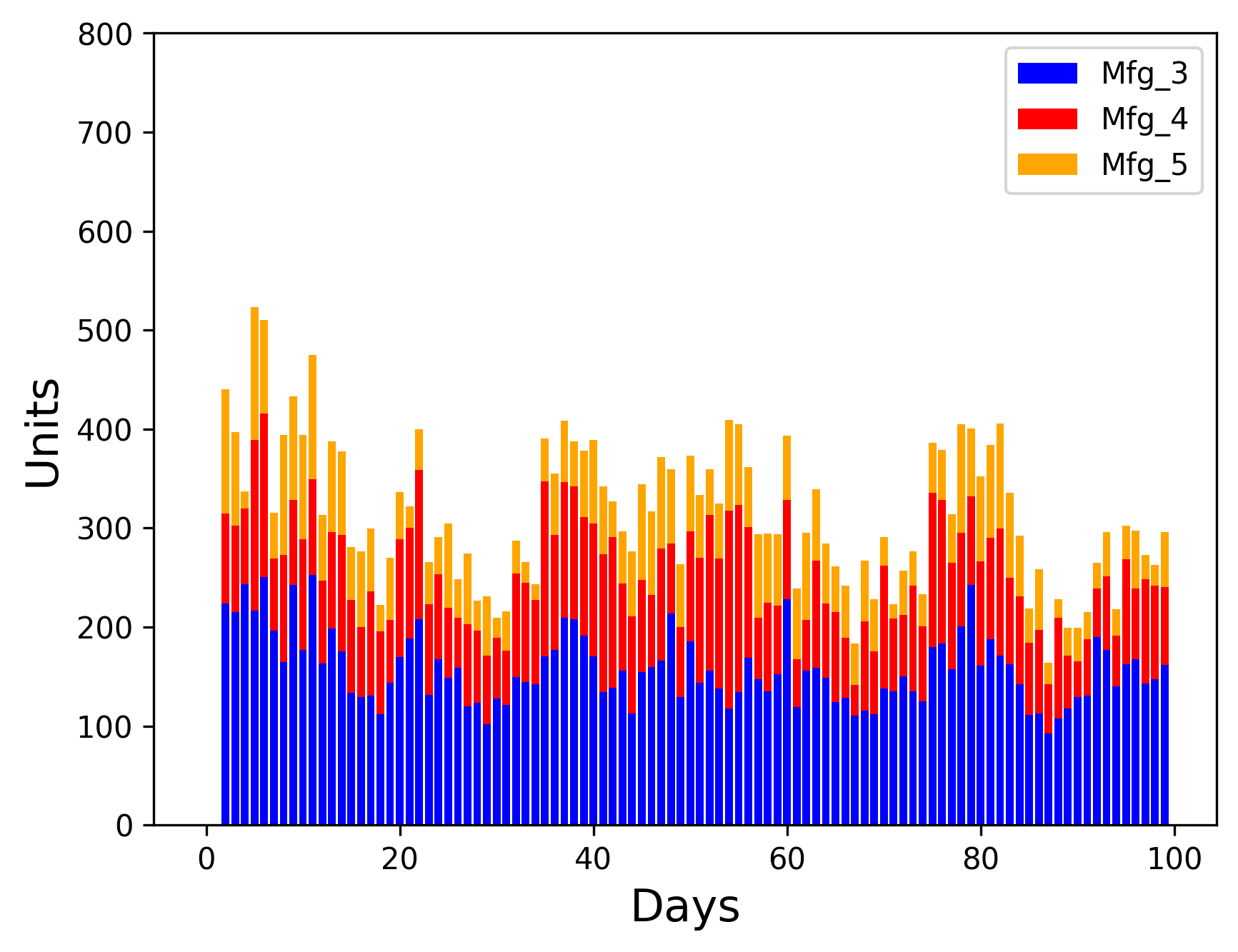}
        \caption{MORL/D - Moderate}
        \label{fig:mfg_morld_moderate}
    \end{subfigure}
    \hfill
    \begin{subfigure}{0.32\textwidth}
        \includegraphics[width=\textwidth]{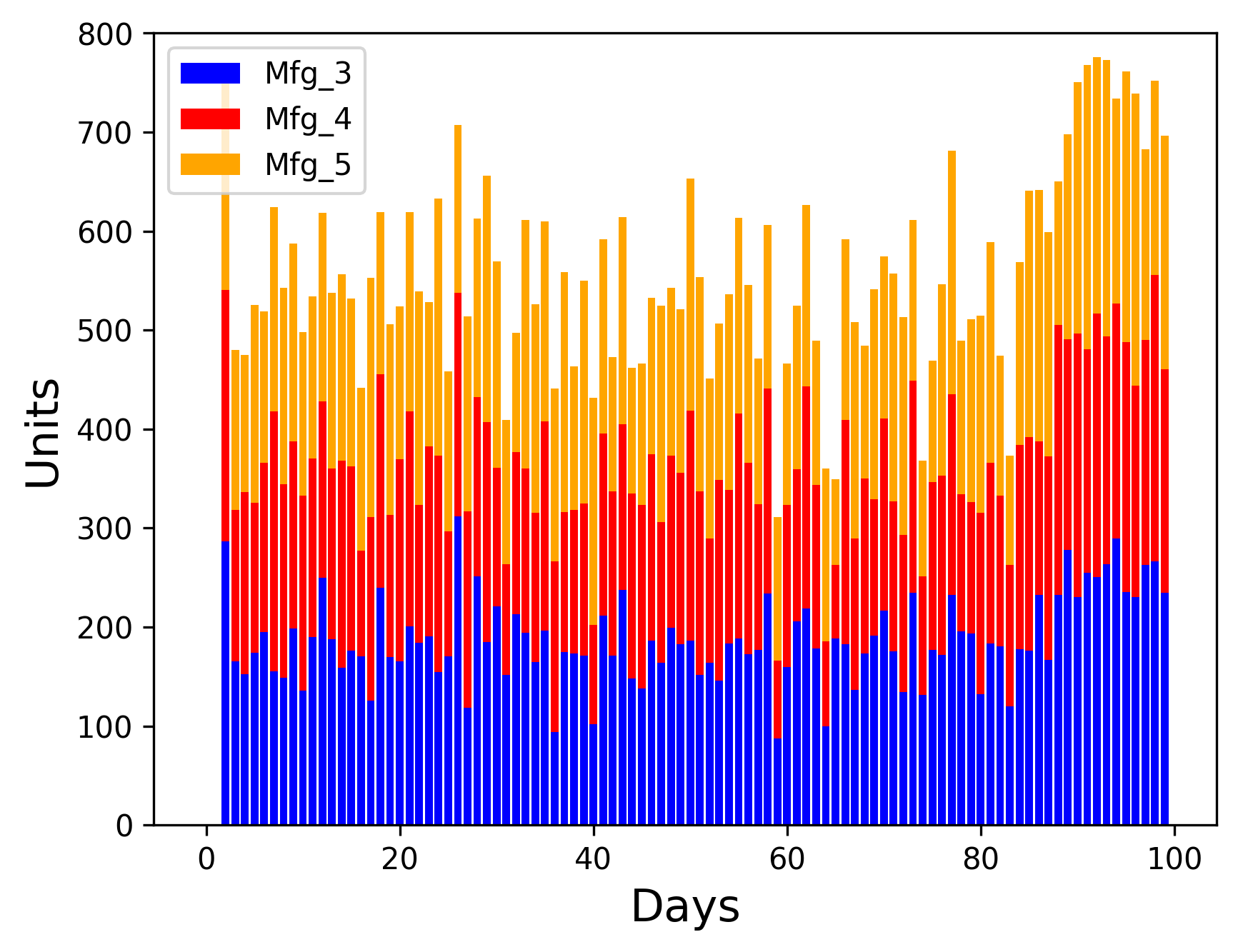}
        \caption{NSGA-II - Moderate}
        \label{fig:mfg_nsga2_moderate}
    \end{subfigure}
    \hfill
    \begin{subfigure}{0.32\textwidth}
        \includegraphics[width=\textwidth]{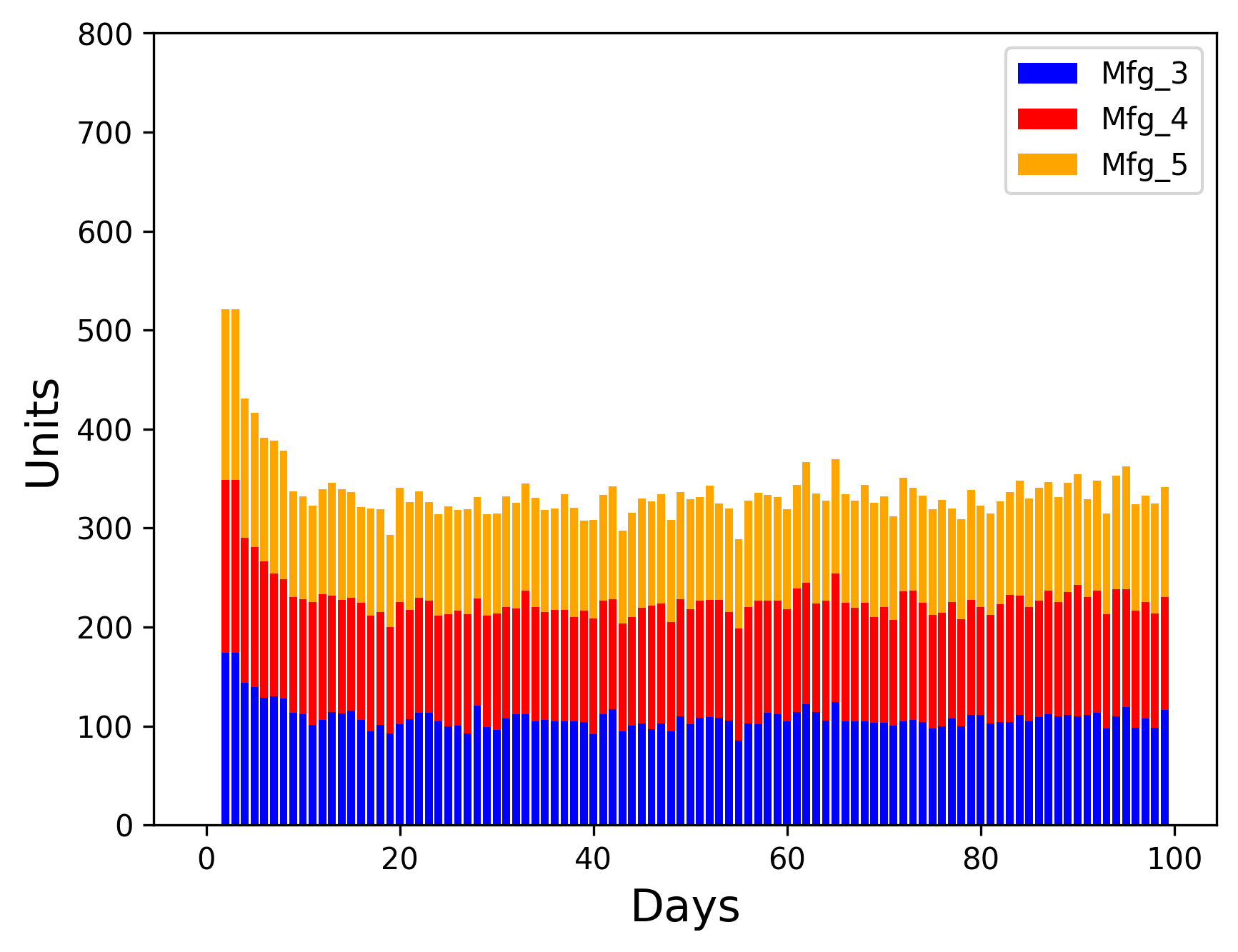}
        \caption{Ours - Moderate}
        \label{fig:mfg_ours_moderate}
    \end{subfigure}

    \begin{subfigure}{0.32\textwidth}
        \includegraphics[width=\textwidth]{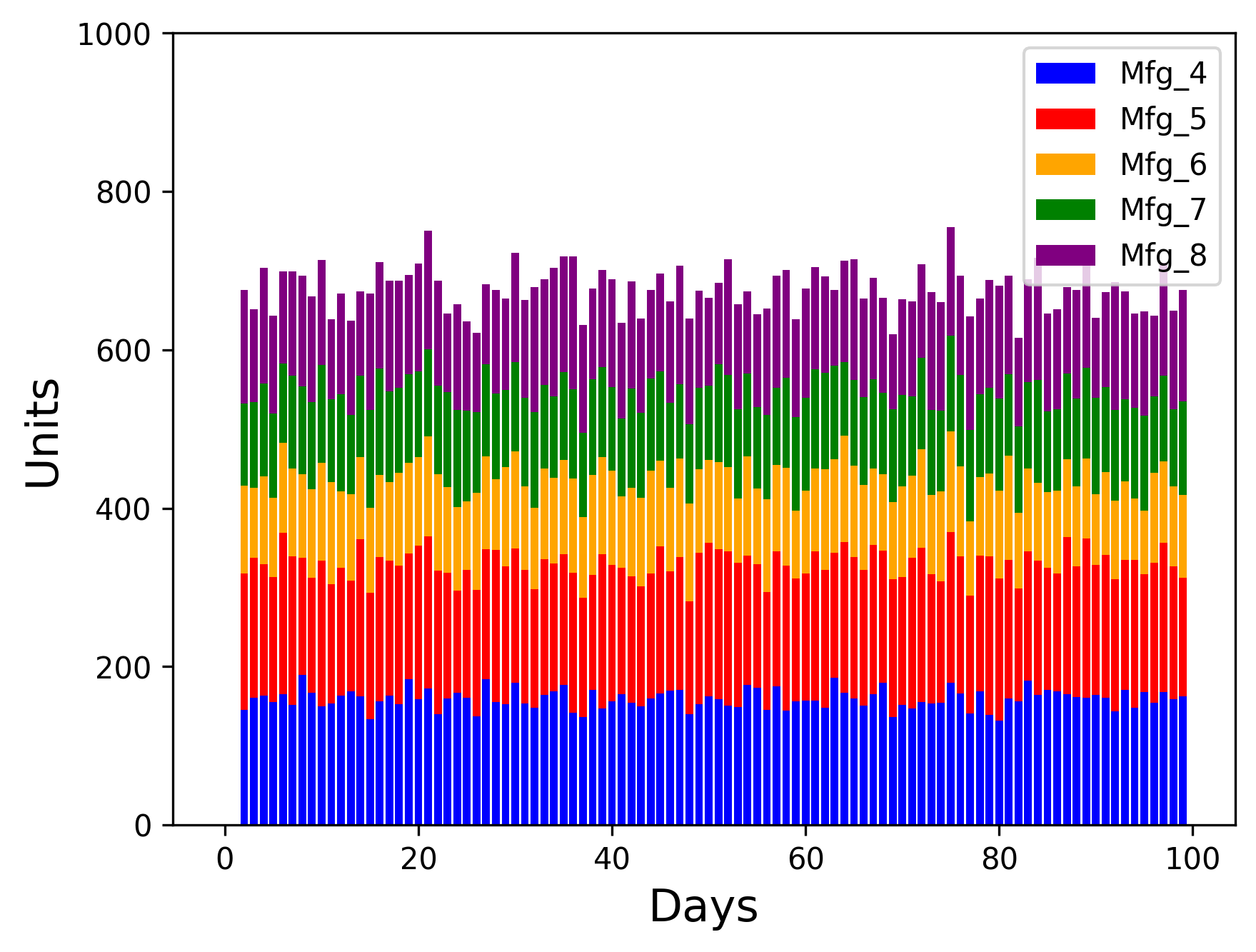}
        \caption{MORL/D - Complex}
        \label{fig:mfg_morld_complex}
    \end{subfigure}
    \hfill
    \begin{subfigure}{0.32\textwidth}
        \includegraphics[width=\textwidth]{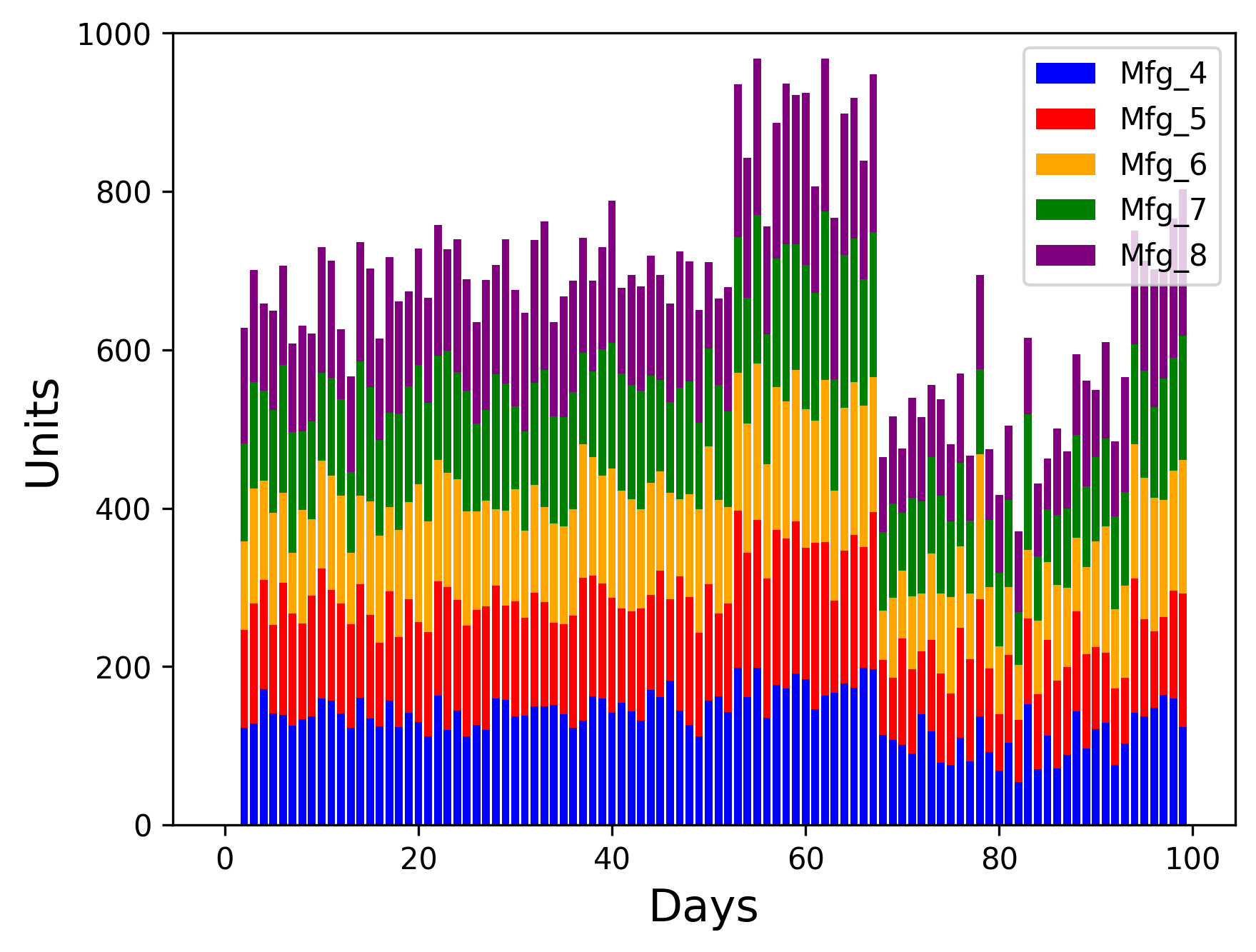}
        \caption{NSGA-II - Complex}
        \label{fig:mfg_nsga2_complex}
    \end{subfigure}
    \hfill
    \begin{subfigure}{0.32\textwidth}
        \includegraphics[width=\textwidth]{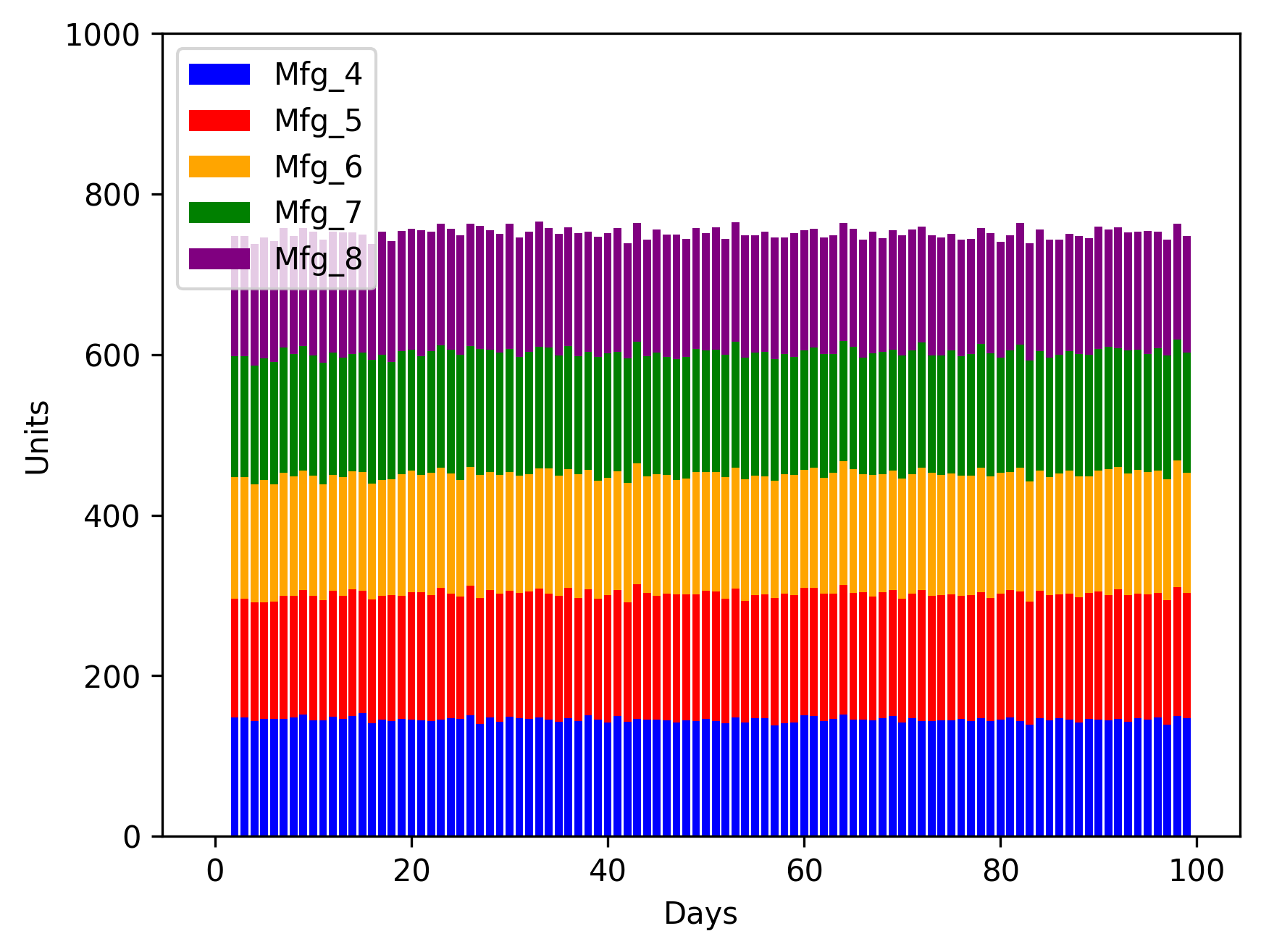}
        \caption{Ours - Complex}
        \label{fig:mfg_ours_complex}
    \end{subfigure}

    \caption{Manufacturing quantities of three algorithms. Our proposed method (\ref{fig:mfg_ours_simple},\ref{fig:mfg_ours_moderate},\ref{fig:mfg_ours_complex}) exhibits the most stable daily production quantities and evenly distributed production amounts across facilities compared to MORL/D and NSGA-II.}
    \label{fig:comp_mfg}
\end{figure*}

% \subsection{Manufacturing Stability}
Figure~\ref{fig:comp_mfg} displays the daily output levels from all manufacturing plants as determined by the solutions obtained from the three methods. We evaluate the operational behaviour of the solutions to understand their practical implications, focusing on manufacturing stability, inventory fluctuations, and unmet demand. A representative solution from the PF approximation set of each method is selected using the approximate proportion of objective values. Our proposed approach consistently yields stable and evenly distributed manufacturing quantities, avoiding the erratic production patterns observed in MORL/D and NSGA-II. This stability reflects the generalisation property of meta-learning, which prioritises consistent behaviour over reactive adjustments.

\begin{figure*}
    \centering
    \begin{subfigure}{0.32\textwidth}
        \includegraphics[width=\textwidth]{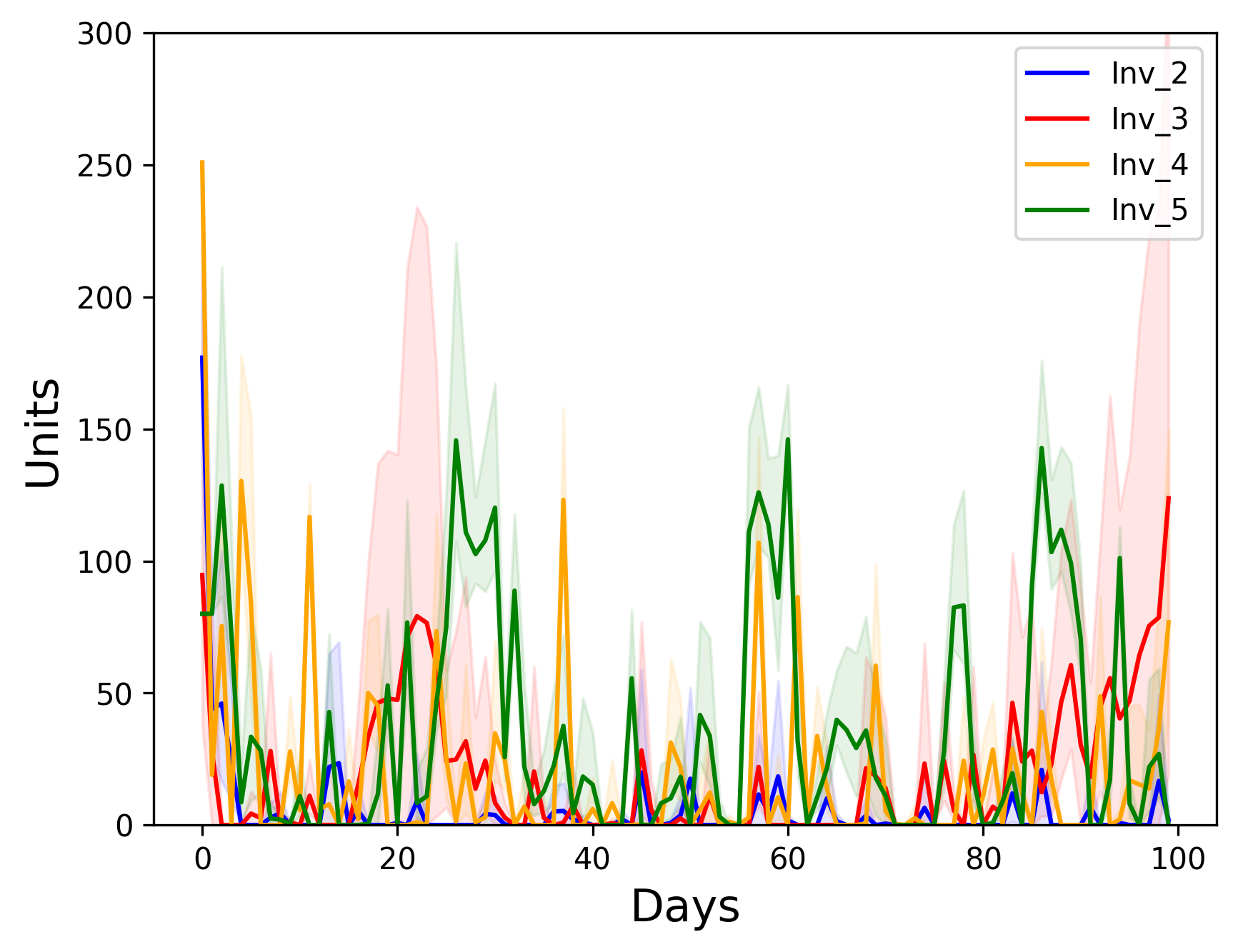}
        \caption{MORL/D - Simple}
        \label{fig:inv_morld_simple}
    \end{subfigure}
    \hfill
    \begin{subfigure}{0.32\textwidth}
        \includegraphics[width=\textwidth]{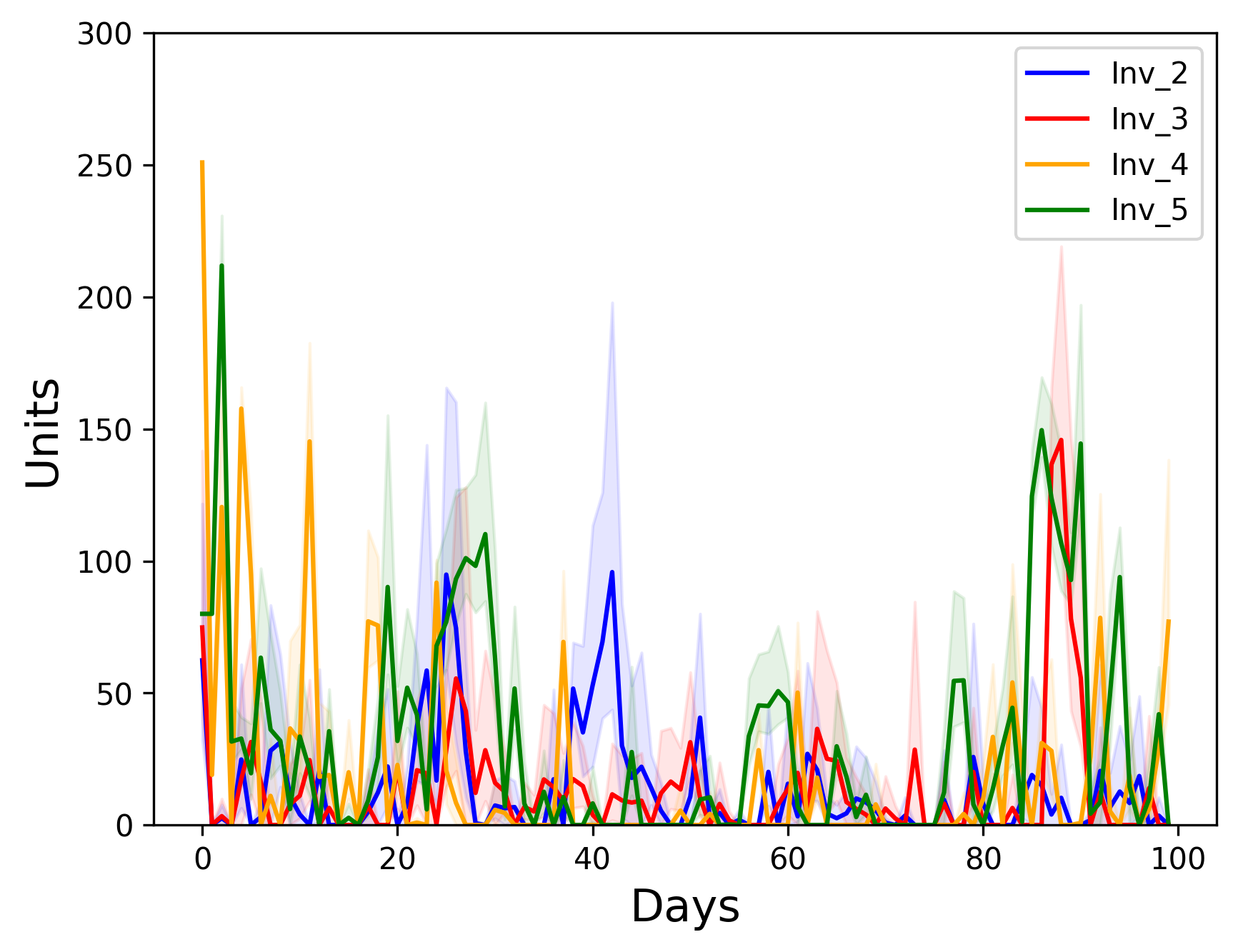}
        \caption{NSGA-II - Simple}
        \label{fig:inv_nsga2_simple}
    \end{subfigure}
    \hfill
    \begin{subfigure}{0.32\textwidth}
        \includegraphics[width=\textwidth]{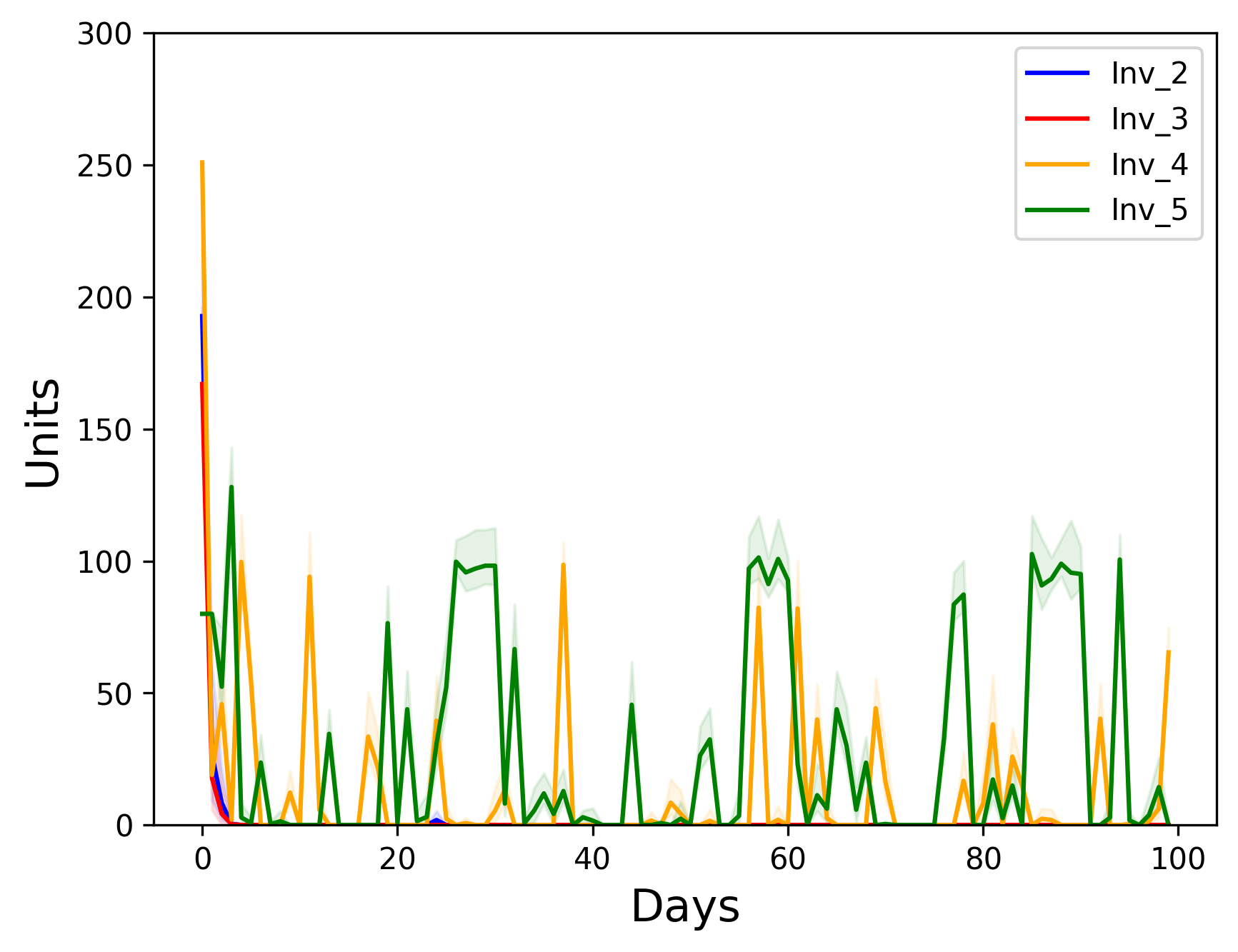}
        \caption{Ours - Simple}
        \label{fig:inv_ours_simple}
    \end{subfigure}

    \begin{subfigure}{0.32\textwidth}
        \includegraphics[width=\textwidth]{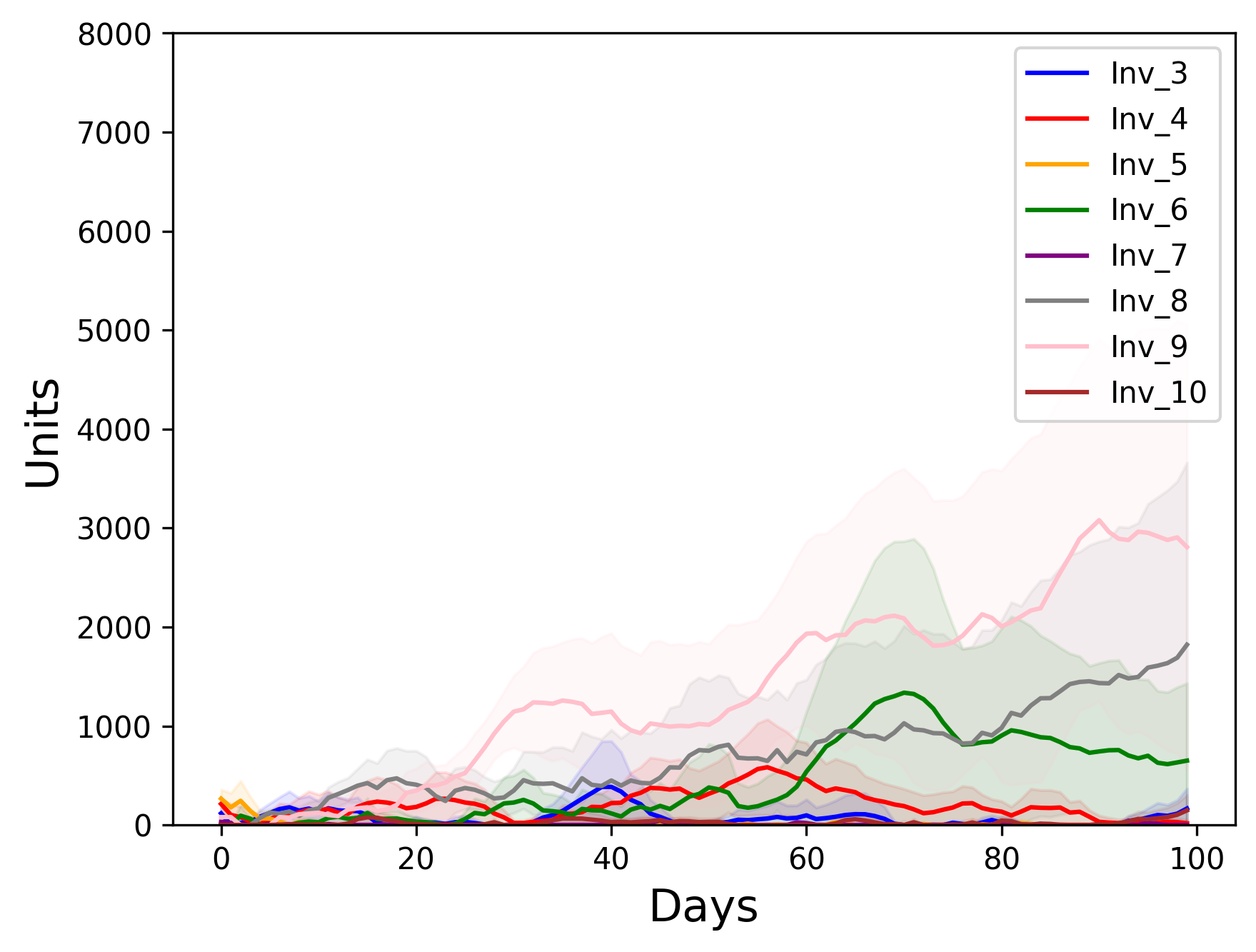}
        \caption{MORL/D - Moderate}
        \label{fig:inv_morld_moderate}
    \end{subfigure}
    \hfill
    \begin{subfigure}{0.32\textwidth}
        \includegraphics[width=\textwidth]{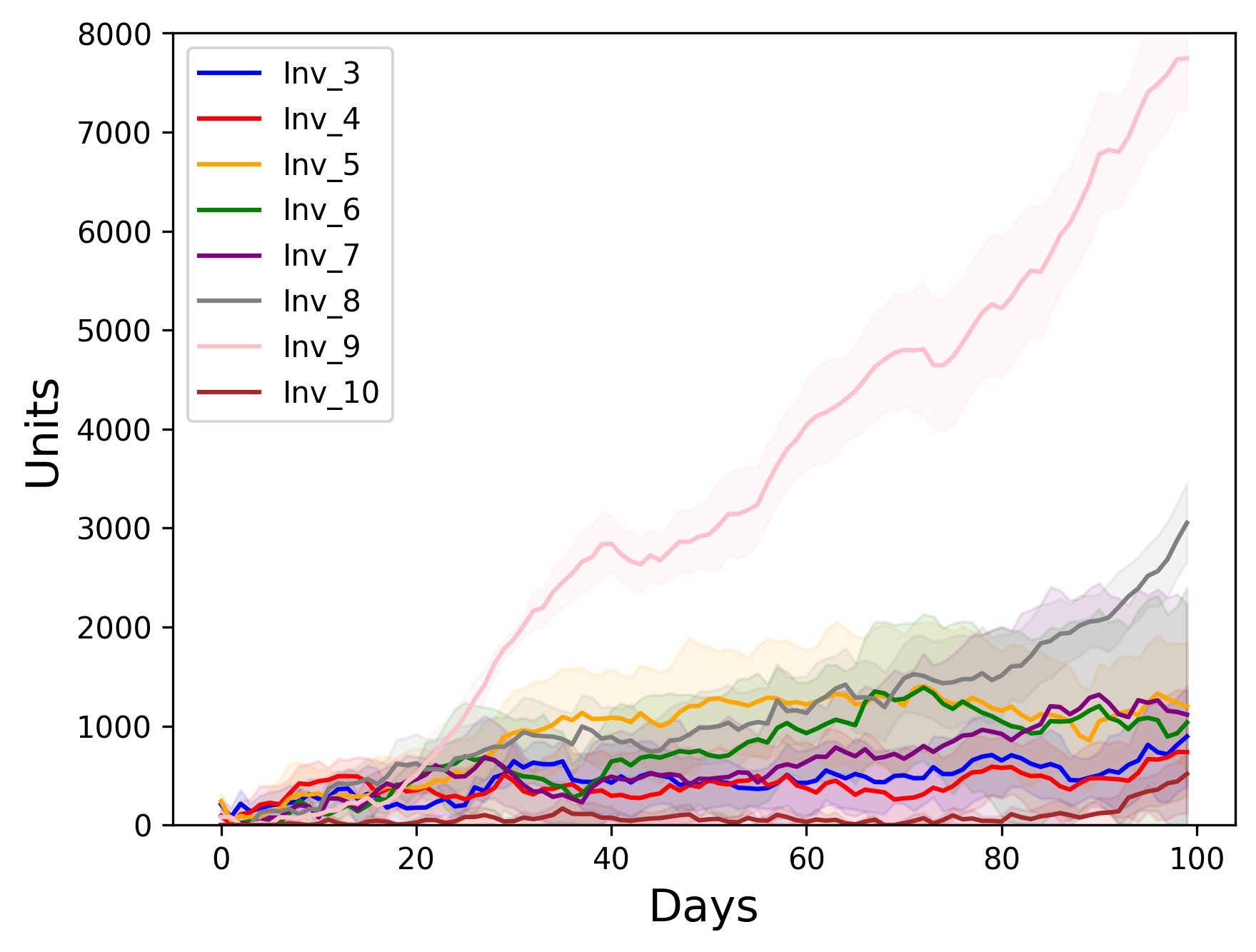}
        \caption{NSGA-II - Moderate}
        \label{fig:inv_nsga2_moderate}
    \end{subfigure}
    \hfill
    \begin{subfigure}{0.32\textwidth}
        \includegraphics[width=\textwidth]{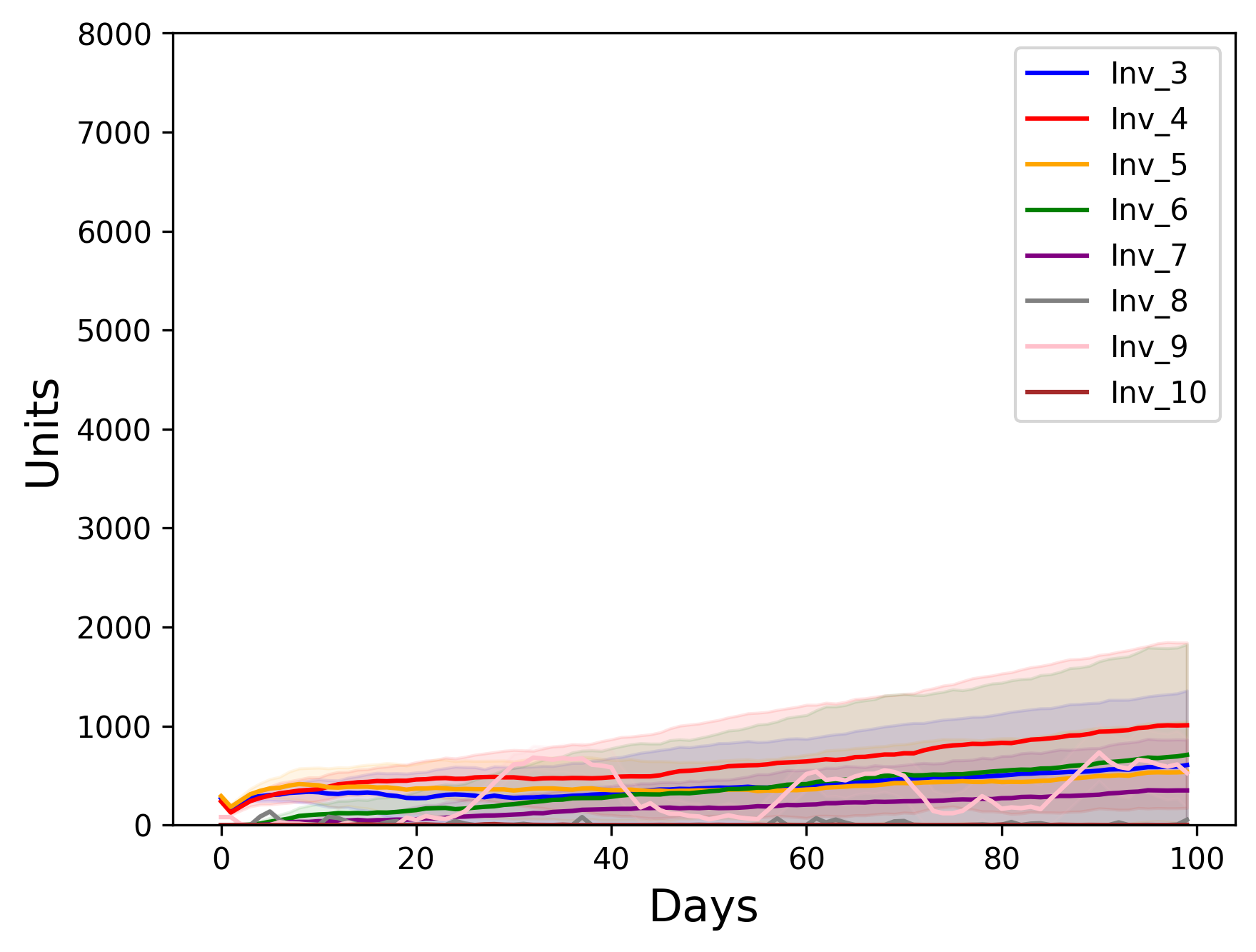}
        \caption{Ours - Moderate}
        \label{fig:inv_ours_moderate}
    \end{subfigure}

    \begin{subfigure}{0.32\textwidth}
        \includegraphics[width=\textwidth]{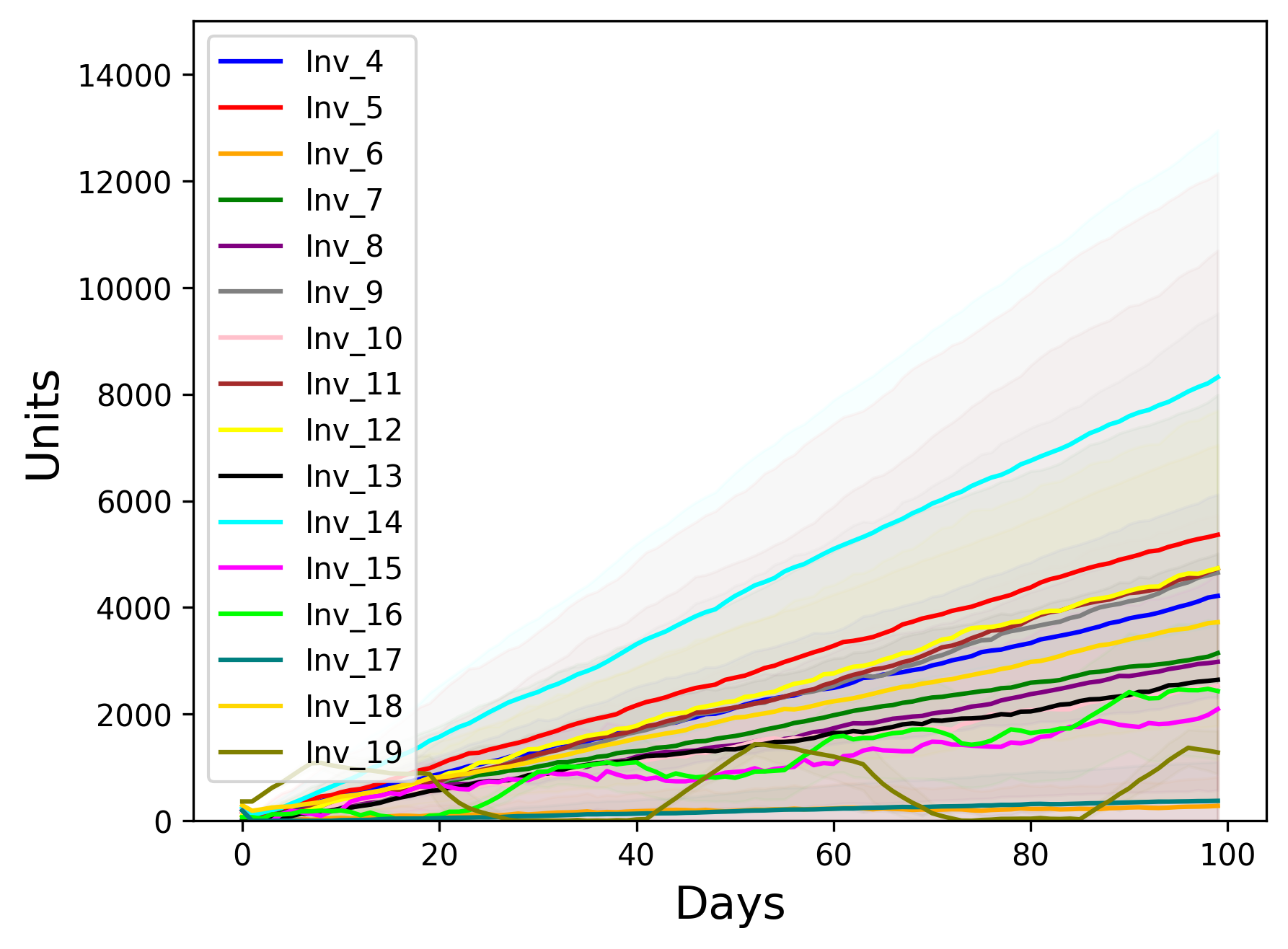}
        \caption{MORL/D - Complex}
        \label{fig:inv_morld_complex}
    \end{subfigure}
    \hfill
    \begin{subfigure}{0.32\textwidth}
        \includegraphics[width=\textwidth]{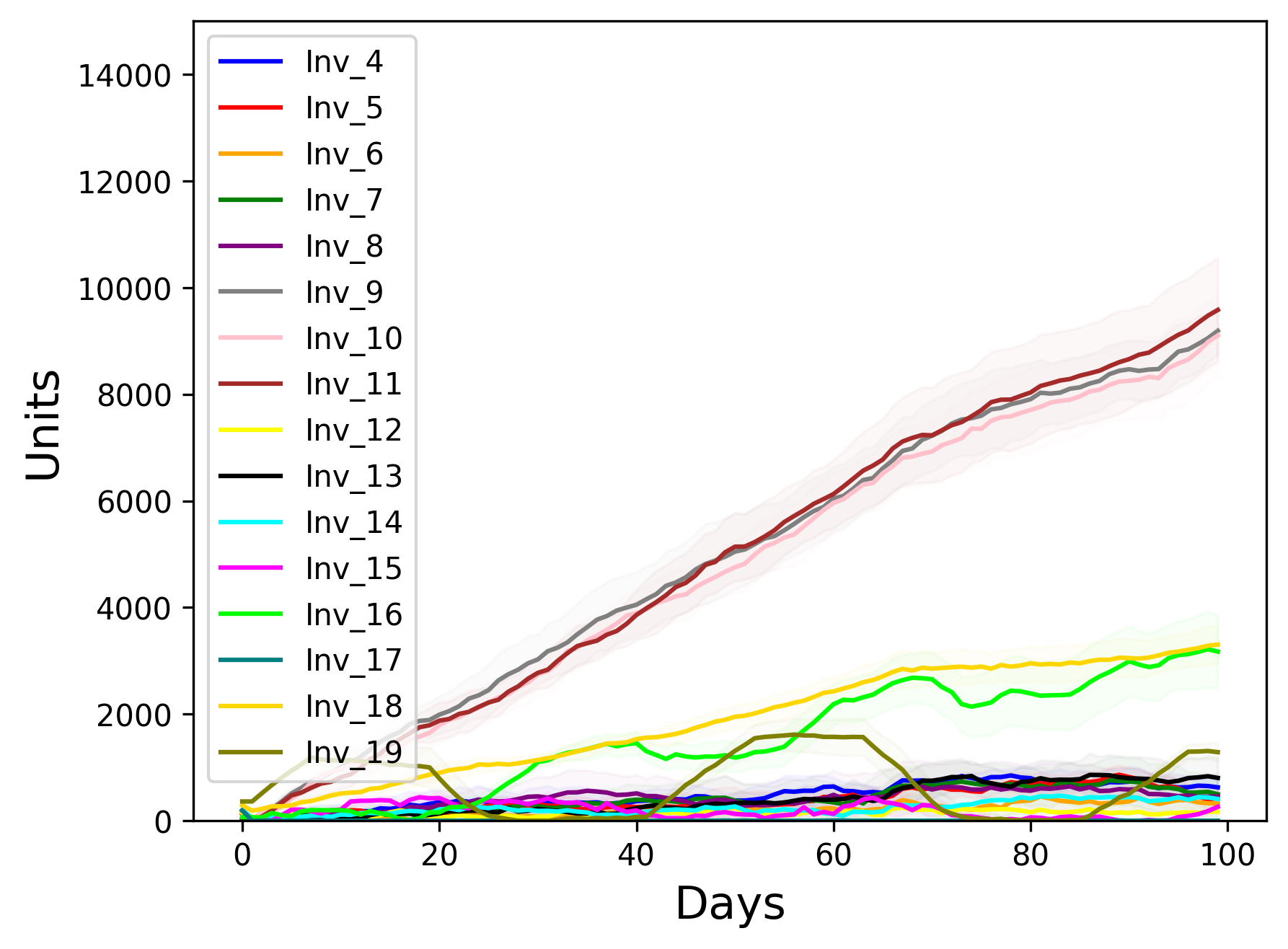}
        \caption{NSGA-II - Complex}
        \label{fig:inv_nsga2_complex}
    \end{subfigure}
    \hfill
    \begin{subfigure}{0.32\textwidth}
        \includegraphics[width=\textwidth]{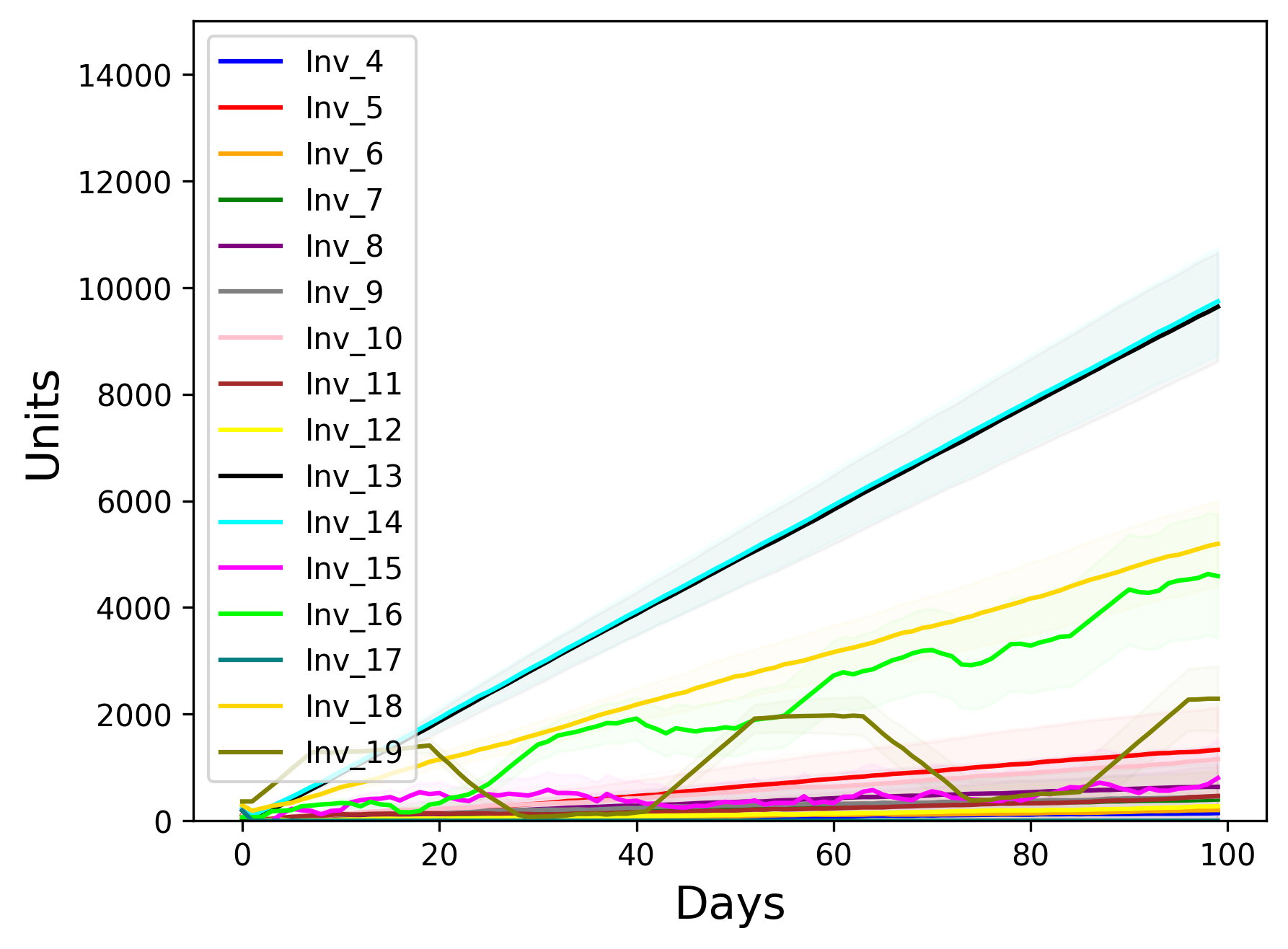}
        \caption{Ours - Complex}
        \label{fig:inv_ours_complex}
    \end{subfigure}
    \caption{The inventory levels yielded from the three algorithms in various complexities. Our proposed method (\ref{fig:inv_ours_simple},\ref{fig:inv_ours_moderate},\ref{fig:inv_ours_complex}) exhibits relatively lean and stable inventory levels daily compared to conventional RL. The average inventory levels tend to increase with the problem complexities for all algorithms.}
    \label{fig:comp_inv}
\end{figure*}

% \subsection{Inventory Stability}
Figure~\ref{fig:comp_inv} depicts the comparison of daily inventory levels across the three supply chain problems when using MIRACL compared to traditional methods. In terms of inventory management, our proposed method maintains leaner and stable inventory levels with low variance, compared to MORL/D and NSGA-II. In complex problems, it buffers selectively at key facilities to manage uncertainty while avoiding widespread overstocking. Regarding unmet demand, our proposed method achieves a comparable demand loss in all problems without excessive inventory. These behaviours highlight the practicality of meta-learning-based approaches for achieving stable and balanced operations in SC environments.

\subsection{Demand Loss} \label{sec:demand_loss}
\begin{figure*}
    \centering
    \begin{subfigure}{0.32\textwidth}
        \includegraphics[width=\textwidth]{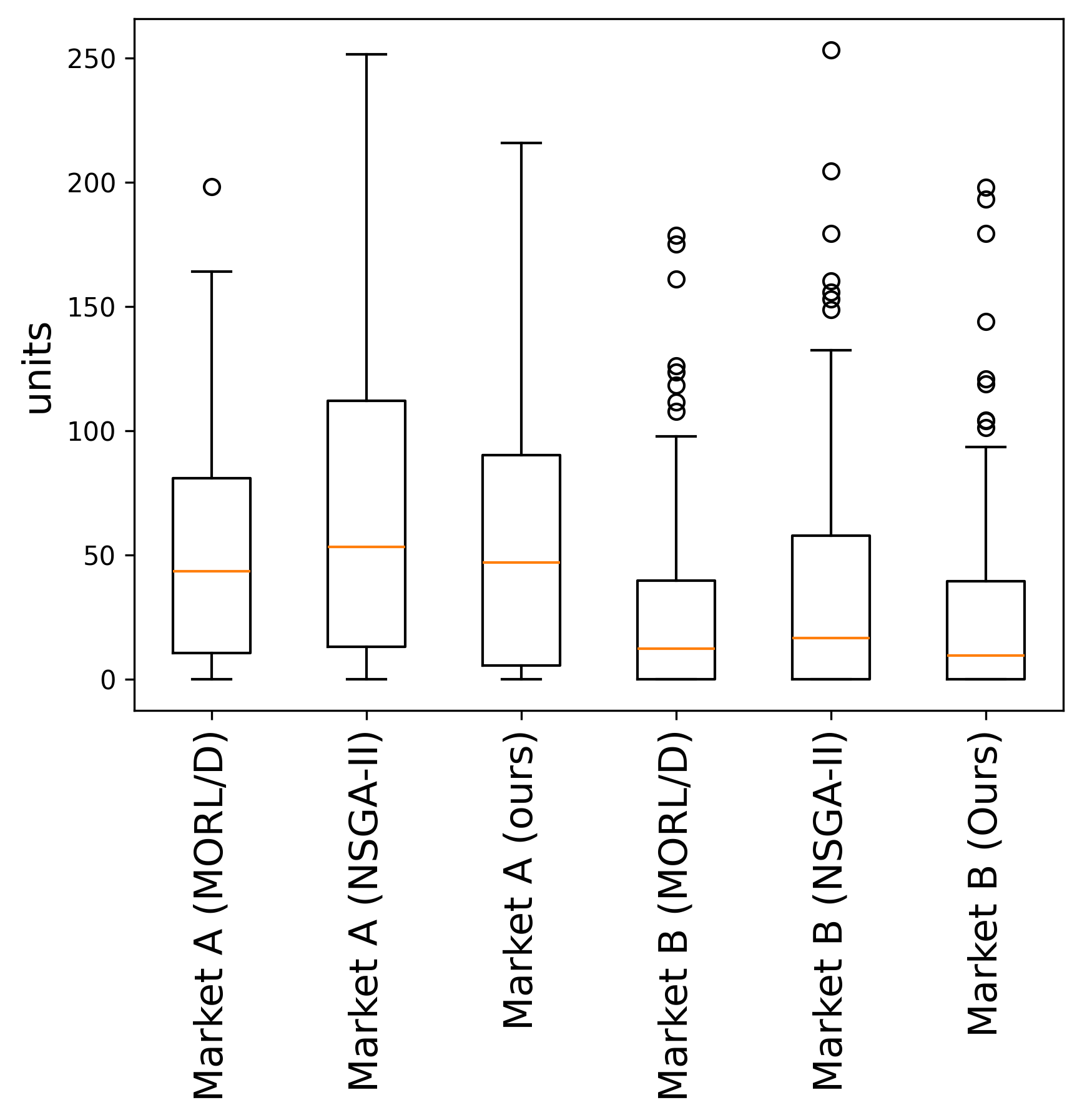}
        \caption{Simple}
        \label{fig:comp_demand_loss_simple}
    \end{subfigure}
    \hfill
    \begin{subfigure}{0.32\textwidth}
        \includegraphics[width=\textwidth]{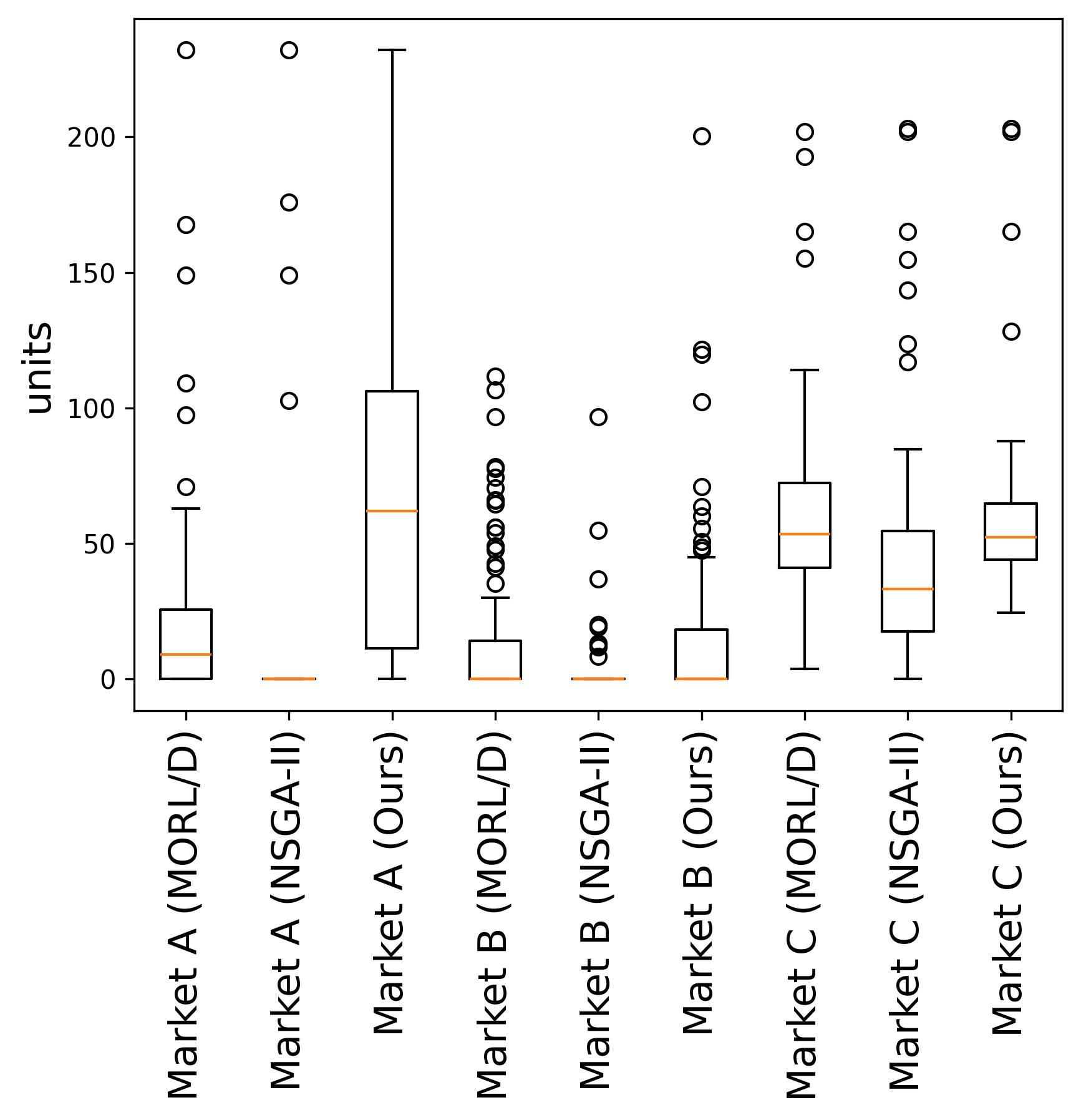}
        \caption{Moderate}
        \label{fig:comp_demand_loss_moderate}
    \end{subfigure}
    \hfill
    \begin{subfigure}{0.32\textwidth}
        \includegraphics[width=\textwidth]{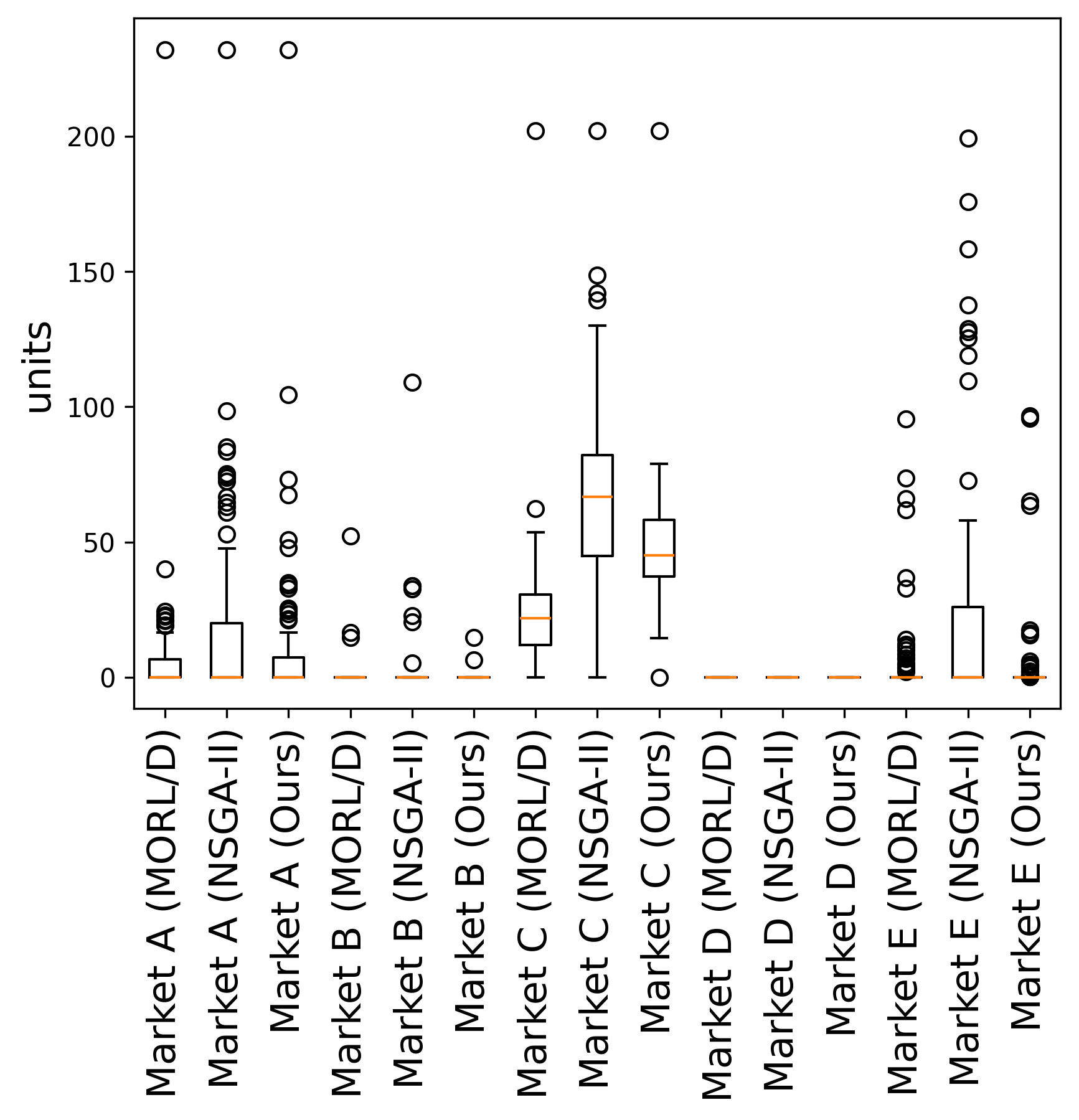}
        \caption{Complex}
        \label{fig:comp_demand_loss_complex}
    \end{subfigure}
    
    \caption{Unmet demand resulting from the three algorithms across various problem complexities. Our proposed method mostly shows proximity values to MORL/D, which is lower than NSGA-II, especially in the simpler problems.}
    \label{fig:comp_demand_loss}
\end{figure*}

Figure~\ref{fig:comp_demand_loss} shows the comparison of the loss of demand in all corresponding markets and in all methods.

